%% file: main.tex
\documentclass[11pt, a4paper, twocolumn, nonumbering, progresssphere]{alphaurbanism}

\usepackage[authoryear, sort&compress, round]{natbib}
\usepackage{multirow}
\usepackage{listings}
\usepackage{caption}
\usepackage{cuted}
\usepackage{subcaption}
\usepackage{amsmath}
\usepackage{booktabs}
\usepackage{xcolor}
\usepackage{graphicx}

\ifdefined\XeTeXversion
    \microtypesetup{tracking=false}
\fi

\newcommand{\beginsupplement}{%
        \setcounter{tocdepth}{3}
        \setcounter{secnumdepth}{3}
        \renewcommand{\thesection}{S\arabic{section}}%
        \setcounter{subsection}{0}
        \renewcommand{\thesubsection}{S\arabic{subsection}}%
        \renewcommand{\theHsubsection}{supplement.\arabic{subsection}}%
        \setcounter{subsubsection}{0}
        \renewcommand{\thesubsubsection}{\thesubsection.\arabic{subsubsection}}%
        \renewcommand{\theHsubsubsection}{supplement.\arabic{subsection}.\arabic{subsubsection}}%
        \setcounter{equation}{0}
        \renewcommand{\theequation}{S\arabic{equation}}%
        \renewcommand{\theHequation}{supplement.\arabic{equation}}%
        \setcounter{table}{0}
        \renewcommand{\thetable}{S\arabic{table}}%
        \setcounter{figure}{0}
        \renewcommand{\thefigure}{S\arabic{figure}}%
     }

\title{AlphaEarth distinguishes cities but compresses urban variation}

\correspondingauthor{andrew.renninger@glasgow.ac.uk}

\author[1]{Andrew Renninger}

\affil[1]{School of Geographical \& Earth Sciences, University of Glasgow}

\begin{abstract}
Cities differ in built form, land cover and development history, complicating comparison across places and time. Satellite foundation models map Earth's surface onto common numerical representations. Yet the tasks and targets used to shape them typically do not focus on cities: globally consistent labels for urban function do not exist, and many datasets---especially land cover and land use classifications---collapse the built environment into few classes. Here we audit the representation, focusing on AlphaEarth but with broader applicability to other Earth embeddings, by probing the geometry and geography of embeddings for 1,000 urban areas in 162 countries. We find that cities occupy a shifted but overlapping region on the hypersphere, 62.7\textdegree{} from the global mean direction, and continent and climate predict 24.3\% of variation among the mean directions of urban centres in excluded countries. Inside cities, degrees of urbanisation carry 8.9\% of the variation, and what they leave holds shared directions whose local orientation varies, not one universal axis of urbanisation. Retained variation is itself unequal: dispersion within urban centres is 14.1\% greater per standard deviation of national development, even after adjusting for population, land area and continent. Further controls suggest cities in developing countries present less contrast in vegetation and texture, and dispersion follows that contrast: full adjustment for it leaves at most 6.4\% of the gradient. Annually, a city's representation moves nearly eight times more than redrawing its own pixels explains, and contracts where the 2022 loss of Sentinel-1B removed a pass direction. AlphaEarth's representations therefore support comparison across regions, while the differences between its annual layers are not yet validated for comparison over time.
\end{abstract}

\begin{document}

\maketitle

\section{Introduction}

Urban environments vary within cities, between cities and over time, and comparative research needs a description that keeps its meaning in every setting it is carried to. Urban typologies supply one by declaring in advance what to measure and how to divide it. Local climate zones sort landscapes by surface structure, cover, materials and human activity, properties chosen because they generate characteristic local temperature regimes \citep{stewart2012local}. Spatial signatures instead cluster measures of form and function into recurrent configurations, and have been fitted across an entire national territory in that form \citep{arribasbel2022spatial,fleischmann2022geographical}. Both systems make heterogeneity between places explicit, and both tie their comparability to a schema: transfer holds where the variables and the classes keep their meaning, and fails where they do not. Further, variation inside a type is not possible in these schemes.

Earth embeddings---which represent raster imagery as vectors---remove the class boundary. Their principal attraction is reuse \citep{rolf2021generalizable,tseng2025galileo,feng2025tessera,alphaearth2025}: expensive processing occurs once and turns satellite archives into features that many later questions can share. With relatively few labelled examples, those features can be paired with a simple classifier, regressor or nearest neighbour search, rather than building a new model from raw observations for every task. The common feature space also permits comparison and search by example: a location of interest can be used to retrieve locations with similar vectors elsewhere using cosine similarity, which for AlphaEarth's unit vectors is simply their dot product \citep{googleearthengine2025similarity}. For urban comparison the gain is real, because cities enter one coordinate system without predefined classifications. A field trained for planetary coverage, continuous as it is, must still spread a bounded representation across every surface on Earth, where urban land is under 1\% of the total land area, and in service of targets given during construction. Uneven representation is thus a concern across Earth embeddings rather than a property of this model alone \citep{danish2026terrafm}: changing a target does not resolve the capacity problem behind it.

Whether it spreads enough of that representation across cities cannot be settled by task accuracy. Evaluation needs labels, and globally consistent labels for urban function do not exist: the schemes that carry functional categories are national or regional, and the products that cover every city grade built-up presence or intensity rather than use. AlphaEarth's own evaluation shows the shape of the gap. Its suite spans 15 assessments over land cover, land use, change detection, crop mapping, tree genera and biophysical regressions, so it is instructive about the problem rather than silent on it \citep{alphaearth2025}. Yet every global or national evaluation in that suite collapses the built environment into comparably few classes; the evaluation carrying functionally differentiated urban categories is confined to Europe; and no evaluation anywhere measures how much variation the representation retains within a class or between cities. That gap follows from what exists rather than from an evaluative decision. Training carries the same asymmetry. The only target with any differentiation inside the built environment uses land cover from the United States, where four developed classes grade intensity rather than function; it was sampled at half of the training rows, weighted below the sensor reconstructions and restricted to the conterminous United States \citep{alphaearth2025}. Where the labels are missing, the representation itself is the only object that can be audited everywhere. 

Earth embeddings arrive in several forms. Most pretrained models are adapted to a task by their users, although there are now formal multimodal benchmarks \citep{lacoste2023geobench,marsocci2024pangaea}; MOSAIKS distributes 2,048 precomputed image features \citep{rolf2021generalizable} while TESSERA and AlphaEarth publish annual global layers at nominal 10\,m resolution \citep{feng2025tessera,alphaearth2025,google2025embeddings}. AlphaEarth is built on a sphere, where each 10\,m cell receives, for each year, a vector of 64 numbers with unit length, so every vector is a point on that sphere and the similarity between two of them is an angle \citep{alphaearth2025}. An encoder reads radar from Sentinel-1 and optical imagery from Sentinel-2 and Landsat \citep{torres2012sentinel1,drusch2012sentinel2,roy2014landsat8,masek2020landsat9}, and during training the model is asked to reproduce a menu of targets, which include terrain, climate, water storage, land cover and text linked to places, alongside the imagery itself. These targets do not have detailed classifications of urban form or function, however. Capacity is also finite. AlphaEarth's own ablation varied the embedding dimension from 32 to 256 along with a noise parameter, and found that the tasks with the most detailed schemas, fine crop types, European land use and American tree genera, did better with more dimensions and less noise than the released model carries \citep{alphaearth2025}. A compact global field can support broad mapping while losing distinctions a particular urban taxonomy needs, or while keeping directions that differ from place to place.

Urban applications of these fields already report signal that changes with the target, the city and the validation design. AlphaEarth predicted the share of commuters driving alone across six United States metropolitan areas with $R^2=0.74$ and the share cycling with $R^2=0.16$ \citep{gong2026urban}, and across 12 cities it separated the presence of slums more readily than their density, returning negative $R^2$ inside the cells slums occupy in every city tested \citep{hou2026slum}. Adding geolocated text on points of interest to the same embeddings improved classification of land use in London by 6.3\%, which places the deficit in semantics rather than in resolution \citep{liu2025aether}. Validation design moves the answer as much as the model does: across 11 models, eight cities and eight tasks, random spatial splits inflated scores and reordered the models relative to splits blocked in space \citep{liu2026cityrep}. Attempts to produce general tests inherit the problems with classification resolution, since few schemes discern variation within cities with global coverage \citep{lacoste2023geobench,marsocci2024pangaea}, and a sequence of local evaluations does not establish that a representation supports urban comparison worldwide. Work on the representations themselves has begun without reaching cities. Across the conterminous United States the number of effective dimensions capturing variance is 13.3 and local tangent spaces rotate substantially \citep{rahman2026geometry}; 2 to 12 of AlphaEarth's 64 dimensions recover 98\% of a baseline for recovering land cover, indicating---though not proving---redundancy \citep{benavides2026alphaearth}. Redundancy against a minimal schema is a different quantity from redundancy for urban comparison, and existing studies do not draw an urban sample. Expanding beyond the United States, no study yet asks how a planetary field allocates its capacity across the world's cities.

We therefore ask four questions of AlphaEarth's released embeddings, sampling 840,776 locations in 1,000 urban areas from 162 countries in each annual layer from 2017 to 2024 \citep{schiavina2019ghsfua,google2025embeddings}. First, do urban samples differ in mean direction and embedding dispersion from the wider global representation? Second, do similarities among cities follow broad geographic context, and how much embedding variation remains inside four standard degrees of urbanisation? Third, is the amount of variation a city retains associated with properties of the city and of its national context? Fourth, do the annual layers move with the city or with the way the city is observed, given that a change of instrument can move an embedding without moving a city? To answer these questions, we introduce a protocol for working with AlphaEarth to understand cities---one that can also be applied to future Earth embeddings. Every observation is treated as a point on the unit sphere, and comparisons between cities are transported to a common tangent frame rather than flattened. We distinguish two moments of a city, its mean direction and its dispersion, and compare both across geographic contexts. Fitted structure is then evaluated with countries excluded, and every claim is checked against the reweighting. This protocol makes the urban claims transportable. Nothing in it is specific to AlphaEarth, and it applies to any planetary field, with modifications for any representation that does not leverage the unit sphere in the manner that AlphaEarth does.

Four results follow. The urban region of the field is shifted but overlapping, and the differences among cities form a continuum rather than a set of global types. Cities have distinct signatures, and geographic context orders the differences between them---albeit without clean separation. Together, continent and climate predict 24.3\% of variation among the mean directions of urban centres in countries excluded from fitting. Cities matched on population alone, within a tenth of a log unit, sit 30.8\% closer when they share both contexts than when they share neither. Within cities, the four degrees of urbanisation account for 8.9\% of embedding variation and 91.1\% remains inside them. A shared path through the urban gradient, according to these degrees, exists but explains 31.9\% of the variation in centred trajectories, and a typical city occupies about six effective dimensions where the pooled residual across all cities occupies 16; further, urban extracts from the field have more effective dimensions than non-urban extracts, but all urbanism has less---indicating that urbanism is complex, but there is global urbanism that is \emph{less} complex than random surfaces on the globe. Capacity is unevenly allocated: one standard deviation higher national development is associated with 14.1\% greater dispersion within urban centres, and 16.7\% more within the whole city. That association survives every annual layer and a comparison matched on land cover, falls by about a third on a common scale when African cities are excluded, and is carried in nearly equal parts by measured vegetation, radar surface and built form, and not at all by population.

Annual motion is large relative to the net displacement it produces, and the difference matters for comparison over time. Across our panel of 1,000 cities, annual city directions accumulate a median 11.45\textdegree{} of path per year but retain only 1.91\textdegree{} per year as net displacement, indicating repeated directional reversal rather than steady drift. A snapshot can move without the city moving in three ways: the pixels drawn in our sample, the observations made and the processing applied. The first is bounded and insufficient, as adjacent years separate a city's mean nearly eight times as far as redrawing its own pixels within a single year does. The second is demonstrably live: when Sentinel-1B failed in 2022, the spread of embeddings in the 159 cities that lost one of its two viewing directions contracted 4.1 percentage points more than in the 841 that kept both, movement in the observing rather than in the city. The third cannot be isolated with public information. A shared direction nevertheless exists: 73.6\% of all 499,500 city pairs end the period closer than they began; the median pair converges 1.74\textdegree{} from 2017 to 2024, with the contraction negative and excluding zero across all degrees. Thus, the released fields measure structure unequally across geography, and the time series is a flawed record of change.

\section{Materials and Methods}
\label{sec:methods}

\subsection{Data and processing}

We begin by sampling and summarising AlphaEarth's dimensions across 1,000 cities; sampling pixels while summarising data at multiple levels of aggregation balances computational demands with our inferential goals. Each sampled location has a unit length AlphaEarth vector with 64 numbers. We average vector directions to describe a degree of urbanisation or a city, then measure embedding dispersion within a city as the mean squared angle of sampled vectors around that city's mean direction. We compare these mean directions and dispersions with a global reference and across geographic and national development contexts. The unit-length constraint also fixes the total second moment per observation, so dispersion is a share of a bounded budget and every comparison of it is comparative by construction.

\subsubsection{Study sample and AlphaEarth observations}

To compare urban variation across a broad range of places, we use ``Functional Urban Areas'' rather than administrative boundaries. These areas join a densely settled core to its surrounding commuting zone and therefore provide a comparable definition of a city across countries. We select 1,000 areas with at least 250,000 residents in 2015 from the GHS--OECD Functional Urban Areas catalogue \citep{schiavina2019ghsfua}. The sample contains 162 countries on five continents. That threshold defines the frame, and it excludes the smaller settlements where most current urban growth is occurring.

To distinguish variation associated with urban intensity from variation among cities, we assign every sampled location to one of four 2020 Degree of Urbanisation classes from the Global Human Settlement Layer \citep{schiavina2023ghssmod}: suburban or peri-urban (PU), semi-dense urban cluster (SD), dense urban cluster (DU) and urban centre (UC). These labels are native at 1\,km; sampling them at finer spacing does not increase their information resolution. We then sample valid observations from the 2024 AlphaEarth field distributed through Google Earth Engine \citep{google2025embeddings}. AlphaEarth describes each location with 64 numerical components at a nominal 10\,m resolution. The released annual layers are the product of model version v2.1, while the published description of AlphaEarth describes v2.0 \citep{alphaearth2025}, a boundary we mark wherever a published quantity is invoked (Supplementary Section~\ref{sec:supp-sampling}). We remove negligible drift from unit length but otherwise analyse the released values directly, which the catalogue documents as already unit-length.

Within each Functional Urban Area, we request as many as 250 observations from each degree of urbanisation at a 30\,m sampling scale. The 2024 table contains 840,776 observations, and every city contributes at least one class. Analyses using all 1,000 cities therefore describe each city's available stratified sample, which can contain between one and four classes. The covariance, transfer and decomposition analyses that need equal samples in every class use 544 cities with exactly 250 observations in each class, giving 544,000 observations across 117 countries. We calculate city and class mean directions directly from the relevant pixels, giving every city equal weight in comparisons among cities; they do not weight the complete urban footprint by physical area or population. Further information about sample composition, sampling and the construction of city averages is provided in Supplementary Fig.~\ref{fig:si-data-description} and Supplementary Section~\ref{sec:supp-sampling}.

The temporal analyses contain annual AlphaEarth summaries for 1,000 cities in 162 countries. Estimands that compare a city's complete path require the same city in all eight annual layers; all 1,000 cities satisfy that requirement and are retained in every year, giving 8,000 city-years. Each city's set of 2020 degrees of urbanisation and its sampled count are held fixed through time. The comparison of city pairs within each degree of urbanisation additionally requires every city to have all four degrees in every year. Of the 1,000 starting cities, 538 cities in 117 countries meet that condition and are retained in each year, giving 4,304 city-years.

\subsubsection{Reference and contextual data}

To place the urban observations within AlphaEarth's wider representation of the planet, we draw a separate reference sample. We divide the Earth into equal area cells, sample AlphaEarth wherever the 2024 field is valid, and keep the observations that carry an ESA WorldCover label---e.g.\ ``built-up'' or ``cropland''. This produces 247,565 observations over land, in 23 of the 24 cells. Equal candidate density among the cells makes the retained sample approximately representative of WorldCover land within AlphaEarth's valid 2024 footprint (Supplementary Section~\ref{sec:supp-sampling}).

To describe the context of each city, we use continent and broad Köppen--Geiger climate family \citep{beck2023koppen}. Population and land area come from the Functional Urban Area catalogue. We measure national development with the 2023 Human Development Index (HDI) from the Human Development Report 2025 time series \citep{undp2025hditimeseries} and treat HDI as national context rather than a city attribute. HDI is unavailable for 15 cities, and analyses of a named urban class also require sufficient observations from that class; the primary model of dispersion within urban centres contains 977 cities in 157 countries. Because HDI is standardised inside each estimation sample, its standard deviation varies across the samples we compare---from 0.110 in the sample that omits Africa to 0.168 in the sample that omits Asia---so coefficients expressed per standard deviation are not interchangeable across those samples. To test whether physical form can account for the HDI association, we sample population from the 2020 GHS population grid \citep{schiavina2023ghspop} and building height, area, volume and fraction from World Settlement Footprint 3D v0.2 \citep{esch2022wsf3d} at the same sampled locations within urban centres. These products describe measured physical form and product coverage; they are not contemporaneous ground measurements for 2024.

\subsection{Modelling and analysis}

\subsubsection{Spherical comparisons}

To preserve the geometry of AlphaEarth's released representation, we treat every vector as a point on the unit sphere in 64 dimensions. A larger angle means that two embeddings are less similar. The angular distance between two normalised vectors $x$ and $y$ is

\begin{equation}
d(x,y)=\arccos\!\left[\operatorname{clip}(x^{\mathsf T}y,-1,1)\right].
\end{equation}

We represent a set of observations by its spherical barycentre, defined as the normalised average direction and called the mean direction hereafter. For city $c$, embedding dispersion is the mean squared angle between its observations and its own mean direction $\mu_c$,

\begin{equation}
D_c=\frac{1}{n_c}\sum_i d(x_{ci},\mu_c)^2.
\end{equation}

These operations allow position and internal variation to be distinguished: two cities can have similar mean directions but different embedding dispersion. When variation from different cities must be compared directly, we map observations to the tangent plane at each city's mean direction and move those tangent vectors to a common reference while preserving their lengths and mutual angles. This puts local deviations around different city means into one coordinate system without treating the spherical field as flat. Definitions and derivations are given in Supplementary Section~\ref{sec:supp-geometry}.

\subsubsection{Urban position and geographic structure}

To ask whether cities occupy the same part of AlphaEarth space as the wider planet, we measure the angle between the equally weighted overall urban mean direction and the global mean direction. We describe the angular spread of city mean directions around the overall urban mean and compare those radii with the distribution of global pixels around the global mean. We then test whether a city's distance from the overall urban mean direction is related to its embedding dispersion. To ask whether cities are individually identifiable, we split each city's 2024 urban pixels into two random halves and compute a mean direction from each half of every city. Each city's second half, and each single pixel within it, is then assigned to the nearest of the 1,000 mean directions built from the first halves, with the split repeated five times; a bootstrap of every city's mean direction at 250 pixels sets the angular scale below which two cities cannot be told apart (Supplementary Section~\ref{sec:supp-geography} and Supplementary Table~\ref{tab:si-city-identification}).

To determine whether similarities among cities follow broad geography, we first cluster the angular distances among all 1,000 city mean directions without supplying geographic labels. We then match each city to cities of similar population under four combinations of shared or different continent and climate; we select matches without using AlphaEarth distance or HDI. Finally, we learn the multivariate differences associated with continent and climate from training countries, predict city mean directions in countries excluded from fitting and measure the share of variation explained in those countries. These analyses test, respectively, whether differences are well described by a hierarchy, whether matched cities are closer in shared contexts and whether geographic structure transfers to new national settings. Exact matching rules, uncertainty procedures and validation splits are described in Supplementary Section~\ref{sec:supp-geography}.

\subsubsection{Variation within standard urban classes}

To measure how much embedding variation the four degrees of urbanisation express, we use the 544 cities with exactly 250 observations in every class. Within each city, we separate total tangent variation into differences among the four class mean directions and differences among observations assigned to the same class. Because the classes are native at 1\,km while AlphaEarth is nominally 10\,m and we sample it here at 30\,m, this partition compares a broad classification with embedding variation at finer spatial scales; it does not by itself assign meaning to the residual. To test whether the residual variation is reproducible rather than noise, we learn subdivisions within each class in one set of countries and evaluate how much they reduce distortion in another. Independently mapped population, vegetation, built fraction, building volume and night lights supply an external check. The country split, dimension reduction and evaluation procedure are reported in Supplementary Section~\ref{sec:supp-geography}.

\subsubsection{National development and urban dispersion}

To test whether AlphaEarth expresses a comparable amount of urban embedding variation across development contexts, we fit ordinary least squares models to the natural logarithm of embedding dispersion. The principal specification is

\begin{equation}
\log D_c=\alpha+\beta H_c+\gamma P_c+\delta A_c
+\eta I_c+\lambda_{k[c]}+\varepsilon_c,
\end{equation}

where $H_c$ is national HDI, $P_c$ and $A_c$ are $\log_{10}$ population and land area, $I_c$ identifies the collection stage and $\lambda_{k[c]}$ represents continent. We centre continuous predictors and divide them by their standard deviation within each outcome sample. We report $100[\exp(\beta)-1]$, the percentage difference in embedding dispersion associated with one standard deviation higher HDI. The outcomes are embedding dispersion across all available urban observations, embedding dispersion over every valid cell of the city's urban centres and mean embedding variation within the four equally sampled classes. Standard errors and intervals allow cities in the same country to be statistically dependent.

The second of these is the principal outcome and a population reduction rather than a sample statistic: Earth Engine returns it over every valid cell of the urban centres the city contains rather than over the 250 sampled pixels. Neither the cell count nor the extent of the urban centre carries the association, and the same estimand recomputed on the 250 sampled pixels gives 16.0\% (9.0 to 23.5\%) on the same cities (Supplementary Section~\ref{sec:supp-development}).
That dependence is uneven. In the primary sample of urban centres in 977 cities, China contributes 109 cities, India 108 and the United States 39, while 40 countries contribute a single city each, so Kish's effective cluster count is 30.7 against 157 nominal countries and a correction indexed to the nominal count is optimistic. We therefore refit the primary coefficient with a wild cluster bootstrap that imposes the null and resamples whole countries, and quote its interval alongside the interval from clustered standard errors where the coefficient is first reported (Supplementary Section~\ref{sec:supp-development}).

We test whether the association transfers to countries excluded from fitting, persists within the two collection stages and continents, depends on any one country or concentrates in a few directions of the representation. We also repeat the same fixed-sample model in every annual embedding layer from 2017 to 2024, give each country equal total weight, replace continent with UN subregion and omit each continent in turn. We summarise each adjustment by the paired difference in the log coefficient rather than by the attenuation ratio it implies: once the adjusted coefficient can change sign, the ratio exceeds 100\% in a share of paired draws and stops reading as a fraction of the gradient removed, whereas the paired difference remains well defined. These supporting specifications and their correction for multiple comparisons are provided in Supplementary Section~\ref{sec:supp-development}.

\subsubsection{Temporal separation and motion}

For the temporal comparison within each degree of urbanisation, we calculate one spherical mean direction for every city, year and degree, as well as a pooled mean direction across all of a city's available urban observations. Among the 538 cities holding all four degrees in every year, we then calculate all 144,453 unordered city-pair distances in every year and degree. Panel medians give every pair equal weight. For each pair, we also count sign reversals across the six boundaries between its seven consecutive annual changes, and we compare the observed counts with the exact expectation under an exchangeable ordering of the eight annual values. For the number of pairs that close or open at every transition, the null is exact and synchronous: we apply each of the 40,320 orderings of the eight calendar years to the whole annual field at once and recount, which preserves the dependence among pairs that share a city. Because pooling within each city needs no such restriction, we also compute the pooled track on all 1,000 cities and their 499,500 pairs, with 1,000 bootstrap draws that resample countries and then cities; the 538-city sample remains where the four degrees are compared on identical pairs. We quantify uncertainty in annual medians and paired 2017--2024 endpoint changes by sampling countries and then cities within sampled countries 500 times while retaining complete annual trajectories, excluding the degenerate zero-distance pairs a city drawn twice would otherwise contribute.

For motion at the level of the city, we use all 1,000 complete trajectories. We log-map the 8,000 annual mean directions, each pooled over all of a city's urban observations, to one common tangent plane, subtract each city's own 2017 tangent position and fit one principal-component analysis to all resulting city-year displacements. The first two scores are used only to display the leading shared directions and magnitudes of change. We quantify motion without dimension reduction directly on the unit sphere: annual path speed is the sum of the seven consecutive great-circle steps divided by seven years, endpoint net speed is the 2017--2024 great-circle distance divided by seven, and straightness is endpoint distance divided by accumulated path length. Intervals for the three median motion statistics come from 5,000 bootstrap replicates that resample countries and then cities. These annual mean directions are repeated cross-sections within fixed degrees of urbanisation, not tracked pixels; their connected paths can contain components of production, observation, sampling and physical change.

To bound one component of that movement, we measure how far an annual mean direction moves when only the sampled pixels change. Within every city, 2020 degree of urbanisation and year holding at least two observations, we partition the sampled pixels eight times into two equal-sized, non-overlapping halves and measure the angular separation of the two half-sample mean directions. Where each half holds $h$ pixels and the full annual sample holds $n$, we estimate the squared error of one full-sample mean direction as the mean squared split separation multiplied by $h/2n$, and the component expected in an adjacent-year separation as the sum of the two annual values. We compare that quantity with observed adjacent-year squared separation across 23,583 year-to-year steps of a city within one degree and 6,930 steps of a whole city. Intervals resample cities as blocks carrying all of their degrees and transitions; a specification weighted by the sampling design is a sensitivity check. This is a spatial sampling floor. It removes the uncertainty that comes from drawing different places inside a fixed degree of urbanisation and leaves annual variation at fixed sites, acquisition conditions, change in the representation across the whole product and physical landscape change in the remainder, so it cannot be read as the share of movement that is reproducible.

We do not treat repeated row indices as fixed cells: each annual export draws its own stratified sample under a year-specific seed and discards point geometry, so matched row identifiers are aliases rather than locations, and the original city tables cannot estimate movement at an identical, unchanged location (Supplementary Section~\ref{sec:supp-temporal-floor}). Two further analyses approach the remainder from opposite ends. To ask how much of the annual movement measured conditions predict, we fit ridge models of each city's 64-dimensional annual change to monthly weather from ERA5-Land, vegetation indices, seasonal timing and retrieval quality from MODIS, and Sentinel-2 acquisition counts and cloud scores \citep{munoz2019era5land,didan2021mod13a2,friedl2022mcd12q2,pasquarella2023cloudscore}, scoring the reduction in squared error on countries and years excluded from fitting. To locate movement where change on the ground is documented, we extract the same 10\,m pixels through all eight annual layers at eleven construction projects in nine countries and compare their displacement with that of fixed locations matched on their starting vectors at least 2\,km away in the same city (Supplementary Section~\ref{sec:supp-movement-tests}).

\subsubsection{Alternative explanations}

Finally, to ask whether less varied measured urban form explains the HDI coefficient, we summarise the distributions of population and four building measures within each city and compare the coefficient before and after adding them to the same model. We then add detailed Köppen--Geiger composition, vegetation level and spread, and radar surface-structure summaries on one fixed sample. Adding the blocks in a fixed order, a ladder, gives each block whatever the blocks before it have left, so we also decompose the same fixed-sample reduction over every ordering of its six blocks and take each block's Shapley value---its contribution averaged over all 720 entry positions---as the primary attribution. That decomposition is exact: the six values sum to the difference between the context and fully adjusted coefficients. Because the ladder's radar block and the surface index described in Supplementary Section~\ref{sec:supp-alternatives} are two representations of the same construct, we fit both on identical rows and separate the effect of the covariate from that of the city summary statistic. Terrain enters as a further block, summarising each urban centre from the 30\,m Copernicus elevation model by the spread of elevation, local relief and topographic position, added after radar surface as an eighth rung of the ladder and as a seventh block in the decomposition on the 937 cities the model covers. As a separate broad land-cover check, we recompute dispersion among built-up pixels, among vegetated pixels and in repeated samples containing exactly 30 pixels of each type per city. To ask whether opportunities for satellite observation account for the association, we measure public Sentinel-1, Sentinel-2 and Landsat coverage at fixed points within urban centres and add scene counts, clear fractions from the Sentinel-2 cloud-score product \citep{pasquarella2023cloudscore} and orbital conditions. We also use the loss of Sentinel-1B between 2021 and 2022 as a discrete change in viewing geometry and test whether it coincides with local change in embedding dispersion. The construction of the physical and environmental summaries, cover matching, mission variables, attenuation estimates and temporal placebo is described in Supplementary Section~\ref{sec:supp-alternatives}.

Unless stated otherwise, intervals are two sided 95\% intervals and resampling preserves the complete city or country cluster appropriate to the estimate. None of the development models identifies a causal effect of national development. Lower embedding dispersion means that AlphaEarth vectors form a tighter angular cloud; it does not by itself show that a physical city is less varied, that a downstream task must perform worse or that any particular training source caused the difference.

\section{Results}
\label{sec:results}

\subsection{Cities occupy a distinct part of AlphaEarth space}

\begin{figure*}[p]
    \centering
    \includegraphics[width=0.96\textwidth,keepaspectratio]{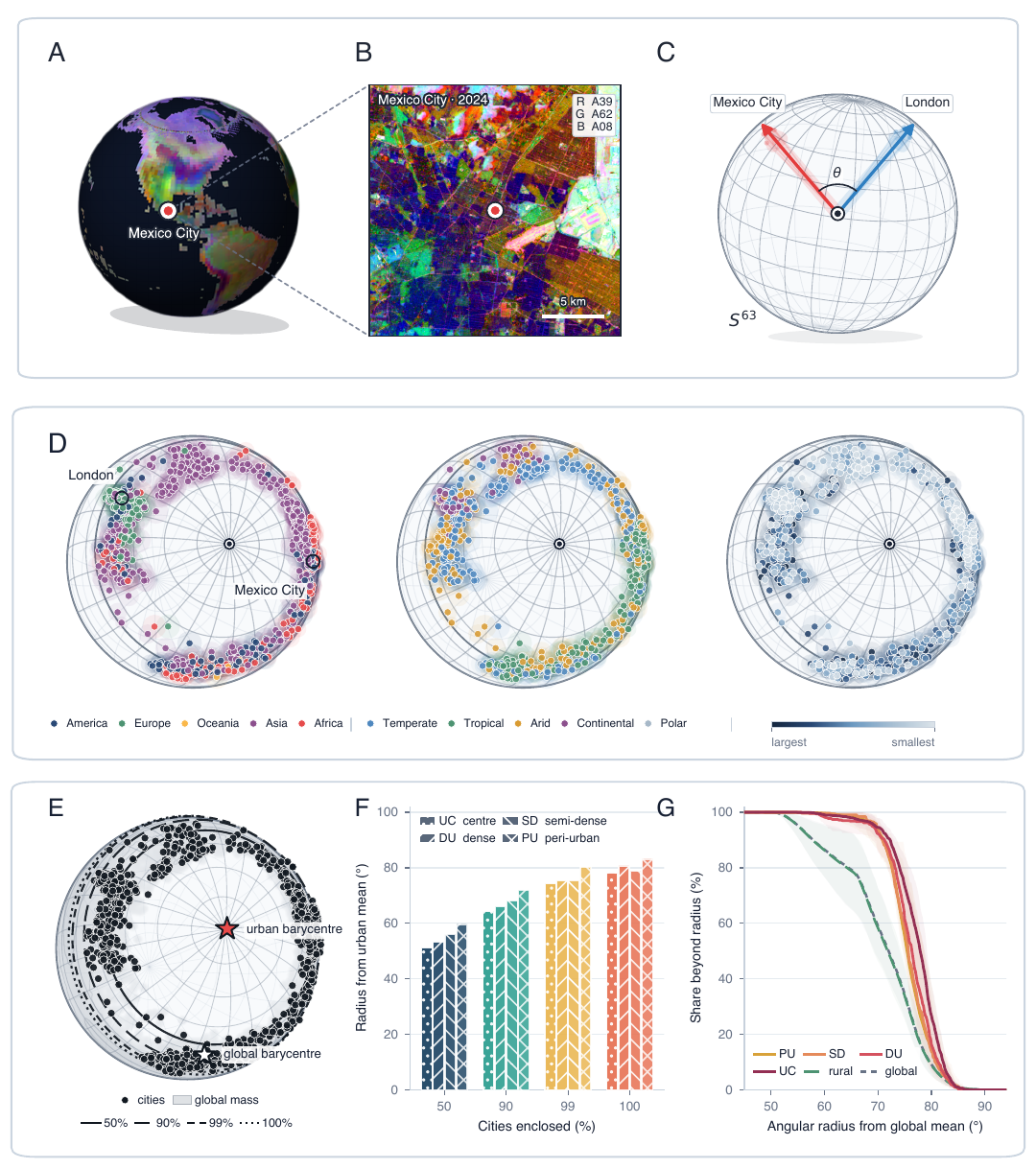}
    \caption{\textbf{Urban observations occupy a shifted but overlapping part of the global field.} Three released coordinates carry the global field into the texture of Mexico City, on separate colour stretches (A,B), and similarity is read as an angle on $S^{63}$---80.61\textdegree{} between Mexico City and London, larger meaning less alike (C). Seen from above the urban mean, city mean directions fill one restricted region, coloured in turn by continent, climate and population and grouped only imperfectly by each, halo area giving each city's internal dispersion (D). On the same view, half the cities lie within 54.69\textdegree{} of the urban mean and the farthest within 80.23\textdegree{}, against the 62.66\textdegree{} between urban and global means (E). Means of urban centres cluster most tightly and means of peri-urban areas most broadly (F), and the share of pixels beyond each angle from the global mean falls later for every degree of urbanisation than for rural and global samples (G). Radial distance is exact and azimuth schematic (Supplementary Sections~\ref{sec:supp-sampling} and \ref{sec:supp-geography}).}
    \label{fig:global-urban}
\end{figure*}

Urban observations occupy a distinct but overlapping part of AlphaEarth's global representation. AlphaEarth assigns each location a unit vector with 64 components, so a larger angle between two vectors means that their embeddings are less similar. We represent each city by the spherical barycentre of its sampled vectors, which we call its mean direction. Giving every city equal weight, the overall urban mean direction is separated from the mean direction of 247,565 globally sampled land locations by 62.66\textdegree{} (Fig.~\ref{fig:global-urban}). Half of the city mean directions lie within 54.69\textdegree{} of the overall urban mean, 90\% within 67.88\textdegree{} and 99\% within 76.81\textdegree{}; the most distant city is 80.23\textdegree{} away. The separation is also a scale marker rather than a distinctive position: measured against the same reference, our sample of global land lies between 28 and 76\textdegree{} from the global mean direction, and 92\% of that reference is rural land whose own barycentre sits 4.64\textdegree{} away (Supplementary Table~\ref{tab:si-global-class-distances}). 

The shift also appears at the pixel scale. Measured from the global mean direction, the median radius is 77.91\textdegree{} for pixels in urban centres, compared with 71.82\textdegree{} for rural pixels and 72.00\textdegree{} for the complete global sample. Further, urban pixels sit closer to the urban barycentre than global pixels sit to the global barycentre---median radii of 59.2\textdegree{} and 72.00\textdegree{}, respectively. The global sample spans radically different surface archetypes---deserts, rainforest, steppe---sitting tens of degrees apart, so their unit vectors cancel heavily when averaged: the mean resultant has length 0.33. Urban pixels are one broad kind of place; their vectors point in broadly similar directions, cancel less and leave a resultant of length 0.50. Urban and rural pixels also allocate variation differently across AlphaEarth's released coordinates: rural land is 97\% of the reference and as dispersed as the whole, with total variance 0.892 against 0.893, while urban pixels carry 0.758 and urban centres 0.703. Both spread that variation over about 19 effective dimensions, so what cities lose is amplitude rather than directions, about 12\% of the standard deviation along each; a single natural cover is lower-dimensional, tree cover at 11 and cropland at 13. The distributions nevertheless overlap across continents, climates, population densities and degrees of urbanisation (Supplementary Fig.~\ref{fig:si-sphere-context}). 

Cities are also individually identifiable in this frame, carrying a distinct signature. Using the 2024 release, splitting each city's urban pixels into two random halves and asking which of the 1,000 mean directions built from the first halves lies nearest the second half, the answer is the city itself 5,000 times in 5,000 trials across five repeats, against a chance rate of 0.1\%, with the half held out sitting a median 2.12\textdegree{} from its own mean direction and 13.01\textdegree{} from the nearest other city's. The identity is carried by the average rather than the pixel: a single 10\,m pixel returns to its own city 52.4\% of the time (48.1 to 56.8\%), and when it errs it errs locally, 54.4\% of misassigned pixels going to a city in the same country at a median distance of 323\,km (Supplementary Table~\ref{tab:si-city-identification}). A pixel that finds its own city among 1,000 half the time finds its own degree of urbanisation among four only a little more often than chance. We compute a mean direction for each class and ask whether each pixel sits closest to the mean of its own class. With the class means learned in other countries, 36.6\% of pixels do, in a balanced audit of 408 cities in 117 countries, against 25\% by chance (95\% interval 36.0 to 37.1\%). With the class means learned from the other half of the same city's pixels, 54.2\% of peri-urban and 56.7\% of pixels in urban centres sit closest to their own class, but only 39.2\% of semi-dense and 33.4\% of dense urban pixels do, across the 544 cities with equal samples in every class. The labels track an overlapping gradient, not four clean pixel classes (Supplementary Table~\ref{tab:si-prediction-benchmarks}).

Semantic audits show that broad land cover and climates are predictable from the representation (median AUROC 0.919 and 0.929), and many physical targets add information beyond degrees of urbanisation. The signal is distributed, however: a model confined to the leading directions of the representation predicts these targets worse than one using all 63, and keeping 90\% of its predictive score requires 32. The representation predicts elevation, vegetation, moisture, bare and built spectral indices, and radar roughness almost completely in countries excluded from fitting ($Q^2$ 0.83 to 0.97), building height, volume and night lights about half as well (0.39 to 0.50), and population density barely (0.09) (Supplementary Fig.~\ref{fig:si-urban-semantics}).

\subsection{Geography structures continuous differences between cities}

\begin{figure*}[p]
    \centering
    \includegraphics[width=\textwidth,keepaspectratio]{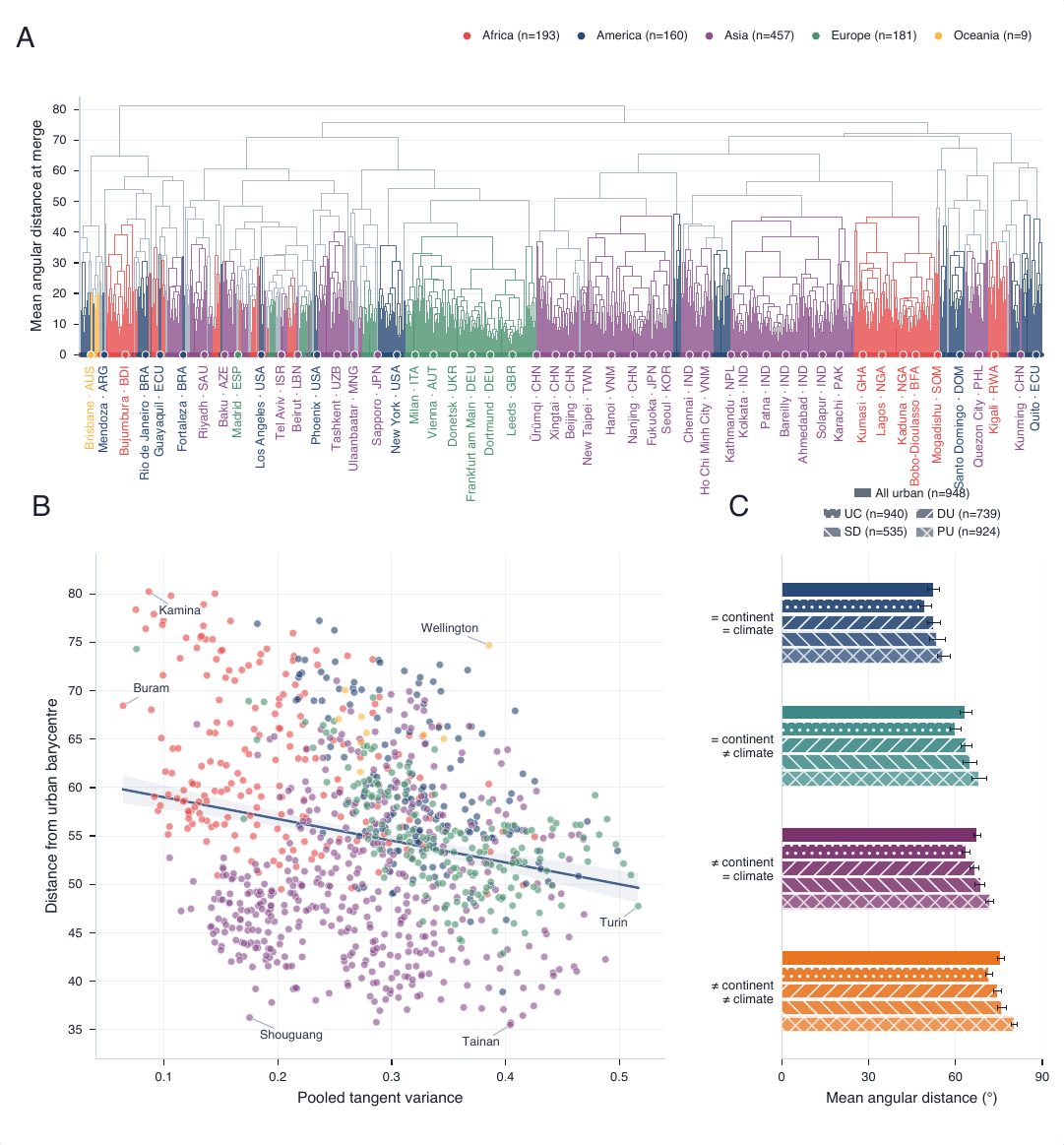}
    \caption{\textbf{City similarity is continuous but geographically structured.} The hierarchy over all 1,000 city mean directions descends in a smooth cascade rather than into separated groups: branch height is angular distance, continent colours the leaves and 50 cities are named, and the best cut manages a silhouette of only 0.266 at 29 groups against a cophenetic correlation of 0.793 (A). Each city's distance from the urban mean, against the variance inside it, shows that where a city sits in that continuum barely predicts how much variation it holds, and the weak decline drawn here is not robust to the catalogue's sampling design (B; Supplementary Section~\ref{sec:supp-geography}). Geographic context does order position. Bars give the mean angular distance from an anchor to its population-matched partner under each pairing of shared or different continent and climate: among 948 anchors in 160 countries, cities sharing both lie 30.8\% closer than cities sharing neither, and the same ordering returns in each of the four degrees of urbanisation (C).}
    \label{fig:city-structure}
\end{figure*}

Pairwise differences among city mean embeddings form a continuum rather than a small set of discrete global types. We represent each of the 1,000 urban areas by the mean direction of every degree of urbanisation available for that city, then cluster all 499,500 pairwise angular distances without using geographic labels. The resulting tree reproduces the original pairwise distances well (cophenetic correlation 0.793), but none of the tested divisions into 2 to 40 groups is well separated. The largest silhouette score is 0.266, at 29 groups, and the curve falls away on both sides of it; partitioning around medoids instead of cutting the tree gives the same reading, with a maximum of 0.264 at 17 groups. Those values nevertheless exceed what a structureless configuration matched to the same distance geometry produces---at 20 groups, 0.246 against a matched maximum of 0.021 over 200 draws---so differences among cities are strongly structured without separating into types. The structure is lumpy at short range: 2.6\% of pairs lie within 30\textdegree{} of each other where a structureless configuration yields 0.01\%, and the closest pairs sit within single national systems---Leeds and Sheffield at 3.43\textdegree{}, Rotterdam and Amsterdam at 4.68\textdegree{}, Mexico City and Puebla at 8.51\textdegree{}---while other close branches cross national and continental borders. Its one large division is continental: the leading axis of city positions is bimodal, with Europe forming a compact mass against the rest of the world and continent explaining 43.7\% of that axis (Fig.~\ref{fig:city-structure}A; Supplementary Fig.~\ref{fig:si-geography-dou}A).

Continent and climate account for a substantial part of this continuum. We match each city to cities of similar population---within 0.10 log units, and on no other variable---under four combinations of shared or different continent and climate, selecting matches without AlphaEarth distance or HDI. Among 948 anchors for which all four matches are available, cities sharing continent and climate are separated by 52.4\textdegree{} on average, compared with 75.6\textdegree{} when they share neither. They are therefore 30.8\% closer in embedding space (Fig.~\ref{fig:city-structure}C). The ordering repeats in each of the four degrees of urbanisation, where the reductions range from 29.4 to 30.9\%, and remains between 30.5 and 30.9\% as the population caliper changes. Matches within the anchor's own country supply 3.9 percentage points of the pooled reduction; excluding them leaves 26.9\%. In a separate test that excludes whole countries from fitting, across the 992 cities with observations in urban centres, continent and climate explain 24.3\% of variation among the mean directions of urban centres in the excluded countries (95\% interval 19.6 to 29.7\%); population alone explains essentially none (Supplementary Fig.~\ref{fig:si-geography-dou}B,C).

A city's average position and the amount of variation inside it are only weakly related. Across all 1,000 cities, dispersion declines with angular distance from the overall urban mean ($r=-0.222$), but an interval that resamples whole countries runs from $-0.529$ to $0.021$ and the relation is not stable under the catalogue's sampling design (Fig.~\ref{fig:city-structure}B; Supplementary Section~\ref{sec:supp-geography}). We therefore do not interpret centrality as a predictor of internal variation.

Mean position and the orientation of internal variation share local structure without defining one hierarchy. Across the same 1,000 cities, pairwise mean-direction distance and distance between transported top-ten covariance subspaces have Spearman $\rho=0.703$, and the two ten-nearest-neighbour sets overlap by 53.9\% on average. Agreement falls to $\rho=0.336$ between cophenetic distances, and only five exact clades of ten or more cities are shared by the two trees, the largest holding 26 (Supplementary Fig.~\ref{fig:si-mean-variance-hierarchies}). Geography thus links where a city lies in the representation to how it varies internally, but neither relation supplies a fixed global taxonomy.

\subsection{Most variation within cities remains inside broad urban classes}

\begin{figure*}[p]
    \centering
    \includegraphics[width=0.96\textwidth,keepaspectratio]{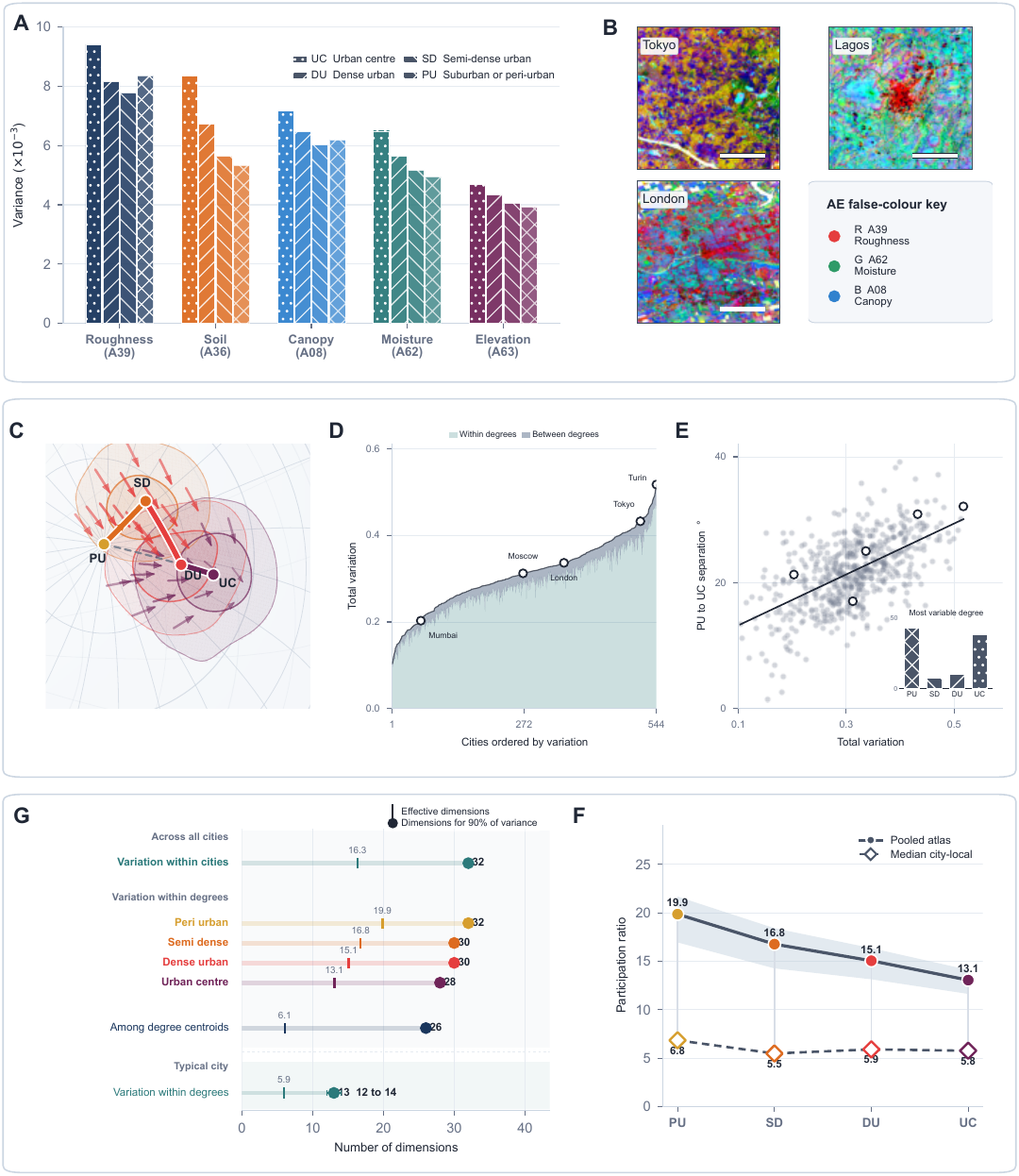}
    \caption{\textbf{Urban classes share one structure and leave most variation inside them.} The four degrees of urbanisation spread their variance across the same coordinates in nearly the same proportions (A), and false-colour views of three coordinates show the texture cities share (B). Centred on its own mean, each city becomes a short path from peri-urban land to urban centre; aligned, the paths share one ordered trajectory but differ in curvature and scale (C). Variation between classes (darker) and within them (lighter) shows the between share averaging 8.88\% and staying a sliver from Mumbai to Turin (D), and widening with total variation (E). Pooled across cities, the classes need 28 to 32 components for 90\% of variance, fewer toward the urban centre, while a single city stays near six (F); the count depends on which covariance is asked about, and the city-centred paths have few dominant directions but need 26 components against 13 for a typical city (G).}
    \label{fig:within-city-structure}
\end{figure*}

Broad urban classes reweight one shared profile of variance across coordinates rather than activate separate parts of the representation. In a balanced audit of 544 cities in 117 countries, correlations between the 64-coordinate within-city variance profiles of the four classes range from 0.874 to 0.983, while their effective coordinate counts range only from 57.1 to 59.2. The coordinate variances and false-colour views, which we label mnemonically with dominant correlates loading on each axis, make this shared structure visible (Fig.~\ref{fig:within-city-structure}A,B). The coordinates themselves are one basis among many, however: rotating the 64 axes would give a different profile of the same variation, so no single coordinate carries meaning of its own (Supplementary Fig.~\ref{fig:si-spherical-pca}G).

The four degrees of urbanisation account for less than one tenth of embedding variation observed at finer spatial scales inside cities. The largest exactly balanced sample contains 544 cities in 117 countries, with 250 observations in each class and 544,000 observations in total. Differences among the four class means account for 8.88\% of tangent variation within cities (8.32 to 9.37\%), while 91.12\% remains within classes. Correcting the class-mean component for finite-sample noise reduces its share to 8.60\%. The labels therefore supply a coarse ordering but leave most internal variation unresolved (Fig.~\ref{fig:within-city-structure}D).

Degrees of urbanisation are coarse. A finer schema recovers only a fifth of that residual: every one of the 544,000 observations also carries a WorldCover class, which works at 10\,m rather than 1\,km; six of its eleven classes occur in our urban sample and a typical city contains five. Splitting each city's variation by land covers instead of by degrees of urbanisation leaves 71.8\% within classes (70.6 to 73.2\%) rather than 91.1\%, and splitting by both schemas at once leaves 65.9\% (64.8 to 67.3\%). Most of the variation inside cities therefore lies within classes whichever schema is used, and what remains is not noise. Groups finer than degrees of urbanisation, clustered from the embeddings alone in 93 countries, carry meaning to 24 others: pixels from those unseen countries mapped onto the nearest cluster share population, vegetation and built form, measurements never used to make the groups (Supplementary Fig.~\ref{fig:si-geography-dou}E,F). The groups are only moderately stable, however (adjusted Rand index 0.50), so they show that finer types exist without fixing what they are.

A common urban gradient exists, but cities depart from it in different directions. We centre each city's four class means on its own mean, so every city becomes a short path from peri-urban to urban centre, and compare those paths across the 544 cities. The average path is nearly a straight line, one direction holding 97.5\% of the variation among its four points, yet that average accounts for only 31.9\% of the variation in the cities' own paths (29.1 to 35.9\%); the remaining 68.1\% is specific to each city and spreads over about twenty directions. Figure~\ref{fig:within-city-structure}C,E shows this combination: the average path is ordered, while its curvature, scale and surrounding variation differ among cities. The middle of the gradient carries the most movement and the most disagreement. In a typical city the step from semi-dense to dense urban land is the longest, a median 11.8\textdegree{} against about 10.1\textdegree{} for the steps on either side, and it is also the step at which cities depart farthest from the common one; the average path is nearly straight because those departures point in different directions and cancel.

Pooling cities expands the apparent dimensionality of the variation that remains within classes. In a sample of 544 cities that have all degrees of urbanisation, pooled participation ratio within classes declines from 19.86 in peri-urban areas to 13.06 in urban centres, while 28--32 principal components are still required to retain 90\% of variance. By contrast, across all 1,000 cities, a typical city has participation ratio 5.95 (interquartile range 4.85 to 6.87) and requires 13 components (12 to 14). Ten directions learned in some countries carry over to others: they capture 60.9\% of the variance in held-out countries, 94.2\% of what ten directions fitted to those countries themselves, the oracle on excluded cities in the figure, would capture. Single cities rotate further, their own ten leading directions sitting a median 33.0\textdegree{} from the shared set (Supplementary Fig.~\ref{fig:si-spherical-pca}). The pooled atlas spans more directions partly because local covariance directions rotate, not because every city uses the complete pooled basis.

Copies of the same footprints on non-urban land show how much of this belongs to cities. Cities carry more variation within their footprints than their copies on non-urban land, about 60\% more, and 40\% more than copies kept in the same climate. The number of directions that variation occupies runs the other way. One city at a time, cities spread their variation over more directions than the land they were copied to. Pooled, the copies spread over more directions than the cities, 26.8 against 21.0, and 23.6 against 18.3 under climate matching. A basis learned from other countries' cities also carries over to a new city 15 points better than one learned from their copies (Supplementary Fig.~\ref{fig:si-placebo-pca}).

\subsection{Dispersion within cities is associated with national development}

The amount of variation retained within each degree of urbanisation is geographically unequal. National Human Development Index (HDI) has a standard deviation of 0.141 in the primary sample of urban centres. After adjustment for population, land area and continent, an increase of this size is associated with 14.1\% greater dispersion within urban centres in 977 cities from 157 countries (95\% interval 6.5 to 22.3\%, or 6.5 to 23.3\% under a wild cluster bootstrap; Fig.~\ref{fig:development-temporal-dynamics}A). Two supporting dispersion outcomes give the same answer on the expanded catalogue. Across all available urban observations the association is 16.7\% in 985 cities from 157 countries (9.3 to 24.6\%); within the four equally sampled classes it is 18.3\% in 537 cities from 113 countries (11.5 to 25.5\%; Supplementary Table~\ref{tab:si-development-sensitivities}). HDI is national context rather than a treatment applied to individual cities, so all of these estimates are descriptive associations.

The association persists across annual layers, but it is sensitive to how geography is handled: removing African cities lowers it by about a third once both estimates are put on the same scale, from 9.8\% to 6.6\% per 0.1 of HDI, and finer geographic controls widen its interval to include zero (Supplementary Fig.~\ref{fig:si-development-robustness}D, E).

The association spans the representation rather than concentrating in a few directions. Across the 992 cities with a complete 250-pixel sample of their urban centres, the leading shared direction contains 17.4\% of internal variation, the first three contain 35.0\% and the first ten contain 62.6\%. In the 977 of those cities with HDI and context, one standard deviation higher HDI is associated with 14.3\% more absolute variation in the first direction, 14.0\% more in the first ten and 19.9\% more in the remaining 53---so the tail gains more than the leading span. Only one of 63 directions retains a change in variance share after adjustment for 63 comparisons, the fifty-eighth, carrying 0.09\% of pooled within-city variation and lying beyond the span where cross-fitted directions agree across country folds, so we do not read it as a direction. Lower dispersion in national contexts with lower HDI is therefore broadband in this coordinate system once geography is adjusted for, not the absence of one or two components.

\subsection{Vegetation, radar structure and built form account for the gradient in nearly equal parts}

Measured urban form removes a substantial part of the association and population removes none of it; detailed environmental structure removes most of what remains. On the fixed Figure~\ref{fig:development-temporal-dynamics}B sample of 942 cities in 155 countries, the baseline HDI association is 15.0\% (7.5 to 22.9\%). Adding population structure leaves 15.0\%, the built-form distribution 8.5\% and settlement vintage 7.2\%. Detailed climate composition leaves 7.1\%, vegetation level and spread 4.5\%, and surface structure observed by radar 1.4\% ($-3.8$ to 6.9\%). Because that final adjusted interval includes zero, 23.9\% of paired bootstrap draws place the implied attenuation ratio above 100\%, where it no longer reads as a fraction of the gradient removed; we therefore report the paired difference in the log coefficient instead: 0.125 log units (0.064 to 0.172), which excludes zero.

That ladder is one of 720 orderings, and the order it uses moves every block's apparent share. Averaging each block's contribution over every entry position gives an attribution that does not depend on order and sums exactly to the same total reduction of 0.125 log units (0.066 to 0.173), and it ranks the blocks differently from the drawn sequence (Table~\ref{tab:shapley-blocks}). Vegetation takes 33.1\% of the reduction, radar surface structure 29.6\% and built form 28.6\%; all three have intervals excluding zero on the coefficient scale, and no two of the three are separated by more than five percentage points. The other three fall far below and their intervals include zero: settlement vintage takes 10.2\%, detailed climate 1.1\% and population $-2.7$\%, a negative value that raises the HDI coefficient slightly rather than explaining it. One caveat attaches to every share in that table. The outcome is reduced over the whole urban centre while the physical, environmental and radar covariates are summaries over 250 sampled locations inside it, so measurement error in the regressors attenuates each block and the shares divide a reduction that is bounded below rather than estimated.

The ladder's ordering reflects entry position rather than explanatory content: built form enters third and collects 1.61 times its contribution averaged over orderings, vegetation enters last and collects 0.60 times its own, and each leading block's marginal contribution moves by a factor of at least 3.7 across orderings (Table~\ref{tab:shapley-blocks}). The ladder remains an accurate description of a sequence of adjustments and a poor attribution, so we treat the shares averaged over orderings as the primary summary and use the ladder only as the figure.

Whether a residual gradient remains after that adjustment depends on how measured surface structure is represented, and we report both specifications. The ladder enters raw Sentinel-1 VH backscatter summarised as a mean and standard deviation over every pixel in the urban centre, and on the 938 cities in 155 countries with complete coverage it leaves 1.4\% ($-3.8$ to 6.9\%), spanning zero. Under the optical--radar index Supplementary Section~\ref{sec:supp-alternatives} documents---winsorised NDVI and the dual-polarisation radar vegetation index, entered as a city median and interquartile range---the same rows leave 6.4\% (0.05 to 13.2\%), which excludes zero. The gap is 0.048 log units. Holding one difference fixed while the other varies isolates two causes, each with an interval excluding zero: 0.027 log units from the covariate, raw backscatter against the index, and 0.020 from the city summary statistic, a population mean over every pixel against a median over 250 sampled ones. The strength of the fully adjusted association therefore depends on the covariate set and is not settled by these data.

\begin{table*}[t]
    \centering
    \fontsize{8}{10}\selectfont
    \setlength{\tabcolsep}{4pt}
    \begin{tabular*}{\textwidth}{@{\extracolsep{\fill}}lcccc@{}}
        \toprule
        {\bf Block} & {\bf $\Delta\beta$ averaged over orderings} & {\bf Share of total} & {\bf Ladder's own step} & {\bf Range over orderings} \\
        & (log units, 95\% CI) & (\%, 95\% CI) & ($\div$ the average over orderings) & (log units) \\
        \midrule
        Vegetation & 0.0415 ($0.011$ to $0.078$) & 33.1 (11.9 to 56.8) & 0.0248 ($0.60\times$) & 0.016 to 0.061 \\
        Radar surface structure & 0.0372 ($0.012$ to $0.063$) & 29.6 (10.8 to 54.2) & 0.0298 ($0.80\times$) & 0.012 to 0.060 \\
        Built form & 0.0359 ($0.010$ to $0.055$) & 28.6 (11.1 to 46.4) & 0.0579 ($1.61\times$) & 0.013 to 0.073 \\
        Settlement vintage & 0.0128 ($-0.002$ to $0.025$) & 10.2 ($-2.3$ to 24.7) & 0.0117 ($0.91\times$) & 0.006 to 0.019 \\
        Climate zones & 0.0014 ($-0.011$ to $0.014$) & 1.1 ($-12.3$ to 9.7) & 0.0013 ($0.91\times$) & $-0.002$ to 0.004 \\
        Population & $-0.0033$ ($-0.014$ to $0.005$) & $-2.7$ ($-16.1$ to 4.0) & $-0.0001$ ($0.02\times$) & $-0.009$ to $-0.0001$ \\
        \midrule
        Total reduction & 0.1254 (0.066 to 0.173) & 100 & 0.1254 & --- \\
        \bottomrule
    \end{tabular*}
    \caption{\textbf{Vegetation, radar surface structure and built form account for the development gradient.} Each block's contribution is averaged over all 720 orderings on the same 942 cities in 155 countries that Figure~\ref{fig:development-temporal-dynamics}B draws, and the six sum exactly to the total reduction of 0.125 log units. The fourth column gives the step the drawn ladder assigns each block and the fifth the range that step takes over the other orderings: built form collects 1.61 times its contribution averaged over orderings by entering third, vegetation 0.60 times its own by entering last. Percentage points per standard deviation are not tabulated because $\exp(\beta)-1$ is not additive.}
    \label{tab:shapley-blocks}
\end{table*}

Broad cover proportions are not the explanation. Recomputing dispersion in repeated samples of exactly 30 built-up and 30 vegetated pixels per city, averaged over 40 repetitions, leaves 17.9\% (10.8 to 25.5\%) across 933 cities in 151 countries; restricting to built pixels alone leaves 11.7\% and to vegetated pixels alone 19.8\%. The environmental measurements are the strongest competing explanation observed here, and they are also the point at which the residual comes to depend on the specification, so they can neither establish nor rule out a disparity in the representation beneath them (Supplementary Fig.~\ref{fig:si-development-robustness}E).

Terrain is a second channel for dispersion, and it behaves differently. Adjusted dispersion rises 37\% for every tenfold increase in local relief within the centre (29 to 47\%), and in a joint model relief, vegetation contrast and radar contrast each add about 10\% per standard deviation with the others held fixed. Relief is nearly unrelated to national development, a rank correlation of 0.17, so it predicts which cities are diffuse without touching the gradient: added after the full ladder it moves the residual from 1.5\% on those rows to 3.3\% ($-1.7$ to 8.5\%) on the 937 cities the elevation model covers, and as a seventh block it takes 1\% of the reduction (Supplementary Fig.~\ref{fig:si-terrain-coherence}). The covariates that do remove the gradient are the entangled ones. The spread of vegetation and of radar backscatter within a centre correlate with HDI at 0.48 and 0.42, and climate family explains a quarter of the variance of the first and a quarter of the variance of HDI, so adjusting for them removes development together with the landscape that accompanies it (Supplementary Section~\ref{sec:supp-results-development}). Holding development fixed separates the two. With fixed effects for country, vegetation contrast, radar contrast and relief keep 86 to 92\% of their pooled coefficients, so landscape predicts dispersion inside countries almost as strongly as between them; entering only the country means of those three covariates takes the HDI coefficient from 14.9\% to 4.0\% ($-3.3$ to 11.9\%); and a tenfold increase in relief raises dispersion by 37.7\% in the lowest third of HDI and 41.8\% in the highest, an interaction of 1.4\% per standard deviation ($-4.7$ to 7.8\%), so the representation reads physical surface the same way at both ends of development (Supplementary Table~\ref{tab:si-within-country}).

Two further audits narrow the mechanism without settling it. The released locations of AlphaEarth's training samples are highly uneven, a quantity we call ``focus''---or, how densely a place is represented among the model's training targets: the median count of 1.28\,km training chips seeded by Wikipedia or GBIF records within 25\,km of a study city rises from 14 in the lowest HDI tertile to 381 in the highest (Supplementary Fig.~\ref{fig:si-development-robustness}A). Yet focus does not carry the gradient. Adding it, together with the density of ecoregion samples and an indicator for the land-cover target confined to the United States, after the environmental covariates leaves 8.5\% (3.4 to 13.8\%) in the 495 cities with complete coverage, and each alone leaves 11.9 to 14.3\% in the 516 cities for which the densities exist. In a separate subset of 214 cities, controlling for public Sentinel-1, Sentinel-2 and Landsat scene counts and clear observations moves a 19.6\% association to 18.6\%, and adding orbit geometry and the Sentinel-1B loss leaves 20.3\%; neither interval is narrow enough to call either a null, and neither the released sample locations nor public scene catalogues reconstruct what the model actually ingested (Supplementary Fig.~\ref{fig:si-development-robustness}A, E).

\subsection{Separation between cities contracts while city paths repeatedly reverse}

\begin{figure*}[p]
    \centering
    \includegraphics[width=0.92\textwidth,keepaspectratio]{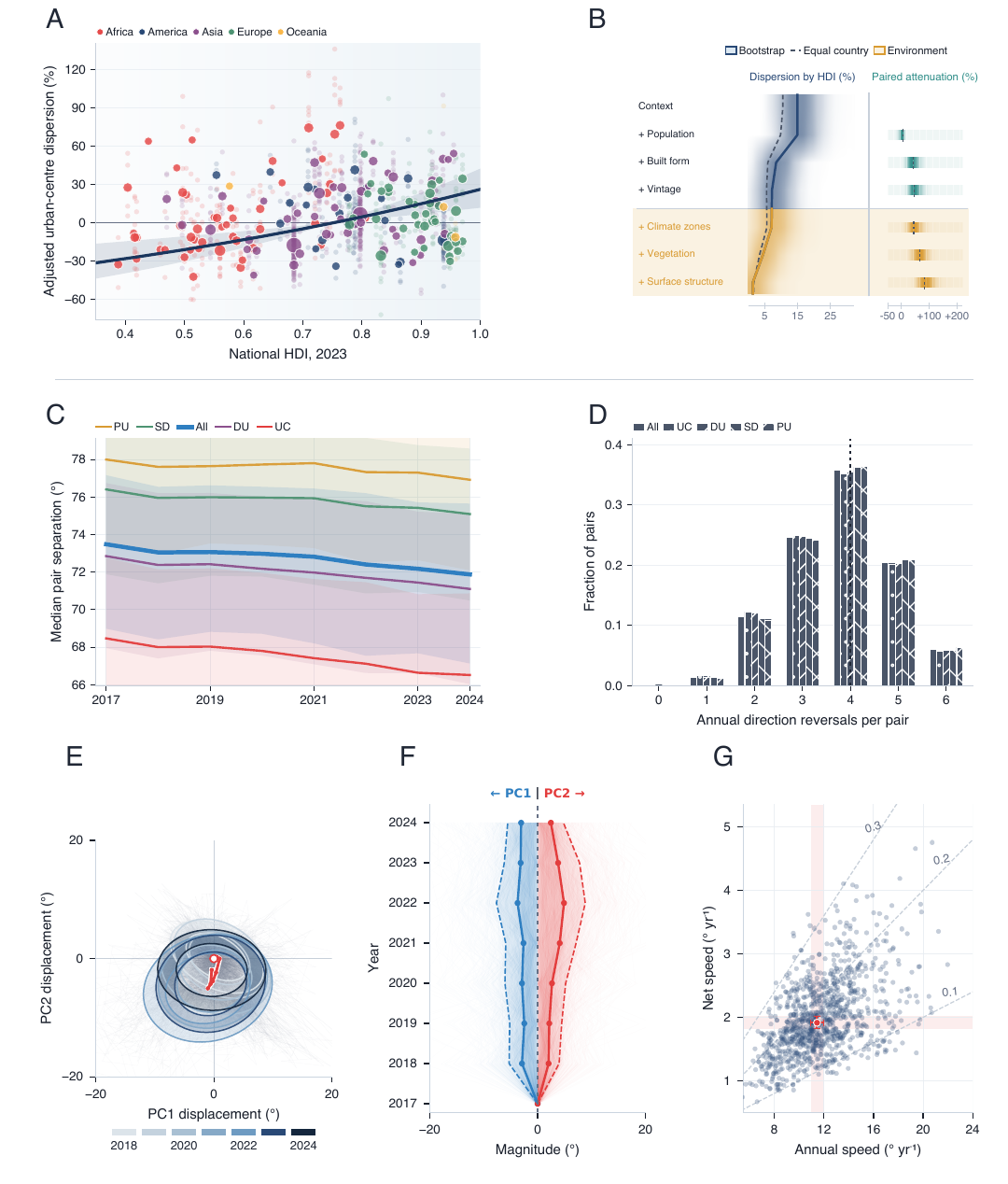}
    \caption{\textbf{Dispersion rises with national development, and between-city separation contracts while individual city paths keep reversing.} Each point is a city's adjusted dispersion within its urban centre against HDI (A), and adding covariate blocks walks the association from 15.0\% at context down to 1.4\% once radar surface enters, the equal-country fit dashed (B). Over time, median separation between the same 144,453 city pairs narrows in every degree of urbanisation, by 1.08\textdegree{} in peri-urban areas and 1.95\textdegree{} in urban centres (C), yet almost no pair follows it steadily: each pair's seven annual steps change sign a median four times, the exchangeable expectation (dashed line) (D). From each city's own 2017 origin, the yearly ellipses of displacement widen in no dominant direction and the mean path (red) barely moves (E,F), and yearly distance travelled against distance netted puts every city far below straight travel, the median (red) moving 11.45\textdegree{} a year and netting 1.91\textdegree{} (G).}
    \label{fig:development-temporal-dynamics}
\end{figure*}

Separation between cities declines modestly across the annual releases. This comparison is made within each degree of urbanisation and requires the same cities with all four degrees in every year, retaining 538 of the 1,000 starting cities in 117 countries and all 4,304 city-years. Across the same 144,453 unordered city pairs, median separation falls from 2017 to 2024 by 1.08\textdegree{} in peri-urban areas, 1.31\textdegree{} in semi-dense, 1.75\textdegree{} in dense urban and 1.95\textdegree{} in urban centres, and by 1.61\textdegree{} after pooling all urban observations within city (Fig.~\ref{fig:development-temporal-dynamics}C). Every interval for these changes from 2017 to 2024 excludes zero. The contraction is a tendency in the panel rather than a path each pair follows. Of the 144,453 pooled pairs, 85 (0.059\%) move closer at every one of the seven transitions and none moves farther at every transition (Fig.~\ref{fig:development-temporal-dynamics}D). Under the exact synchronous null, which reorders the eight calendar years for the whole field at once, 3.6 such pairs are expected, the 97.5th percentile is 15, and the observed calendar ranks first among all 40,320 orderings ($p=1/40{,}320$). That asymmetry---85 monotonically closing pairs against none monotonically opening, and 278 against five across the four degrees of urbanisation, each at or near the top of its own exact null---is the sharpest evidence of a shared direction here, because any exchangeable null makes the two counts equal in expectation. Pooling within each city does not require all four degrees. On all 1,000 cities and 499,500 pairs, median separation falls by 1.86\textdegree{} from 2017 to 2024 (1.43 to 2.22), the median pair converges 1.74\textdegree{} (1.45 to 1.97), 73.6\% of pairs end closer than they began (70.1 to 75.9\%), and 253 pairs close monotonically against four that open, where the synchronous null expects 12.4 with a 97.5th percentile of 51. The median pair reverses direction four times in its seven annual changes, exactly what a random ordering of the years would produce, so the shared contraction rides on year-to-year motion that has no memory. Starting in 2018 instead leaves a fall of 1.27\textdegree{} (0.90 to 1.60) and 769 closing against 46 opening pairs, still at $p=0.001$ under the exact null.

City paths show the same separation between annual movement and accumulated drift. This estimand does not compare urban classes and therefore retains the largest valid complete temporal sample: all 1,000 cities in 162 countries and all 8,000 city-years. After translating each city's mean direction over all its urban observations to its own 2017 origin, the first two shared displacement directions explain only 9.4\% and 7.8\% of change variance, and individual paths repeatedly change direction (Fig.~\ref{fig:development-temporal-dynamics}E,F). Directly on the unit sphere, the median city accumulates 11.45\textdegree{} of path per year (95\% interval 10.96 to 11.99) but has an endpoint net speed of only 1.91\textdegree{} per year (1.82 to 2.02). Median endpoint-over-path straightness is 0.17 (0.16 to 0.18; Fig.~\ref{fig:development-temporal-dynamics}G). The annual representations therefore move substantially but retain only a small share of that movement as straight endpoint displacement. A change to the urban fabric accumulates: a city that travels 11\textdegree{} a year and ends 2\textdegree{} a year from where it started has, for the most part, not been describing construction.

Part of that movement could be an artefact of sampling, because we draw pixels for each city at random each year: a city's mean direction could shift between years simply because different places inside it were sampled. We measure how much movement that alone produces by splitting a single year's sample into two halves, comparing their two mean directions and rescaling to the full sample size. Different places inside the same city and year move the mean by 1.57\textdegree{} for a whole city and 2.99\textdegree{} within one degree of urbanisation, against observed movement between adjacent years of 12.46\textdegree{} and 12.83\textdegree{}, so sampling accounts for under 2\% of the squared movement of whole cities (98.4\% lies beyond it, 98.3 to 98.5\%) and about 5\% within degrees (94.6\%, 93.9 to 95.0\%); the ordering holds in every transition and every degree, so the wobble is not an artefact of sampling. Two documented changes to the model between releases, a fix to a Sentinel-2 timecode fault and an end to sub-sampling frames at inference \citep{alphaearth2025}, could each move a layer without moving a city, and neither can be dated from the public record.

\subsection{Sentinel-1B loss changes local dispersion but leaves the global variance hierarchy intact}

One source of spurious movement can be isolated. Across all 1,000 cities in 162 countries, the 159 cities that lost a Sentinel-1 pass direction contract by 4.08 percentage points more than the 841 comparison cities from 2021 to 2022 (95\% interval $-5.88$ to $-2.50$; the exposed cities lie in 41 countries). No placebo transition reproduces that contraction, although the same group expands 2.36 points more than its comparison across 2023 to 2024 (0.78 to 4.01). The 46 cities, in 15 countries, containing a fixed point that received no Sentinel-1 scene at all in 2022 show a much larger difference of $-19.93$ points ($-25.81$ to $-10.87$), indicating that the response scales with how much observation was lost, as a sensor effect would and ground change would not. Its timing is not unique: the same group had already contracted 2.92 points more than its comparison across 2017 to 2018 ($-5.42$ to $-0.57$), about a seventh of the difference in the outage year, and no other placebo contrast in either arm excludes zero. Every degree of urbanisation moves the same way, by 2.87 to 5.65 points after direction loss and 13.53 to 23.02 points where a point lost every scene, with the largest contractions in dense urban areas and urban centres. Change adjusted for baseline has a modest continuous relation with mean Sentinel-1 loss (Spearman $\rho=-0.22$), with an interval that resamples whole countries running from $-0.307$ to $-0.086$ (Supplementary Fig.~\ref{fig:si-s1b-shock}).

The perturbation is locally measurable but does not define a stable global outage direction. Directions fitted in the discovery and holdout samples have cosine 0.162, and the held-out coefficient is not distinguishable from zero. Projecting the discovery direction from the variance decomposition of the 538 cities holding all four degrees removes 1.26\% of total trace variance and changes no share between cities, between degrees or within degrees by more than 0.07 percentage points under equal-city weighting, or 0.10 points under design weighting. The stability shows that the candidate nuisance direction does not organise the global variance hierarchy, so the cross-sectional comparisons survive the outage even though the time series does not; it does not rule out sensor contributions to the HDI association or reconstruct AlphaEarth's inputs.

\subsection{In the embeddings, construction appears but weather confounds}

Measured conditions predict part of the movement. Ridge models of monthly weather anomalies from ERA5-Land, radiation, precipitation, snowfall and snow cover, reduce the squared error in predicting a city's 64-dimensional annual change by 12.7\% (95\% interval 11.2 to 14.0\%) against year effects alone, scored on countries and years excluded from fitting across all 1,000 cities and 7,000 transitions. Seasonal vegetation from MODIS raises the reduction to 15.6\% and retrieval quality to 17.0\%, although the last increment spans zero ($-0.4$ to 3.2 points), and Sentinel-2 acquisition counts lower it to 16.4\% because measured opportunity does not transfer across years. Snow acts locally: physical snow-cover change alone reduces error by 12.0\% in the 379 cities whose peak climatological monthly cover reaches 1\% and by 0.04\% in the other 621, and both gains and losses of snow accompany larger movement. Removing one month's weather costs most in January in six of seven transitions. These are predictive scores on public proxies for what the model saw; they do not partition movement into causes, and the error they leave is not thereby identified as change on the ground (Supplementary Section~\ref{sec:supp-weather-wobbles}).

There is, however, evidence for real change in the data. At eleven construction projects in nine countries, from Paris's olympic village to Cairo's new capital, we follow the same 159,545 pixels through the annual layers and match each to a similar fixed location at least 2\,km away in its city. Over each project's own interval the median excess displacement beyond the matched controls is positive in every case, from 7.8\textdegree{} in Paris to 57.5\textdegree{} in Phoenix and 20.2\textdegree{} across the eleven, and the village mean moves 14.0\textdegree{} closer to Les Halles between 2019 and 2024 while the largest share of its pixels nearest any one Paris place falls from 34.2\% to 18.6\%. Real redevelopment therefore contributes to annual movement, but the controls are not certified unchanged and eleven selected projects cannot say what share of movement across 1,000 cities is physical change (Supplementary Section~\ref{sec:supp-construction}).

\section{Discussion}

Earth embeddings promise one geometry in which every city can be compared, and AlphaEarth delivers that geometry without delivering a neutral comparison. For cities, a representation works when it preserves relevant local distinctions, transfers under geographic ablation, allocates comparable capacity to the places being compared, and changes when the city changes rather than when sampling or sensors change. AlphaEarth meets the first two, fails the third by a margin we can measure but cannot attribute, and leaves the fourth uncalibrated despite evidence of excess movement at selected construction sites. Distinguishing cities and compressing urban variation are compatible claims: the first rests on the margins between precisely estimated city means, the second on the amplitude urban pixels carry relative to land as a whole. These results support comparative urban analysis across regions, but they do not make the representation self-interpreting, geographically neutral or temporally faithful.

Planetary coverage does not impose a fixed urban taxonomy, and it does not remove the need for one. City mean directions vary continuously, and no division of the 1,000-city tree cuts cleanly, although the differences among cities are far from structureless. Local climate zones and spatial signatures declare their variables and types in advance \citep{stewart2012local,arribasbel2022spatial}; an embedding offers a common metric before an urban target is chosen. That flexibility is also a constraint. Continent and climate explain a quarter of the variation among city means in countries excluded from fitting, and cities matched on population alone sit 30.8\% closer when they share both. A neighbour in embedding space is therefore a neighbour in a joint physical and observational representation, not necessarily a functional urban analogue, and the analyst must state whether geographic context is part of the intended similarity or a nuisance to be controlled.

The coarse schema leaves most of the variation the field retains unresolved, and much of that residual is structured rather than noise. Subdivisions learned in some countries transfer to others and separate independently measured vegetation, population and built form, so an embedding can distinguish environments that a coarse urban schema treats as equivalent, which is the argument for embeddings over typologies. Finer land cover reclaims about a fifth of what the degree of urbanisation leaves. Cities are individually richer than the land around them and collectively simpler than the same footprints extracted elsewhere, varying along directions they share; at the same time the shared urban gradient explains under a third of the variation in cities' paths through it, and each city's leading directions sit a median 33\textdegree{} from the pooled set. There is a shared direction that generally captures ``urban'', but any given city's own such direction aligns with it only partly. Further, split a city's pixels in half at random and each half is closer to the other than to any half of any other city, in 5,000 of 5,000 draws. AlphaEarth thus distinguishes cities from rural and wild areas, resolves not just an urban signature but a gradient from less to more urban, and marks that gradient in each city with properties unique to it.

How much variation is represented depends on where the city is. Cities in national contexts with lower HDI have less measured embedding dispersion. If two cities are described with different amounts of retained variation, a distance, a cluster boundary or a nearest neighbour does not mean the same thing in each, and a classifier or similarity search has less to learn from in some contexts even when every location carries the same 64 components. Embedding dispersion is variation retained by AlphaEarth, and measured environment is the strongest competing explanation: vegetation, radar surface and built form carry nine tenths of what adjustment removes, in nearly equal parts, and population none. Those covariates are themselves patterned by development and by climate, but our evidence suggests that landscape, not development, is to blame: vegetation, radar and terrain contrast predict dispersion as strongly inside countries, where development is fixed, as between them.

This clarifies what successful urban use looks like. AlphaEarth is well suited to exploratory comparison, retrieval, stratification and feature construction, and it can support supervised models when the target is externally defined and validation excludes the geographies to which the model will be transferred. It is less defensible as an off-the-shelf measure of form, function, development or inequality. Foundation representations reduce the cost of producing features. They do not remove the need to define the construct, calibrate it against external observations or evaluate it in the places where conclusions will be drawn.

Temporal use is potentially more fraught. Annual city paths accumulate far more movement than they retain as displacement, and movement that doubles back year after year is not the signature of a changing urban fabric, which accumulates. Three terms of that movement are now measured, none of them a change to the urban fabric. Redrawing a city's own pixels within a year produces about 13\% of it. Public records of weather, vegetation and retrieval quality reduce the error in predicting it by 17.0\% in countries and years excluded from fitting, with snow the sharpest local term, and the error they leave is not thereby identified as change on the ground. The loss of Sentinel-1B shows that the observational term is live: AlphaEarth was built to absorb exactly this loss \citep{alphaearth2025}, yet cities that lost a pass direction contracted measurably more than those that kept both, while the perturbation defines no stable global direction and leaves the cross-sectional hierarchy intact. The satellite record is shared by every field built from it, so an outage that reaches one embedding through a disclosed invariance can reach another through an undisclosed one, and from 2022 onward a city observed from one pass direction and a city observed from two differ in dispersion for reasons of observing alone, by about 4 points at the outage boundary, so exposure belongs in any comparison that spans them. Change on the ground is visible where it is documented: at eleven construction projects the same pixels move farther than matched locations in their cities, although those controls are not certified unchanged. Validating change in general still requires a temporal noise floor, the displacement expected when the underlying place is unchanged, and the sampling, weather and sensor terms bound only parts of it. Until that floor is calibrated, comparison over time remains unvalidated, while several causes of movement that are not change have been demonstrated.

Several limitations bound the interpretation. The city sample excludes smaller settlements and does not weight observations by area or population, and the catalogue mixes two sampling designs, which we treat as a live sensitivity throughout (Supplementary Section~\ref{sec:supp-geography}). The fixed 2020 degrees of urbanisation are coarse labels attached to much finer observations, and are delineated from census inputs whose quality tracks development, a bias whose direction we leave unresolved. Representation geometry can show where information is concentrated and how fitted structure transfers, but cannot by itself determine whether the retained distinctions correspond to the concepts urban researchers care about.

None of this is unique to AlphaEarth. Any field trained for planetary coverage spreads a bounded representation across every surface on Earth \citep{danish2026terrafm}, and the properties found here---unequal capacity, rotating local directions and unvalidated annual differences---are the ones to test for in the next one; the estimands are defined on the unit sphere and apply unchanged to any embedding carrying the same norm constraint. The representation expands what can be compared across cities, while leaving the central inferential work, semantic validation, geographic calibration and temporal reliability, to the analyst.

\section{Data \& Code Availability}
The sampled atlas is deposited
\href{https://github.com/asrenninger/alpha-urban}{here} and an interactive exploration of our results is \href{https://asrenninger.github.io/alpha-urban}{here}.

\bibliography{references}
\label{LastMainMatterPage}

\clearpage
\input{si}

\end{document}

%% file: si.tex
\beginsupplement
\section*{Supplementary Information}
\label{sec:supplement}

\subsection{Sampling and data processing}
\label{sec:supp-sampling}

To compare cities under one definition, we take them from the GHS--OECD Functional Urban Areas catalogue, which defines a city as a dense core with its commuting zone. We select 1,000 areas with at least 250,000 residents in 2015, spanning 162 countries on five continents, and assign the locations inside them to the four degrees of urbanisation recorded for 2020: suburban or peri-urban (21), semi-dense (22), dense (23) and urban centre (30). These labels are drawn at 1\,km and remain fixed across years, and sampling them at 30\,m does not raise their resolution.

To represent each city across its whole gradient of settlement density rather than in proportion to the ground each class covers, we stratify by degree of urbanisation. From the 2024 AlphaEarth field in Google Earth Engine \citep{google2025embeddings} we take up to 250 valid land observations per city and class. The released values are already unit-length by construction \citep{alphaearth2025}, so we renormalise only negligible drift. The expanded catalogue of 1,000 cities yields 840,776 observations in 162 countries: 553 cities have all four degrees of urbanisation and 544 reach exactly 250 observations in each, giving 544,000 observations for the quantities that require equal samples in every degree. An earlier sample covering 530 of those cities, used for the sensitivity checks below, holds 446,761 observations. Each analysis begins from the largest sample its inputs allow; smaller subsamples are named where they are used. One earlier table stores each axis under a signed-square transform and would need inverting before use; we read nothing from it, and every table we do read already holds unit vectors.

Because the loss of Sentinel-1B has to be seen both at fixed locations and in cities, its analyses run on two independent frames. What was acquired, and what a replay within the same year shows, are measured at a global grid of fixed points, of which 4,374 are eligible; movement, dispersion and severity within cities use all 1,000 catalogue cities in 162 countries with complete eight-year pooled mean directions, giving 8,000 city-years. Analyses by degree keep every series that can be estimated rather than requiring all four degrees to be present, and city and degree means are normalised resultants, the average of the unit vectors rescaled to unit length. Every city carries equal weight, so comparisons among cities describe the stratified sample, not the area or population of the full footprint.

To place urban observations against the rest of the planet, we draw an independent global reference. It begins with 40,000 candidates in each of 24 equal-area cells, defined by longitude and sine of latitude, of which 307,273 return an embedding and 247,565 of those carry an ESA WorldCover land label. One rule removes 28,139 locations of permanent water, 23,262 south of 60\textdegree{}S that WorldCover does not label and 8,307 unlabelled along the ocean fringe; 23 of the 24 cells retain land. Because candidate density is constant, the retained observations carry equal weight, and the degree of urbanisation here describes the sample rather than selecting it. Every reference quantity counts each location once, though the deposited table lists them under two sampling designs.

Because differences among cities may follow climate, size or national wealth, we attach external fields to each city. We reduce the 30 Köppen--Geiger classes to Tropical, Arid, Temperate, Continental or Cold and Polar families and give each city its predominant family; population and land area come from the Functional Urban Area catalogue; and national 2023 HDI, missing for 15 cities, cannot identify differences within a country. The primary model of development within urban centres therefore holds 977 cities in 157 countries, and the two supporting dispersion outcomes retain 985 cities in 157 countries when all available degrees are used and 537 in 113 when four are required. Physical covariates are read at the 250 sampled locations in urban centres: 2020 GHS population at 100\,m and static WSF3D building fields at about 90\,m. The primary outcome is not read there. It is computed over every valid cell of the urban centres in the Functional Urban Area, at a coarser implied spacing, so outcome and covariates are measured on different footprints and the adjustment carries classical measurement error.

\subsection{Spherical geometry}
\label{sec:supp-geometry}

Because AlphaEarth places every observation on the surface of a sphere, each comparison we make has to be an angular one. The model supplies a vector $z_i$ with 64 components for every observation, already of unit length by construction \citep{alphaearth2025}, and we remove only numerical drift from that length,

\begin{equation}
x_i=\frac{z_i}{\lVert z_i\rVert_2}, \qquad x_i\in S^{63}.
\end{equation}

The separation between two observations is then the shortest angle between their directions,

\begin{equation}
d(x,y)=\arccos\!\left[\operatorname{clip}(x^{\mathsf T}y,-1,1)\right].
\end{equation}

To summarise a whole set of directions with a single one, we take its normalised resultant, which we call its barycentre,

\begin{equation}
\mu=\frac{\sum_i w_i x_i}{\left\lVert\sum_i w_i x_i\right\rVert_2}.
\end{equation}

This is an extrinsic directional mean, not the intrinsic mean that directly minimises squared angular distance. Urban barycentres give one equal weight to each city and the global reference one equal weight to each observation.

A barycentre fixes where a city sits but says nothing about how its observations vary around that position. To express that variation while retaining spherical distance, we map observations to the tangent plane that touches the sphere at $\mu$. If $\theta=d(\mu,x)$, the spherical logarithm is

\begin{equation}
\log_{\mu}(x)=\frac{\theta}{\sin\theta}
\left(x-\cos\theta\,\mu\right).
\end{equation}

The length of this tangent vector equals the original angular distance, so dispersion within city $c$ follows directly,

\begin{equation}
D_c=\frac{1}{n_c}\sum_i d(x_{ci},\mu_c)^2.
\end{equation}

Cities have different mean directions, so their tangent planes differ too. Where tangent vectors from different cities must be compared, we move them to a common tangent plane by parallel transport along the shortest spherical path, which preserves vector lengths and mutual angles. This allows us to compare directions of variation without treating the released AlphaEarth coordinates as a flat Euclidean field. Because every vector has unit norm, the total second moment per observation is fixed, so dispersion is a share of a bounded budget rather than an unbounded quantity.

\subsection{Semantic, geographic and urban comparisons}
\label{sec:supp-geography}

To place cities against the planet as a whole we measure the angle between the urban barycentre, which weights every city equally, and the global barycentre, which weights every observation equally; city radii are measured from the urban barycentre and the share of pixels lying beyond each angle from the global one, so the two summaries are not one projection. The global reference is itself a sample of the planet, so the angle could depend on which cells happened to be drawn: we resample the 23 equal-area cells 400 times with the reference directions held fixed, and the related urban--rural comparison uses the same global reference and 1,000 bootstraps that resample whole cells. Those cells are badly unbalanced---only 21 of 23 hold pixels from urban centres, the largest holds 22.6\% of them, and Kish's effective cluster count, the number of equally sized cells the design is worth, is 8.97---and holding the references fixed removes their sampling variation as well. The intervals therefore cover less than their nominal 95\%, and we read them as a lower bound on uncertainty rather than as calibrated intervals.

\paragraph{Semantic transfer.}
To test whether the directions carry meaning outside the places used to fit them, we predict 18 predefined WorldCover, Köppen--Geiger and GHSL contrasts from directions fitted with one equal-area cell left out at a time, two of the contrasts independent of the urban labels and GHSL settlement only a positive control. Isotropic ridge models are fitted in the tangent plane of each fold over five folds that keep every country whole, and scored by $Q^2$, the reduction in squared error beyond a model that uses the degree of urbanisation alone; fits from the first sample are then applied without refitting to the cities added later. Because apparent capacity could belong to a city's context rather than to the embedding, a stricter audit compares a ridge with a fixed penalty and a reduced-rank regression of rank one against conventional covariates in folds grouped by country, judged by bootstraps over countries and cities and by permutations of whole cities. Refitting its pooled tangent reference on training cities alone moves $Q^2$ by at most 0.002. The available targets describe surface, environment, built form and intensity, not human function.

\paragraph{City context and hierarchy.}
Because a city's place in the urban field might simply predict how varied it is inside, we relate the two across all 1,000 city samples, with a bootstrap over cities, a sensitivity check on the 544 cities with equal samples in every degree and, for the pooled correlation, an interval that resamples whole countries. The catalogue mixes a core chosen on purpose with two later extensions drawn at random, so we report the estimate weighted equally, weighted by the design and computed inside each stage of collection, rather than the pooled figure alone. The hierarchy applies average linkage to all 499,500 pairwise angles among the 1,000 means without using labels. To separate geography from size, each anchor is then matched to the closest city in $\log_{10}$ population under the four combinations of shared continent and climate; all four matches must fall inside a caliper of 0.10 log units, the largest difference in population a match may carry, and neither AlphaEarth distance nor HDI enters the selection. Intervals resample anchor countries within continents with the matched partner held fixed. Whether that structure transfers to new national settings is tested with multivariate tangent models carrying continent and climate, compared with population quintile, over ten repeated fivefold splits that keep every country whole; negative $Q^2$ values in excluded countries are kept rather than truncated.

\paragraph{Identifying cities.}
Whether one city can be confused with another is a question about identity rather than separation on average, so we put it to a classifier. For each of the 1,000 cities we split its 2024 urban pixels into two random halves and compute a mean direction from each half. Every city's mean direction from the second half, and then each of 50 single pixels drawn from that half, is assigned to the nearest of the 1,000 mean directions built from the first halves, so that no pixel is scored against a mean it helped to form. The split is repeated five times with fresh seeds, and we record the share of assignments that return the correct city among the nearest one, five and twenty; intervals resample countries. Where an assignment is wrong we record whether the chosen city shares the country, continent and climate family of the true one and how far away it lies, against the exact expectation under a uniform draw from the other 999 cities; because the assignment of whole cities made no errors, the same geography is reported for the nearest wrong city instead. To set the scale below which two cities cannot be told apart, we bootstrap each city's mean direction 200 times from 250 of its own pixels, the cap per degree in the sampling design, and again at the city's own pooled count, and count the pairs of cities separated by less than twice the median angular standard error. The design, seeds and decision rule were registered before any accuracy was computed; the geography of the nearest wrong city and the curve of accuracy against pixels per city were added afterwards to qualify a ceiling result.

\paragraph{Predicting degrees of urbanisation.}
To ask whether the four degrees are recoverable from the embedding at all, nearest-centroid models fitted with whole countries excluded compare vectors in their absolute position with tangent vectors centred on each city and transported to a common plane. A second check works inside a single city, fitting centroids on one random half of its observations and scoring them on the other, so that a city's own context cannot help or hinder the comparison; it uses the subsample with equal samples in every degree. Continuous targets need no balance across the four degrees and use all 1,000 cities, where a ridge model fitted in folds that keep every country whole predicts each cell mean from the degree of urbanisation and the 63 tangent coordinates, against fits within the same fold that use the degree alone. Supplementary Table~\ref{tab:si-prediction-benchmarks} carries the samples, the metrics and the intervals from resampling countries and then cities.

The variance partition works inside the tangent plane of each of the 544 cities with equal samples in every degree, separating displacement among the four degree means from displacement within a degree. Cities in the same country are not independent, so the primary intervals come from 2,000 resamples that draw countries and then cities inside them; resampling cities alone would give an interval 1.75 times narrower. Because the variation left within a degree could be noise rather than structure that reappears elsewhere, we learn subdivisions in one set of countries and score them in another: 389 discovery cities in 93 countries against 155 confirmation cities in 24. Principal components are fitted on the discovery cities alone and retain 20 directions, four k-means clusters per degree give a 16-group refinement that is illustrative rather than selected, and scoring in the confirmation countries uses the complete 63-dimensional space. Stability is measured with the adjusted Rand index, which scores how far two partitions of the same points agree, between subdivisions learned on split halves of the discovery cities.

\paragraph{Finer schemas.}
The share of variation left within a class belongs to one schema at one scale, and a finer schema is already in the sample, so we can ask how much of that share is the representation and how much is the coarseness of four degrees. Every one of the 544,000 observations with equal samples in every degree carries an ESA WorldCover label---eleven classes at 10\,m, six of them occurring in the urban sample and a median of five inside a single city---so the identical partition can be run against it, changing only the number of classes. WorldCover leaves 71.81\% of the tangent variation inside a city within its classes (70.64 to 73.23\% when whole countries are resampled; 72.14\% after correction for noise) against the 91.12\% the four degrees leave, and the two schemas crossed leave 65.89\% (64.77 to 67.26\%). A class holding only a handful of a city's pixels could carry a share by chance, so we pool every class with fewer than ten of them; that moves the WorldCover share by 0.03 percentage points. A finer schema, motivated independently of this work, therefore recovers about a fifth of what the degrees of urbanisation leave unresolved.

Local climate zones would ask a different question---100\,m and seventeen classes, of which ten are built, sorting surface structure rather than cover \citep{stewart2012local,demuzere2022lcz}---and that question remains untested here. Coverage is not the obstacle, the global map being continuous over land. Label quality is uneven instead, with overall accuracy 74.5\% $\pm$ 15.1\%, the separation of built from natural surfaces near 95\% and F1 by class between 50\% for compact high-rise and 78\% for open low-rise. The direction of that error is not fixed in advance: misclassification unrelated to the embedding inflates the variance that appears within a class and would favour our claim, but error that tracks the embedding, as a schema built from surface structure might produce, could move the share either way. The binding obstacle is our own sample, drawn without keeping the coordinates of each cell, so the 544,000 pixels cannot be joined to any external map. The one table that does carry coordinates holds 1,160 points in urban centres over 21 equal-area cells, the largest holding 35.9\%, and supports neither a partition inside a city nor an interval that resamples whole countries. Settling the question requires the same 2024 stratified sample drawn again, unchanged except that each row keeps its coordinates; the deposit specifies that collection in full, with its design requirement and its fallback.

\paragraph{Spherical covariance and dimension.}
To ask how many directions the representation actually uses, we read the spectrum of each covariance, the variances along its principal directions taken in order. Spectra over a whole city and covariances inside a single city take all 1,000 cities, while the quantities needing exactly 250 observations in every degree---spectra within each degree, the path a city traces through the four degrees, transfer between countries and the rank a classification task needs---take the 544-city subsample. Pooled spectra use one declared common tangent plane, and covariances from single cities are transported to a common reference. For eigenvalues $\lambda_j$ the participation ratio, which counts how many directions carry the variance in effect, is $(\sum_j\lambda_j)^2/\sum_j\lambda_j^2$, and variance rank is the smallest $k$ reaching 90\%. Neither is the rank a task needs.

Because a basis scored where it was fitted flatters itself, five folds that keep every country whole fit each covariance reference and each basis of rank $k$ on the training countries alone, and what they capture in the excluded countries is compared with the full 63-dimensional covariance and with an oracle of the same rank fitted to the excluded fold itself, which is optimistic by construction. Overlap of projectors and principal angles compare subspaces without assigning signs to axes that are nearly tied. The probe of the rank a task needs instead leaves one continent out. At rank ten we set an oracle fitted inside the city itself against bases shared from the training countries, weighted by city or by pixel, against one shared tangent centre, against covariance measured along straight lines and against ten of the released coordinates. Finally, 500 random orthogonal bases show how much variance the best single coordinate can appear to carry; the leading eigenvalue does not change under those rotations.

The common path through the four degrees is aligned across cities before display by generalised Procrustes analysis, which turns each city's path onto a shared orientation. That alignment is fitted in $O(2)$ rather than $SO(2)$, so it can reflect a path as well as rotate it, and it reflects 14 of the 544 cities. It is a display convention and no reported quantity depends on it.

\paragraph{Placebo footprints.}
To learn how much of the pooling result belongs to cities rather than to any landscape of the same size and shape, we copy each of 504 Functional Urban Area polygons to non-urban land. Each polygon is moved without rotation, in an equal-area projection, to a destination outside every catalogued urban area that keeps its coastal, river or inland setting, with 28 pairs in each of six destination continents and three settings. We draw 2,048 distinct 30\,m grid cells uniformly inside the source polygon and read the 2024 field, land cover, settlement class and elevation at those cells and at their translated positions; water is excluded by land-cover class, and drawing continues until each footprint holds 2,048 land cells or has requested four times that number. Covariances inside a footprint, transport to a fixed global reference, folds that exclude whole countries and a bootstrap that resamples the cities' countries and the copies' countries independently then follow the city analysis, the two training bases in a fold excluding the receiving countries being scored and sharing the same training pairs. Settlement labels are copied from each city to its translated cells for the partition comparison only and never enter the principal components.

\subsection{National development models}
\label{sec:supp-development}

The principal model is

\begin{equation}
\log D_c=\alpha+\beta H_c+\gamma P_c+\delta A_c
+\eta I_c+\lambda_{k[c]}+\varepsilon_c,
\end{equation}

where $H_c$ is national HDI, $P_c$ and $A_c$ are $\log_{10}$ population and land area, $I_c$ identifies the stage at which the city entered the sample and $\lambda_{k[c]}$ represents continent. We centre continuous predictors and divide them by their standard deviation within each outcome sample, then report $100[\exp(\beta)-1]$, the percentage difference associated with one standard deviation higher HDI. Because the standard deviation of HDI ranges from 0.110 to 0.168 across the subsamples we compare, coefficients from different subsamples are contrasts on different scales and cannot be read against one another. The three outcomes are dispersion across all available urban observations, dispersion over every valid cell of the urban centres and mean variation within the four classes sampled equally.

The second is the primary outcome, and it covers a whole footprint rather than a sample of it: every valid cell of the urban centres the Functional Urban Area contains enters, at an implied spacing near 95\,m rather than the 30\,m at which the pixel samples are drawn. A larger footprint contributes more cells, so the outcome could track the size of the urban centre rather than development; adding the logarithm of the cell count leaves the coefficient at 14.1\% (6.7 to 22.0\%), so it does not. The same quantity recomputed on the 250 sampled urban-centre pixels gives 16.0\% (9.0 to 23.5\%).

Because HDI varies at country level, standard errors and confidence intervals allow arbitrary dependence among cities in the same country and use a small-sample correction with a $t$ reference distribution. A coefficient fitted and read on the same countries can describe those countries alone, so ten repeated fivefold splits test prediction in countries excluded from fitting. We also fit within each stage of collection and each continent, omit each country in turn and screen the five continents with the Benjamini--Hochberg procedure, which controls the false discovery rate, treating the three related dispersion outcomes as one planned family of supporting definitions rather than independent discoveries.

That correction is set by the number of countries, which the effective cluster count says overstates what the design is worth, so we do not rely on it alone. We refit the principal coefficient under a wild cluster bootstrap: the null is imposed, each replicate multiplies a whole country's restricted residuals by one random sign, and the interval is the set of coefficients that 9,999 draws do not reject at 5\%. Its $p$-value against zero is 0.0003, and Table~\ref{tab:si-development-sensitivities} sets that interval beside the one clustered on countries and a cluster jackknife. Prediction in excluded countries could also flatter itself if the folds were fixed while the data were resampled, so each of its 2,000 replicates draws countries within continent and refits the whole plan of ten splits by five folds before scoring.

Where an adjustment ladder is reported---blocks of covariates entering in a fixed order, one after another---the primary summary is the paired difference in the log coefficient rather than the attenuation ratio, the share of the base coefficient that adjustment removes. The ratio's denominator never approaches zero here, but its numerator grows once the adjusted coefficient can change sign. At the final rung of the fixed ladder, 23.9\% of paired bootstrap draws put the ratio above 100\%, where it no longer reads as a fraction of the gradient removed.

A ladder also assigns each block whatever the blocks before it have left, so a block that enters early can take credit that a later block would have taken; we therefore decompose the same reduction over all orderings of its six blocks. Context stays in every model rather than forming a block of its own, so the six players are the ladder's own six increments. We fit all 64 subset models on identical rows and average each block's marginal contribution over the 720 orderings, so the six values sum exactly to the difference between the base and fully adjusted HDI coefficients. Shares can be reported where the attenuation ratio cannot, because their denominator is the reduction from the base model to the full one, which runs from 0.034 to 0.217 log units across the 2,000 paired draws and never approaches zero. We also report the collapse into the earlier design's four blocks, which groups population, built form and settlement vintage as settlement.

The ladder's final block and the surface index described in Supplementary Section~\ref{sec:supp-alternatives} are two ways of representing measured surface structure, and they differ at once in the covariate and in the statistic that summarises a city, so a gap between them could come from either. We fit both on identical rows, with two hybrids that hold one difference fixed while the other varies, so the combined gap separates into its two causes; paired differences use the ladder's own 2,000 replicates drawing countries within continent.

To determine whether lower total dispersion reflects a few inactive directions or a contraction spread across all of them, we decompose it by direction across the 992 cities with a complete sample of 250 observations from urban centres, and fit the HDI models on the 977 that carry HDI and context. Directions fitted on the same cities they are scored on would find structure that does not travel, so within each of five folds that keep countries whole and balance continents we estimate the common reference and the principal directions from the other countries alone, then project the excluded cities onto them. We model both the absolute variance along each direction and its share of total variance, correcting the 63 tests in each family separately for false discovery. Sums over the first 1, 2, 3, 5, 10 and 20 directions, and over the remaining 53 after the first ten, compare spans without depending on how the axes are rotated inside them. The specification carries fixed effects for continent and for the stage at which a city entered the sample, and because the family of shares is sensitive to that choice the variant without continent is reported alongside it.

\subsection{Physical form and observation checks}
\label{sec:supp-alternatives}

To ask whether measured physical form accounts for the gradient in dispersion with development, we summarise each city's population and building fields as Shannon entropies, which measure how evenly a city's values spread across bins: ten bins of positive values fixed globally, plus a class for physical zero where that applies, normalised and corrected for small-sample bias by the Miller--Madow adjustment. WSF3D fields require 30 valid observations, the four entropies over building fields together form an index of morphology, and whether the product maps a city at all enters separately.

We then compare the HDI coefficient before and after adding each summary, their composites and all summaries together on identical city samples. Attenuation is

\begin{equation}
100\frac{\beta_{\mathrm{base}}-\beta_{\mathrm{adjusted}}}
{\beta_{\mathrm{base}}}.
\end{equation}

Intervals come from 2,000 paired bootstraps drawing countries within continent; five bins, twenty bins and interquartile ranges are sensitivity checks. This is adjustment, not causal mediation.

Climate adjustment uses nine Köppen--Geiger groups that between them cover every city, keeping tropical savanna, hot desert and hot steppe separate with the warm temperate share as the omitted reference. The surface index, read at the same points, combines Sentinel-2 NDVI with Sentinel-1 DpRVI \citep{mandal2020dprvi}, where $q=\sigma^0_{VH}/\sigma^0_{VV}$ and ${\rm DpRVI}=1-(1-q)/(1+q)^2$. Values beyond the 1st and 99th percentiles are pulled back to them and each term is standardised, giving $[z({\rm NDVI})+z({\rm DpRVI})]/\sqrt{2}$; models enter its median and interquartile range across a city, with EVI2, raw VH and the contrast between VV and VH as alternatives declared in advance. Within the fixed 942-city ladder sample, requiring 200 observations valid for both terms leaves 938 cities in 155 countries. These variables sit close to the model's own inputs, so they test whether measured surface accounts for the gradient rather than isolating a confounder from outside it.

Terrain enters as a separate block. From the 30\,m Copernicus digital elevation model we summarise each urban centre by the spread of elevation, the ninetieth percentile of local relief within 500\,m and the spread of topographic position at 1\,km, each logged and standardised; four cities in the Caucasus fall outside the model's coverage and one further centre lacks a summary, leaving 987 with terrain. The block enters after radar surface as an eighth rung of the fixed ladder on the 937 of its 942 cities the model covers, and as a seventh player in the order-invariant decomposition on the same rows, with intervals from the same 2,000 paired draws. Slope summaries shipped with the atlas were computed on an unprojected mosaic and correlate at 0.22 with slope from a projected surface, so they are not used. Coherence, the mean resultant length of a centre's pixels, is dispersion read backwards: the small-angle relation $\bar R\approx 1-D/2$ holds to within half a per cent and the two rank cities at $-0.998$, so associations are reported on dispersion and named cities on coherence.

The fixed 942-city ladder in Figure~\ref{fig:development-temporal-dynamics}B differs from this description in two measured ways: it enters raw Sentinel-1 VH summaries rather than the index above, and eight raw Köppen zone fractions rather than nine grouped shares. We report the radar difference both ways on identical rows below, because it changes the conclusion. The climate coding does not: the nine documented groups lower every rung by 0.8 to 1.8 percentage points and reorder none, leaving a final rung of 0.1\% ($-4.1$ to 4.6\%) and a documented-index rung of 4.6\% ($-0.9$ to 10.4\%).

A city that is imaged more often could be represented more richly for that reason alone, and the opportunity to observe is itself unevenly spread, so we count it at eight fixed points within urban centres in 214 cities and 133 countries for the adjustment models, and map it across 529 cities in 162 countries in Supplementary Fig.~\ref{fig:si-data-description}F. The 2024 catalogue includes Sentinel-1 GRD, Sentinel-2 Level-1C and Landsat 8/9; an optical opportunity counts as clear under Cloud Score+ ($\geq0.5$) or the standard Landsat quality flags. Models add these counts, the clear fractions, orbit geometry, imbalance between passes and the measured loss of Sentinel-1B to the HDI specification, with intervals as above.

\paragraph{The loss of Sentinel-1B.}
The loss of public Sentinel-1 coverage was timed by events outside this analysis and fell unevenly across places, which makes it a natural test of whether AlphaEarth follows what was observed rather than what is on the ground. We ask whether change in AlphaEarth coincides with it at the 2021--2022 boundary, follows local severity and is absent in other transitions, repeating one design at three scales: fixed global points, whole cities and the dispersion inside their degrees of urbanisation. A local response need not correspond to a direction that transfers between cities, so a fourth test asks whether it does.

\paragraph{Fixed points across the globe.}
Of 4,800 points uniform in longitude and sine latitude, 4,374 have finite 2019--2024 vectors outside WorldCover water and positive 2021 Sentinel-1 coverage. Scene counts by platform and pass define four mutually exclusive states: no loss or gain; some scenes lost while both passes are retained; one pass lost with coverage still positive in 2022; and no 2022 scene at all. The outcome at a point is its 2021--2022 angular movement minus its median across four other adjacent transitions. Uncertainty resamples 12-by-6 equal-area cells, and labelling other transitions as though they carried the loss supplies placebos: a contrast that appears in a year without a loss would show the design responding to something else. A replay within a single year removes duplicate acquisition dates and compares the 2021 VV/VH medians with the view from Sentinel-1A alone, and with Sentinel-1A restricted to the pass that survived into 2022, holding location and year fixed.

\paragraph{Movement of whole cities.}
All eight annual mean directions for 1,000 cities give motion between adjacent years measured on the sphere itself, transported to a 2021 tangent plane. Cities move from year to year for reasons unconnected to radar coverage, so we subtract the motion of cities in the same continent that kept their pass directions, leaving the city being measured out of that background, and centre the remaining magnitude on the city's six other transitions. The states are mutually exclusive, and a city that lost every scene takes priority: 806 cities retain their directions, 148 lose one without falling to zero and 46 contain a degree of urbanisation that lost every scene. Median contrasts resample countries, with a comparison made inside a single country as a stricter check. Which background is removed matters: removing a global one rather than one estimated within each continent leaves a contrast of about $+0.89$\textdegree{} for the cities that lost a direction, with an interval that crosses zero.

\paragraph{Dispersion within a degree and across a city.}
Exposure for the dispersion analysis uses the same fixed diagnostic points, up to eight per city and per degree of urbanisation as recorded in 2020. A degree loses a direction if any point goes from both passes in 2021 to fewer in 2022; it loses every scene if any covered point has no 2022 scene at all. The outcome uses every pixel in the annual summaries by degree, not the diagnostic points. For dispersion $D_{cst}$, change is

\begin{equation}
100\left(\frac{D_{cst}}{D_{cs,t-1}}-1\right).
\end{equation}

so median contrasts have percentage-point units, and holding the exposure labels fixed while applying them to all seven transitions supplies placebos. We estimate outcomes by degree and for pooled cities, each city weighted equally across its degrees, with a check that weights each city by the inverse of its probability of inclusion, in case the result depends on which cities the catalogue was more likely to include. The continuous dose is the mean loss in log scene count, and its outcome subtracts each city's median of the six other changes; its correlation with contraction carries an interval that resamples whole countries rather than a $p$-value that would treat cities as independent.

\paragraph{Whether the loss defines a shared direction.}
If the loss moved every affected place the same way, it would define one direction that could be found in one set of locations and recovered in another. Motion on the sphere at 2,400 points in 40 discovery cells selects the direction associated with the loss, and 1,974 points in 17 untouched cells test it and fit an independent comparison. We then carry that fixed direction to the 2024 sample and remove it before recomputing the variance hierarchy. Removing it requires all four degrees and so retains 538 cities.

\subsection{The sampling floor beneath annual movement}
\label{sec:supp-temporal-floor}

Part of a city's annual movement could be an artefact of sampling, because we draw pixels for each city afresh in every year: a city's mean direction could shift between years simply because different places inside it were drawn. To know how much movement that alone can produce, we bound the displacement expected when the underlying place is unchanged. That floor, which we call the temporal noise floor, has three components: sampling different places inside the same degree; variation at a fixed site from year to year and from one acquisition to another; and change in the representation across the whole product. Only the first can be estimated from the original city samples.

The analysis uses all 1,000 cities, every available 2020 degree of urbanisation and all eight annual layers, each year holding exactly 840,776 rows across 3,379 combinations of city and degree with each city's set of degrees and sampled counts constant, so annual movement is not manufactured by a shifting mix of degrees. Of those combinations, 3,369 hold at least two observations and can be split, leaving the floor estimable for 23,583 of 23,653 adjacent transitions within one degree and 6,930 of 7,000 adjacent city transitions.

Two halves of one year's sample differ only in which places were drawn, so the separation between their mean directions measures what drawing alone can produce. Within each city, degree and year we partition the sampled cells eight times into two equal-sized, non-overlapping sets and record the angular separation $P$ between the two half-sample mean directions. If each half holds $h$ cells and the full annual sample holds $n$, the estimated squared error of one full-sample mean direction is

\begin{equation}
\widehat V=\overline{P^{2}}\,\frac{h}{2n},
\end{equation}

and the component expected in an adjacent-year separation is the sum of the two annual values. Splits taken at the level of a city pool the corresponding halves from each degree before forming the city mean, and use the same scaling. For a declared set $G$ of transition rows, we report

\begin{equation}
R_G=1-\frac{\sum_{g\in G}\widehat N_g}{\sum_{g\in G}D_g^{2}},
\end{equation}

the share of the mean squared annual separation between mean directions that survives subtraction of the estimated spatial component. Intervals resample cities as blocks carrying all of their degrees and transitions, 2,000 times. Weighting every combination of city and degree equally, and every city equally, are primary at their respective levels; a weighting by the sampling design, which treats the original 530 cities as certain to be included, is a sensitivity check.

$R_G$ is a finite-sample excess ratio, not a reliability ratio: the other two components of the floor and real landscape change all stay in its numerator, so it cannot be read as the share of movement that is physical or reproducible.

A repeated row identifier is not a repeated cell, and we say so explicitly because the overlap between years looks total. Each year is drawn with its own seed and without keeping the geometry of a cell, and the row identifier is then built from city, year and position in the returned table, so all 840,776 identifiers recur in every adjacent pair and the degree label necessarily matches. Yet rows matched by identifier agree exactly on all five fields that should not change---elevation, slope, local relief, 2020 population and land cover---for only 0.0064 to 0.0090\% of rows, and on elevation alone for 0.012 to 0.016\%, and the longitude and latitude columns describe the city rather than the cell. The overlap is therefore an artefact of how the rows are labelled, and we do not treat the data as a panel of fixed sites.

Separating the remaining components requires extracts that follow fixed locations through the annual layers, which Supplementary Section~\ref{sec:supp-movement-tests} describes for eleven construction projects. Holding a location fixed does not establish that its surface is unchanged, so a negative control must be declared separately: surfaces known to be stable would supply it, and dated construction, land-cover conversion or vegetation disturbance supply positive controls. The construction comparisons take the latter step and do not establish the full floor. A claim to monitor change still requires movement to exceed that floor and to persist in locations and countries excluded from fitting once the opportunity to observe them is accounted for.

\subsection{Weather and construction as tests of annual movement}
\label{sec:supp-movement-tests}

The sampling floor leaves three components of annual movement unseparated: variation at fixed sites, observing conditions and change in the representation itself, alongside any real change on the ground. Two further analyses approach the remainder from opposite ends. The first asks how much of the movement measured weather, surface state and observing conditions predict. The second follows fixed pixels at dated construction projects, where change on the ground is documented, and compares their movement with matched locations elsewhere in the same city.

\paragraph{Predicting movement from weather and surface state.}
All 1,000 cities in 162 countries are retained: 8,000 city-years over 2017--2024 and 7,000 adjacent transitions, with 1,000 cities per cohort. No urbanisation-support or predictor-completeness restriction is imposed. ERA5-Land supplies monthly radiation, precipitation, snowfall and snow cover over fixed 2020 urban footprints \citep{munoz2019era5land}. MODIS supplies vegetation indices and seasonal timing and state \citep{didan2021mod13a2,friedl2022mcd12q2}; its quality block records retrieval and snow flags, viewing geometry and valid support. Sentinel-2 opportunity combines acquisition counts and Cloud Score+ visibility \citep{pasquarella2023cloudscore}. The proxies stand in for imagery AlphaEarth accepted, which is undisclosed.

Ridge models predict the full 64-dimensional transported tangent change. Tests cross five country groups with seven transition years, excluding both from training; preprocessing, imputation and penalty selection use training folds alone. Weather anomalies use 1991--2020 climatology. Gain measures reduction in pooled squared error against an intercept and centred-year baseline; unseen years receive the training grand mean. Paired intervals resample countries 5,000 times, holding predictions and observed years fixed. They do not include uncertainty from refitting or from sampling different years.

\paragraph{Fixed pixels at construction projects.}
The original city tables cannot follow a location through time (Supplementary Section~\ref{sec:supp-temporal-floor}), so we extract separate fixed sets of pixels at 10\,m through all eight annual layers at eleven construction projects, each with a declared analysis interval. In Paris we follow the same 7,847 pixels within a 500\,m radius of the olympic village centre and compare them with 17 named places across the city. The circle samples the site and its surroundings rather than the development boundary. We normalise each vector, average within each place and normalise again to form its annual mean direction. Places were chosen before their similarities were calculated, although Paris was already a known example of change. Village--place distances use the full vectors and the comparator's direction in the same year; the sphere in Supplementary Fig.~\ref{fig:si-paris-construction} illustrates the trajectory but does not preserve every pairwise angle.

To ask whether movement at a project exceeds the city's background, we match each target pixel to a similar fixed location elsewhere in its city at the first year of the project's analysis interval. Controls come from a 250\,m lattice, at least 2\,km outside the target bounding box. We greedily match by baseline cosine similarity, with no control used more than 32 times. For target $x_i$, matched control $c_i$ and baseline $s$, excess displacement is $d(x_{it},x_{is})-d(c_{it},c_{is})$. We also measure the change in distance to the nearest same-year named-place mean, subtracting the matched control's change. A positive residual means that the target has moved farther outside this local vocabulary; a negative one means it has moved closer. Medians and interquartile ranges describe pixels, not uncertainty across independent projects.

The eleven cases span nine countries and 88 extracted city-years. They include Port Lands in Toronto, SoFi/Hollywood Park in Los Angeles, Moorebank Intermodal in Sydney and Cairo's new capital. Others cover Giga Texas in Austin, Expo 2020 in Dubai, TSMC in Phoenix and airport projects in Santiago, Bangkok and Mexico City. Paris spans 2019--2024, Los Angeles 2017--2022, Mexico City 2018--2022 and the other eight cases 2017--2024. Fixed pixels are required to measure change at identical locations; no common eight-year city panel or complete set of urbanisation degrees is imposed. Local vocabularies contain 6--17 places. Cairo's control area includes its governorate because the earlier functional boundary excludes the new capital; Phoenix also excludes the surroundings of the semiconductor facilities in its comparison.

\subsection{Supplementary Results}
\label{sec:supp-results}

Each analysis below names its sample, what was excluded from it and the unit that carries its uncertainty. Where a result still rests on a fixed subsample rather than the full 1,000 cities, we say so.

The degree of urbanisation is recoverable, but not as four clean pixel classes (Supplementary Table~\ref{tab:si-prediction-benchmarks}). In a balanced audit of 408 cities with complete countries excluded from fitting, balanced accuracy is 0.366 from vectors in their absolute position and 0.375 from vectors centred on their own city, against 0.25 by chance. The ends of the scale carry it: recall is about 0.60 in peri-urban areas and urban centres, and the middle classes stay weak even when centroids are computed inside each city. A model can also fail across countries simply because cities differ; splitting each city's own observations in half removes that possibility, and among the 544 cities with equal samples in every degree it reproduces the same shape---0.542 and 0.567 at the two ends, 0.392 and 0.334 in the middle.

Built form transfers to unseen countries far better than population density does. Beyond what the degree of urbanisation already gives, $Q^2$ in countries excluded from fitting is 0.090 for population density, 0.495 for building height and 0.390 for building volume. Models fitted once and applied without refitting to a further 270 cities give 0.250, 0.566 and 0.556.

\begin{table*}[!t]
    \centering
    \fontsize{8}{10}\selectfont
    \setlength{\tabcolsep}{4pt}
    \begin{tabular*}{\textwidth}{@{\extracolsep{\fill}}lcccc@{}}
        \toprule
        \multicolumn{5}{l}{\bf A. Predicting the degree of urbanisation with complete countries excluded (408 cities)} \\
        \midrule
        {\bf Model} & {\bf Balanced accuracy} & {\bf Mean F1 over classes} & {\bf Mean error in class steps} & {\bf Within one class} \\
        \midrule
        Absolute unit vector
        & 0.366 (0.360--0.371) & 0.322 (0.314--0.330)
        & 1.038 (1.018--1.055) & 0.699 (0.691--0.708) \\
        \midrule
        City-relative tangent
        & 0.375 (0.369--0.383) & 0.342 (0.334--0.352)
        & 0.999 (0.981--1.017) & 0.719 (0.710--0.728) \\
        \bottomrule
    \end{tabular*}

    \vspace{0.8em}
    \begin{tabular*}{\textwidth}{@{\extracolsep{\fill}}lrrrr@{}}
        \toprule
        {\bf Model} & {\bf PU recall} & {\bf SD recall} & {\bf DU recall} & {\bf UC recall} \\
        \midrule
        Absolute unit vector  & 0.618 & 0.117 & 0.124 & 0.602 \\
        \midrule
        City-relative tangent & 0.616 & 0.148 & 0.164 & 0.573 \\
        \bottomrule
    \end{tabular*}

    \vspace{1.0em}
    \begin{tabular*}{\textwidth}{@{\extracolsep{\fill}}lrr@{}}
        \toprule
        \multicolumn{3}{l}{\bf B. Recovery from the other half of a city's own pixels (544 cities with equal samples in every degree)} \\
        \midrule
        {\bf Class} & {\bf Accuracy} & {\bf Median centroid gap (degrees)} \\
        \midrule
        Suburban / peri-urban (PU) & 0.542 & 4.17 \\
        \midrule
        Semi-dense urban (SD)      & 0.392 & 3.80 \\
        \midrule
        Dense urban (DU)           & 0.334 & 3.82 \\
        \midrule
        Urban centre (UC)          & 0.567 & 4.03 \\
        \bottomrule
    \end{tabular*}

    \vspace{1.0em}
    \begin{tabular*}{\textwidth}{@{\extracolsep{\fill}}llrrr@{}}
        \toprule
        \multicolumn{5}{l}{\bf C. Predicting continuous settlement characteristics} \\
        \midrule
        {\bf Target} & {\bf Transform} & {\bf Cells} & {\bf Cities} & {\bf Countries} \\
        \midrule
        Population density (GHSL) & log1p & 3,363 & 1,000 & 162 \\
        \midrule
        Building height (WSF3D)   & log1p & 3,336 &   998 & 162 \\
        \midrule
        Building volume (WSF3D)   & log1p & 3,336 &   998 & 162 \\
        \bottomrule
    \end{tabular*}

    \vspace{0.8em}
    \begin{tabular*}{\textwidth}{@{\extracolsep{\fill}}lrrrr@{}}
        \toprule
        {\bf Target} & {\bf $Q^2$ in excluded countries} & {\bf Training cities} & {\bf Extension cities} & {\bf $Q^2$ on extension cities} \\
        \midrule
        Population density (GHSL) & 0.090 & 730 & 270 & 0.250 \\
        \midrule
        Building height (WSF3D)   & 0.495 & 728 & 270 & 0.566 \\
        \midrule
        Building volume (WSF3D)   & 0.390 & 728 & 270 & 0.556 \\
        \bottomrule
    \end{tabular*}

    \vspace{0.8em}
    \begin{minipage}{\textwidth}
        \textit{Notes.} Panel A uses a balanced sample of 408 cities in 117
        countries at 50 observations per cell, not recomputed on the expanded
        catalogue; parentheses are 95\% country-then-city bootstrap intervals.
        Panel B uses the 544 cities with equal samples in every degree, with accuracy
        cross-fitted over repeated half-splits and the centroid gap the median
        angular disagreement between opposite halves. Panel C starts with 1,000
        cities and requires 30 valid measurements per target cell; $Q^2$ is the
        error reduction beyond a fold-local model that already knows urban
        class, and the final column trains on 730 cities and predicts 270
        without refitting. Panels A and B classify pixels while panel C predicts
        cell means, so their metrics are not comparable.
    \end{minipage}
    \caption{\textbf{The four degrees of urbanisation resist recovery from individual embeddings, while continuous settlement characteristics transfer well.} Prediction collapses in the two middle classes and survives only at the ends of the scale, whether complete countries are excluded from fitting (panel A) or each city's own observations are split in half (panel B). Among the continuous targets, predicted as means for each city and class (panel C), building height and volume transfer several times better than population density.}
    \label{tab:si-prediction-benchmarks}
\end{table*}

\subsubsection{Samples and coverage}

\begin{figure*}[p]
    \centering
    \includegraphics[width=\textwidth,keepaspectratio]{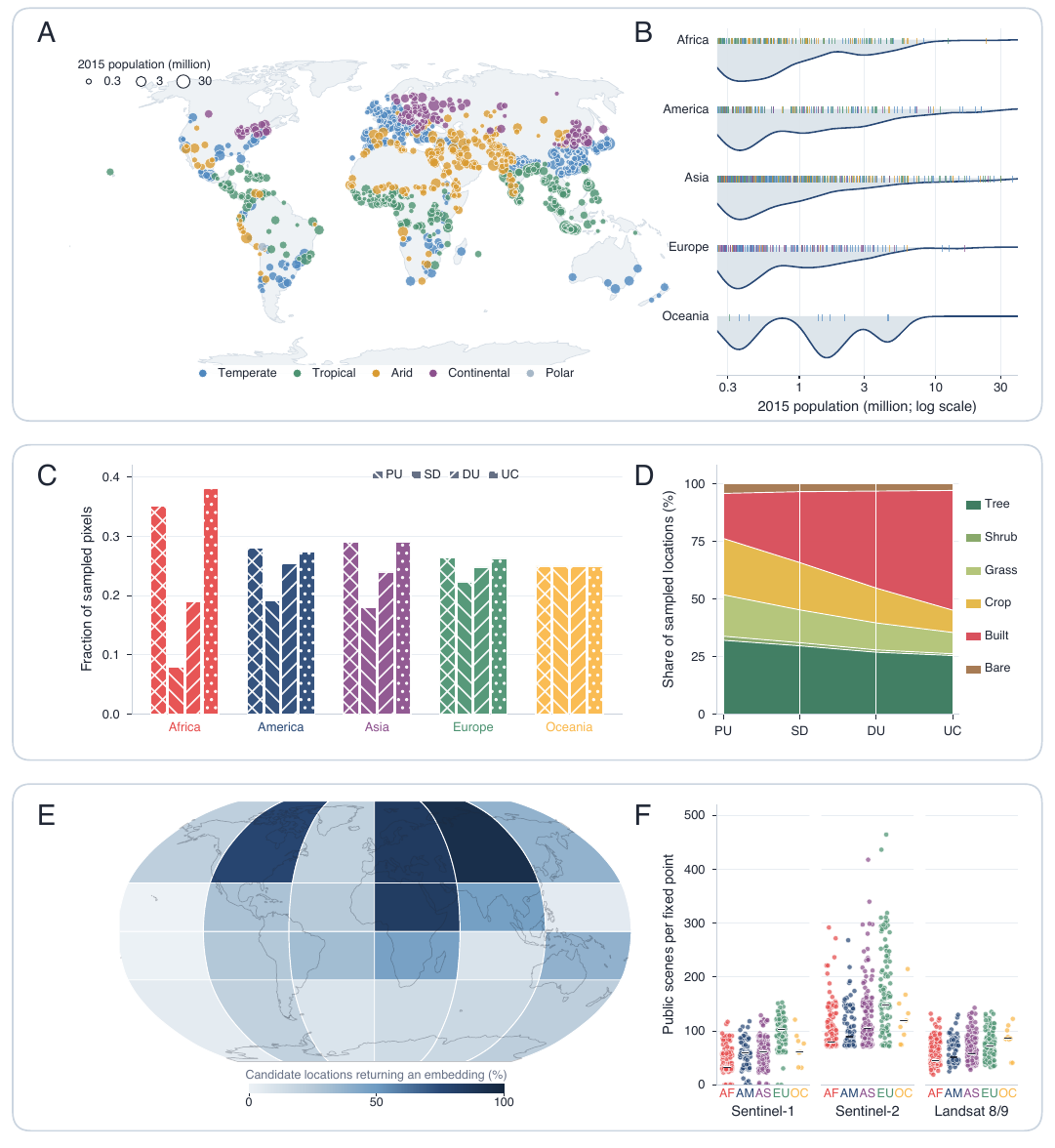}
    \caption{\textbf{The catalogue reaches 162 countries, and the samples drawn from it are balanced by design rather than by area or population.} All 1,000 catalogued urban areas are mapped with their climate and their 2015 population (A,B). What each analysis then sees differs by design: the 2024 sample stratified by degree draws 840,776 observations from all 1,000 cities and leaves continents with uneven shares of their sampled pixels, whereas the WorldCover comparison with equal samples in every degree holds 544 cities in 117 countries to exactly 250 observations per class, 544,000 in all, across which cover moves from tree and crop towards built surface between peri-urban (PU) and urban centre (UC), through semi-dense (SD) and dense urban (DU) (C,D). Beyond the cities, 307,273 of 960,000 equal-area candidates return an embedding, and the opportunity to observe from public satellites, measured at fixed points in 529 cities in 162 countries, varies with instrument and continent, abbreviated AF, AM, AS, EU and OC (E,F).}
    \label{fig:si-data-description}
\end{figure*}

The catalogue holds 1,000 urban areas in 162 countries above a 2015 population of 250,000, and it is a sample rather than a census of every city that qualifies (Supplementary Fig.~\ref{fig:si-data-description}). The 2024 table draws 840,776 locations from them, stratified by the four degrees of urbanisation as fixed in 2020 and capped at 250 observations per city and class.

Analyses that need every city and class equally represented use the 544 cities in 117 countries that carry 250 observations in each class, or 544,000 observations. Across that sample built-up share rises from 19.5\% in peri-urban locations to 52.0\% in urban centres as tree and crop shares fall---a composition of the sample rather than of physical area or population. AlphaEarth returns an embedding at 307,273 of 960,000 equal-area candidates, and the global reference keeps the 247,565 of those that carry an ESA WorldCover land label. The opportunity to observe a city from public satellites is mapped across 529 cities in 162 countries, and the models that adjust for it are confined to 214 cities in 133 countries; none of these external catalogues is an AlphaEarth input.

\subsubsection{Planetary context of the urban representation}

\begin{figure*}[p]
    \centering
    \includegraphics[width=\textwidth,keepaspectratio]{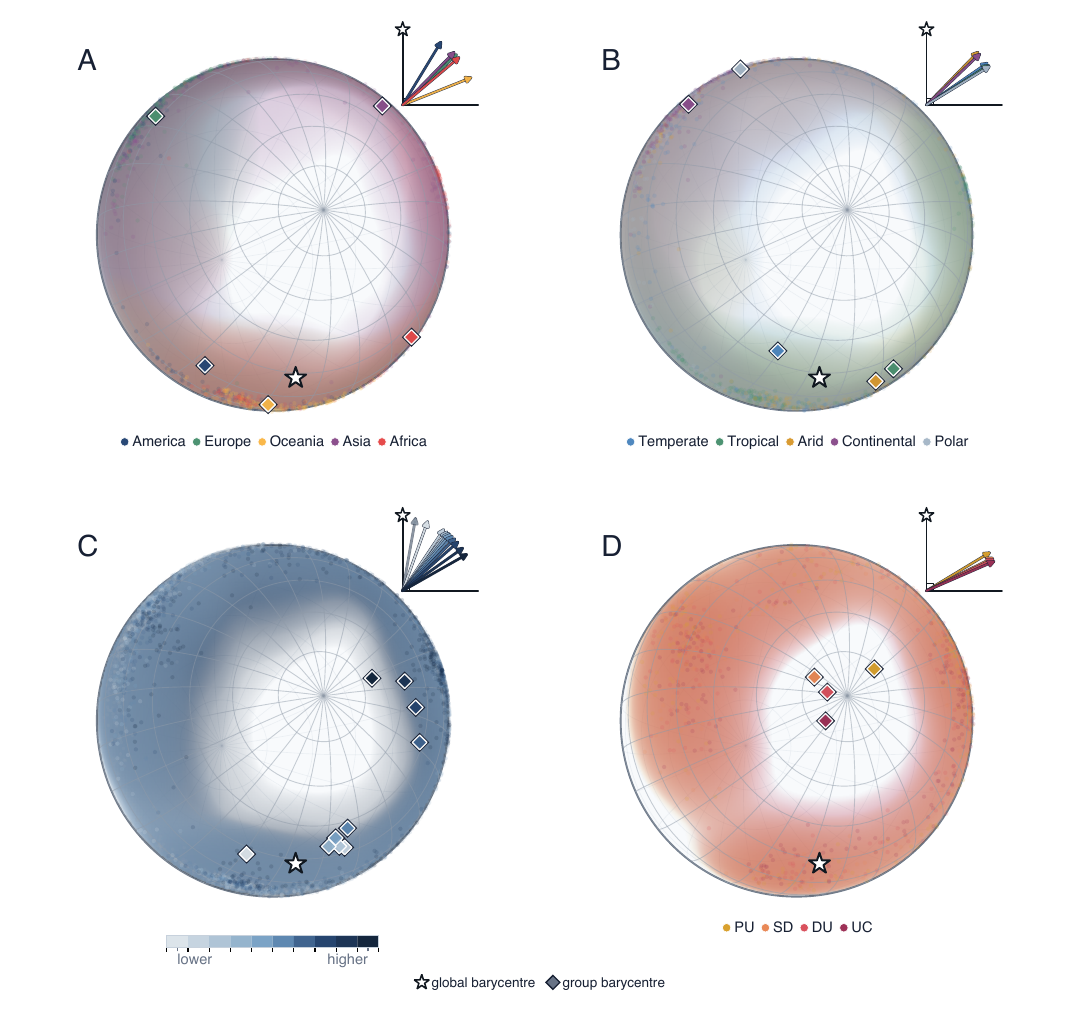}
    \caption{\textbf{Contextual groups sit at displaced mean directions and still overlap heavily, so no one context sorts the representation.} Continent and climate densities pull their means apart yet share most of their range (A,B). Population density moves outward in an orderly way, from 19.8\textdegree{} off the global mean in the lowest positive decile to 60.4\textdegree{} in the highest, whereas the four degrees of urbanisation crowd into a narrow band of similarly displaced radii, 58.8 to 66.5\textdegree{} (C,D). Panels A to C rest on the 247,565-pixel global land sample and panel D on all 1,000 cities. Distance from the centre of each disc and the inset arrows give exact angles.}
    \label{fig:si-sphere-context}
\end{figure*}

Contextual groups overlap but sit at different mean directions. Population density moves outward in order, from 19.8\textdegree{} off the global mean in the lowest positive decile to 60.4\textdegree{} in the highest. By contrast the means of the four degrees of urbanisation, taken over 3,379 combinations of city and degree and 215,258 pixels, all lie between 58.8 and 66.5\textdegree{} away (Supplementary Fig.~\ref{fig:si-sphere-context}). All of it is measured in the original $S^{63}$ geometry, and all of it is descriptive: these patterns do not separate the effects of geography, climate, population and urbanisation.

The same reference puts the separation between urban and global mean directions on a scale. Each of its 247,565 locations carries an ESA WorldCover land label and a degree of urbanisation, so every major cover class on Earth can be measured against the same global barycentre on the same points (Supplementary Table~\ref{tab:si-global-class-distances}). No distinctive cover sits close to it: cropland lies 47.96\textdegree{} away, built-up land 66.44\textdegree{} and mangroves 76.03\textdegree{}. The reason is that 92.2\% of the reference is very-low-density rural land whose own barycentre is 4.64\textdegree{} from the global one. The global mean direction is the mean direction of remote rural land, and 62.66\textdegree{} is where an urban barycentre falls on that scale rather than a distance that sets cities apart from everything else.

\begin{table}[!t]
    \centering
    \fontsize{8}{10}\selectfont
    \setlength{\tabcolsep}{4pt}
    \begin{tabular*}{\columnwidth}{@{\extracolsep{\fill}}lrr@{}}
        \toprule
        {\bf Class} & {\bf Points} & {\bf Angle (\textdegree)} \\
        \midrule
        \multicolumn{3}{l}{\it ESA WorldCover} \\
        Mangroves                 &    332 & 76.03 \\
        Snow and ice              &  4,595 & 70.78 \\
        Built-up                  &  1,827 & 66.44 \\
        Bare or sparse vegetation & 42,330 & 56.63 \\
        Moss and lichen           &  6,151 & 53.04 \\
        Shrubland                 & 19,075 & 52.20 \\
        Cropland                  & 23,696 & 47.96 \\
        Herbaceous wetland        &  4,137 & 47.90 \\
        Tree cover                & 83,946 & 39.33 \\
        Grassland                 & 61,476 & 28.33 \\
        \midrule
        \multicolumn{3}{l}{\it Degree of Urbanisation} \\
        Urban centre              &  1,098 & 66.78 \\
        All urban (class 21 or above) & 5,724 & 61.27 \\
        Rural cluster             &  1,429 & 53.53 \\
        Low density rural         & 11,355 & 49.77 \\
        Very low density rural    & 228,269 & 4.64 \\
        \bottomrule
    \end{tabular*}
    \caption{\textbf{Every major class sits tens of degrees from the global barycentre, which puts the 63\textdegree{} urban separation on a scale.} Angles run from 28.33\textdegree{} for grassland to 76.03\textdegree{} for mangroves, and a single reference pixel sits a median 72.00\textdegree{} away, so distance from the global mean direction is the ordinary condition and not a mark of distinctiveness. Only very-low-density rural land sits close, at 4.64\textdegree{}. Angles are measured on the 247,565 reference points themselves, every one of which carries a WorldCover land label, and they weight pixels rather than cities, so they calibrate the 62.66\textdegree{} figure, which weights cities equally, rather than coinciding with it.}
    \label{tab:si-global-class-distances}
\end{table}

\subsubsection{Physical and environmental correlates of urban separation}

\begin{figure*}[p]
    \centering
    \includegraphics[width=0.97\textwidth,keepaspectratio]{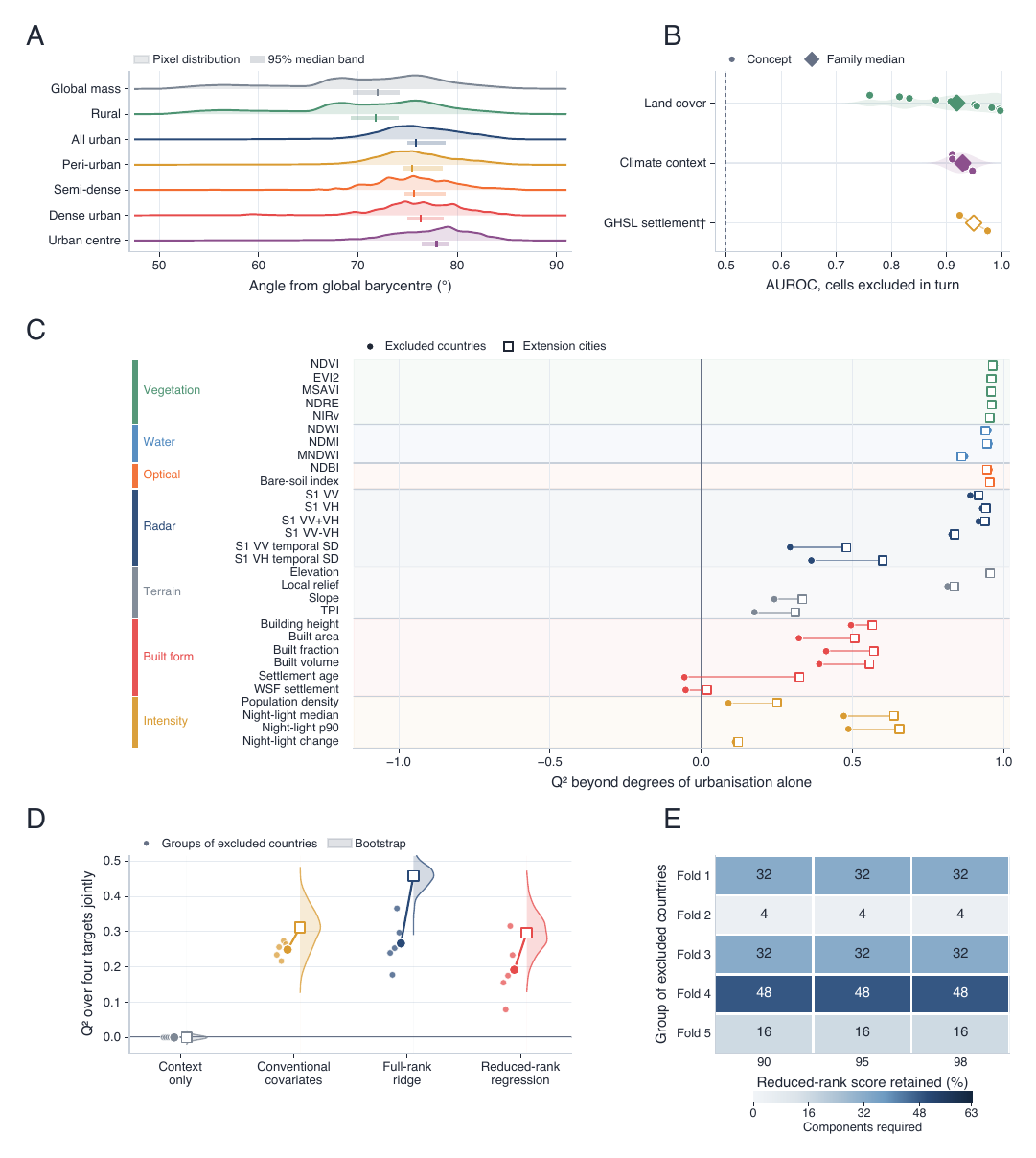}
    \caption{\textbf{Urban separation aligns with measured surface and environmental variables but stays high-dimensional.} Pixels in urban centres lie 6.09\textdegree{} farther from the global mean than rural pixels, and broad land cover and climate transfer to equal-area cells excluded from fitting at a median AUROC, the chance that a positive location scores above a negative one, of 0.919 and 0.929 (A,B). In excluded countries, vegetation, water and radar targets transfer better than built form or intensity (C). None of that collapses into a few directions. After context adjustment a model confined to a few directions reaches $Q^2$ of 0.192 and 0.296 against 0.267 and 0.458 for one using all of them, and holding its score in principal components fitted at the mean direction of all 1,000 cities takes a median 32 components---48 in one group of excluded countries---a count that shifts with the basis while $Q^2$ does not (D,E). Panels C to E use 728 cities in 162 countries and a further 270 extension cities in 46 countries.}
    \label{fig:si-urban-semantics}
\end{figure*}

Angular distances, directions fitted inside a fold and fitted subspaces are unchanged by a common orthogonal rotation of the embedding, so correlates mapped independently can be read against them. Pixels in urban centres lie 6.09\textdegree{} farther from the global mean than rural pixels (95\% interval 3.40 to 8.56\textdegree{} when the equal-area cells are resampled; Supplementary Fig.~\ref{fig:si-urban-semantics}A), and in equal-area cells excluded from fitting the median AUROC---the chance that a positive location scores above a negative one---is 0.919 across 11 land-cover concepts and 0.929 across five climate families. GHSL settlement contrasts are a positive control, since the same family of labels defines the urban comparison.

Within cities, models fitted with whole countries excluded predict independently mapped targets beyond a baseline that already knows the degree of urbanisation. Median $Q^2$ is high for vegetation targets, much lower for built form, and weaker and less consistent again for population and light. Yet the reduced-rank model, though it beats permutations of whole cities ($p=0.001$), does not pass the stopping rule set before the analysis: $Q^2$ is 0.192 in the excluded countries and 0.296 in the extension cities, against 0.267 and 0.458 for the full-rank ridge and 0.249 and 0.312 for conventional covariates. Holding 90\% of that score takes a median 32 principal components fitted at the mean direction of all 1,000 cities, and 48 in one fold. The count belongs to the basis rather than to the representation: it moves under random rotation while $Q^2$ does not.

\begin{figure*}[p]
    \centering
    \includegraphics[width=\textwidth,height=0.68\textheight,keepaspectratio]{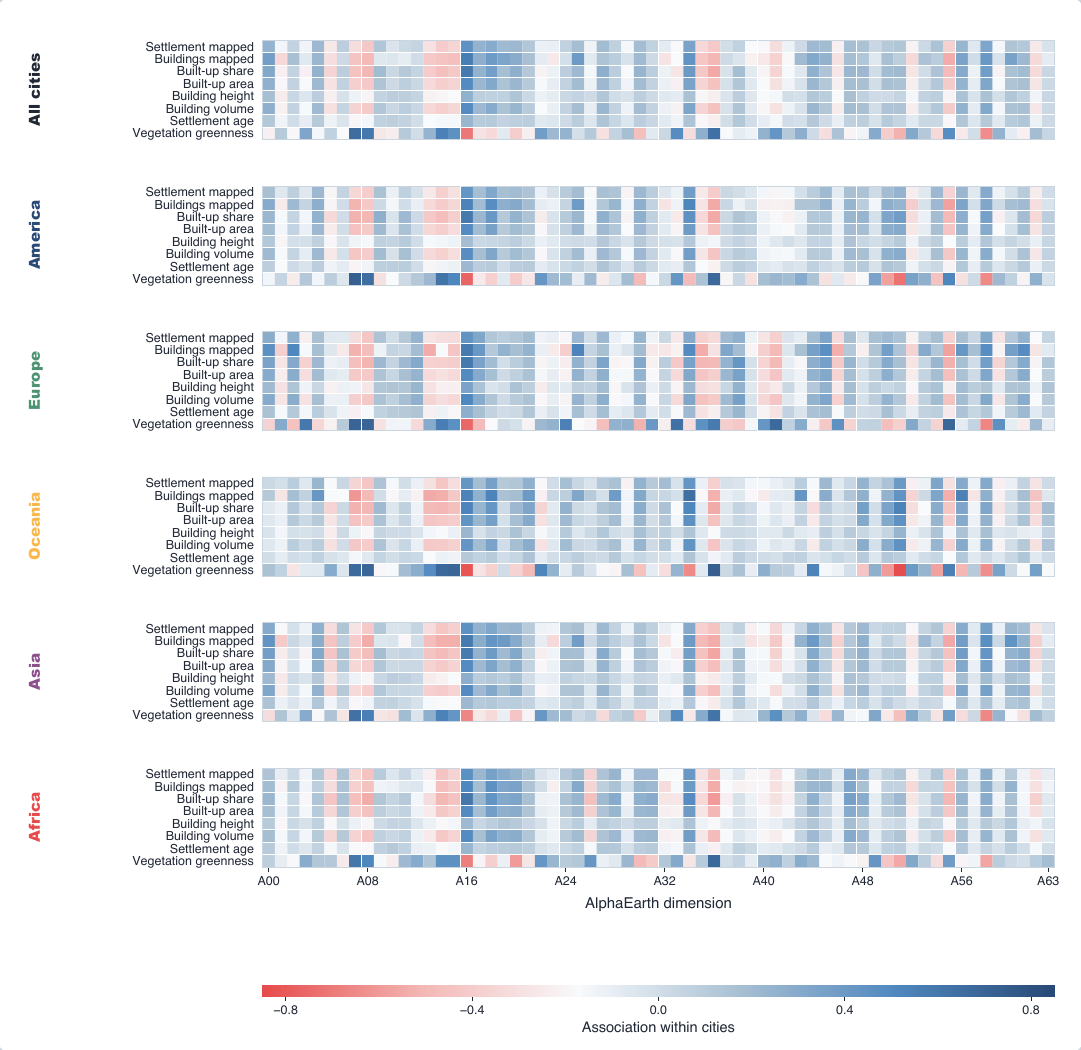}
    \caption{\textbf{Built-form and vegetation associations spread across the released coordinates rather than settling on any one of them.} Colour gives the mean association inside a city, each city weighted equally, across eligible members of the 1,000 cities sampled in 2024, reaching $-0.84$ and $+0.73$ on a fixed $\pm0.85$ scale. Every target draws on broad bands of coordinates rather than one apiece, the built-form rows share a profile that vegetation greenness reverses at several coordinates, and similar bands recur across continents. A row recording whether a target is mapped cannot separate physical absence from missing source coverage, and the settlement and building layers are nominally 2019 while settlement age covers 1985--2015.}
    \label{fig:si-native-coordinate-associations}
\end{figure*}

The available targets describe environment, surface composition, built form and intensity, not human function: no independent activity or land-use labels were available, and night light is an intensity proxy. The fixed-basis displays in Supplementary Figs.~\ref{fig:si-native-coordinate-activity} and \ref{fig:si-native-coordinate-associations} likewise show variance and physical associations spread across released coordinates. They do not define intrinsic axes or support causal interpretation.

\subsubsection{Identifying cities from their own pixels}
\label{sec:supp-identification}

Every city is identified from half of its own pixels. Across five random splits, the mean direction of a city's second half lies nearest the mean direction of its own first half in 5,000 of 5,000 trials, against a chance rate of 0.1\%. The half held out sits a median 2.12\textdegree{} from its own mean direction and 13.01\textdegree{} from the nearest other city's, a median margin of 10.87\textdegree{}, and the smallest margin in any trial is 0.25\textdegree{}, Leeds against Sheffield. A single pixel is a different object. Assigned to the nearest of the 1,000 mean directions, it returns to its own city 52.4\% of the time (48.1 to 56.8\% when countries are resampled), lies within the nearest five 82.4\% of the time and within the nearest twenty 96.6\%, against chance of 0.1, 0.5 and 2.0\%. Averaging is what buys the identification: the mean of five pixels names the city 86.3\% of the time, of twenty 97.9\% and of sixty 99.8\% (Supplementary Table~\ref{tab:si-city-identification}).

The errors are local. Of the 118,925 misassigned pixels, 54.4\% go to a city in the same country where a uniform draw would send 3.6\%, 92.4\% stay on the same continent and 78.5\% in the same climate family, and the median distance to the wrongly chosen city is 323\,km against 6,780\,km under the null. The nearest wrong city, the one a city would be confused with if it were confused with anything, is sharper still: 70.7\% lie in the same country, at a median 173\,km, and for 805 of the 1,000 cities it is the same city in all five splits. Pixel accuracy also falls with isolation, from 93.3\% among the 14 cities whose nearest neighbour lies more than 30\textdegree{} away to 18.6\% among the 20 whose nearest neighbour lies within 6\textdegree{}, while the assignment of whole cities is correct in every bin.

Resolution runs out only between neighbours. Bootstrapping each city's mean direction from 250 of its own pixels gives a median angular standard error of 2.12\textdegree{} (5th to 95th percentile 1.37 to 2.74\textdegree{}), and two of the 499,500 city pairs sit closer than twice that scale: Leeds and Sheffield at 3.43\textdegree{} and Sheffield and Nottingham at 4.04\textdegree{}. A rule that combines the two cities' own errors admits 25 pairs, 0.005\% of the catalogue, every one a pair of cities in the industrial belt of England, the Randstad, the Flemish diamond, the Rhine or the corridor from Boston to Providence. At the count the atlas actually carries, a median 1,000 urban pixels per city, the standard error falls to 1.14\textdegree{} and no pair remains within the noise; the closest pair sits 1.50 times its own noise scale apart.

\begin{table*}[!t]
    \centering
    \fontsize{8}{10}\selectfont
    \setlength{\tabcolsep}{4pt}
    \begin{tabular*}{\textwidth}{@{\extracolsep{\fill}}lccc@{}}
        \toprule
        \multicolumn{4}{l}{\bf A. Identification accuracy over five random splits (1,000 cities, 162 countries)} \\
        \midrule
        {\bf Query} & {\bf Nearest one (\%)} & {\bf Nearest five (\%)} & {\bf Nearest twenty (\%)} \\
        \midrule
        Mean direction of a city's held-out half & 100.0 (100.0--100.0) & 100.0 (100.0--100.0) & 100.0 (100.0--100.0) \\
        \midrule
        A single pixel from that half            & 52.4 (48.1--56.8)   & 82.4 (79.1--85.8)   & 96.6 (95.8--97.5) \\
        \midrule
        Chance                                   & 0.1 & 0.5 & 2.0 \\
        \bottomrule
    \end{tabular*}

    \vspace{0.8em}
    \begin{tabular*}{\textwidth}{@{\extracolsep{\fill}}lrrrr@{}}
        \toprule
        \multicolumn{5}{l}{\bf B. Where the errors go} \\
        \midrule
        {\bf Quantity} & {\bf Pixel errors} & {\bf Null} & {\bf Nearest wrong city} & {\bf Null} \\
        \midrule
        Same country (\%)        & 54.4 & 3.6   & 70.7 & 3.1 \\
        \midrule
        Same continent (\%)      & 92.4 & 31.6  & 96.7 & 30.4 \\
        \midrule
        Same climate family (\%) & 78.5 & 29.5  & 85.9 & 28.7 \\
        \midrule
        Median distance (km)     & 323  & 6,780 & 173  & 7,014 \\
        \midrule
        Within 500\,km (\%)      & 66.0 & 1.6   & 80.5 & 1.3 \\
        \bottomrule
    \end{tabular*}

    \vspace{0.8em}
    \begin{tabular*}{\textwidth}{@{\extracolsep{\fill}}lrrrrr@{}}
        \toprule
        \multicolumn{6}{l}{\bf C. The noise scale of a city's mean direction and the pairs that fall within it} \\
        \midrule
        {\bf Pixels per city} & {\bf Median error (\textdegree)} & {\bf 5th to 95th} & {\bf Twice the median} & {\bf Pairs below} & {\bf Pairs below, combined} \\
        \midrule
        250                        & 2.12 & 1.37 to 2.74 & 4.23 & 2 & 25 \\
        \midrule
        Own count (median 1,000)   & 1.14 & 0.80 to 1.56 & 2.28 & 0 & 0 \\
        \bottomrule
    \end{tabular*}

    \vspace{0.8em}
    \begin{tabular*}{\textwidth}{@{\extracolsep{\fill}}lrrrrrrrrrr@{}}
        \toprule
        \multicolumn{11}{l}{\bf D. Accuracy against the number of pixels averaged in the query} \\
        \midrule
        {\bf Pixels}       & 1 & 2 & 3 & 5 & 8 & 12 & 20 & 35 & 60 & 100 \\
        \midrule
        Nearest one (\%)   & 52.8 & 69.2 & 78.2 & 86.3 & 91.7 & 94.9 & 97.9 & 98.6 & 99.8 & 99.8 \\
        \midrule
        Nearest five (\%)  & 82.6 & 92.3 & 96.2 & 98.3 & 99.3 & 99.8 & 100.0 & 100.0 & 100.0 & 100.0 \\
        \bottomrule
    \end{tabular*}

    \vspace{0.8em}
    \begin{minipage}{\textwidth}
        \textit{Notes.} Panel A scores each city's held-out half, and 50 single
        pixels from it, against the 1,000 mean directions built from the other
        halves; parentheses are 95\% intervals that resample countries. Panel B
        follows the 118,925 pixel errors and, because whole cities are never
        misassigned, the nearest wrong city in each of the 5,000 trials, each
        against the exact expectation of a uniform draw from the other 999
        cities. Panel C bootstraps each city's mean direction 200 times at 250
        pixels and at its own pooled count, and counts the pairs among 499,500
        that lie within twice the median error, or within twice the combined
        error of the two cities. Panel D averages a fixed number of held-out
        pixels before assignment, against the full mean directions of the other
        halves; its one-pixel row repeats panel A under fresh draws.
    \end{minipage}
    \caption{\textbf{Every city is identified from half of its own pixels, a single pixel finds its city half the time, and the pairs the representation cannot separate are neighbours.} Averaging carries the identity: a city's held-out half is never misassigned, a single pixel is misassigned half the time, and twenty pixels are enough to name the city in 98\% of trials. Where pixels err they err locally, and the only pairs within the noise of a 250-pixel mean direction are neighbouring cities in the same conurbation.}
    \label{tab:si-city-identification}
\end{table*}

\subsubsection{Geographic context and the degree of urbanisation}

\begin{figure*}[p]
    \centering
    \includegraphics[width=0.95\textwidth,keepaspectratio]{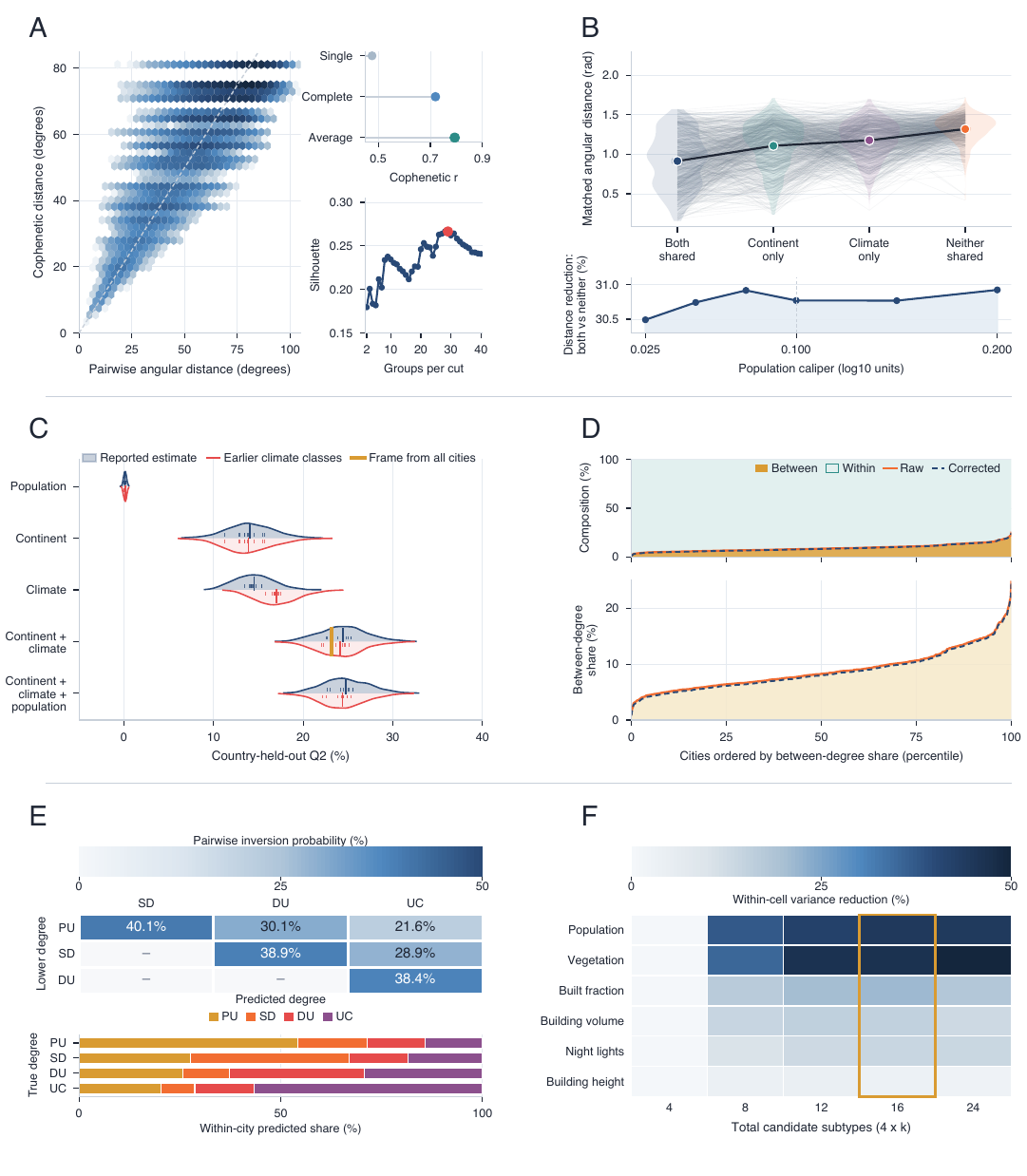}
    \caption{\textbf{Geographic context transfers across countries, while the degrees of urbanisation leave much of the structure inside a city unresolved.} The 1,000-city hierarchy reproduces its own pairwise distances well and still cuts badly: the tree matches the distances it was built from at a correlation of 0.793, yet the best cut separates its groups at a silhouette of only 0.266, at 29 groups, where 1 would be clean (A). Shared continent and climate keep their advantage whichever population caliper is used, the reduction holding between 30.5 and 30.9\% (B), and context predicts mean directions in wholly excluded countries at a $Q^2$ of 0.243 (C). Inside cities the picture reverses: only 8.88\% of variation separates the four class means (D), and on those same cities the class ordering inverts across countries excluded from fitting (E). Validated on cities held out from fitting, splitting the degrees further reduces the spread of population and vegetation most (F).}
    \label{fig:si-geography-dou}
\end{figure*}

The hierarchy of 1,000 city means preserves 499,500 pairwise angles at a cophenetic correlation of 0.793, which scores how faithfully a tree reproduces the distances it was built from, but no tested cut is well separated. The largest silhouette score, which compares how close a city sits to its own group against the nearest other group, is 0.266 at 29 groups, interior to the tested 2--40 grid, and partitioning around medoids gives 0.264 at 17 groups (Supplementary Fig.~\ref{fig:si-geography-dou}A). Weak separation could still beat no structure at all, so we compare it with 200 configurations matched to the same distance geometry but carrying no groups: separation exceeds every one of them, which rules out a featureless cloud and nothing more. A gap statistic selects no division and the distance distribution is unimodal. The tree is a descriptive ordering, not a global taxonomy of cities.

At the primary population caliper, 948 anchors from 160 countries have all four matches. Cities sharing continent and climate are separated by 52.4\textdegree{} on average (95\% interval 50.5 to 54.4\textdegree{} when countries are resampled within continents), compared with 75.6\textdegree{} (74.6 to 76.8\textdegree{}) when neither is shared, a 30.8\% reduction. Matches are selected without AlphaEarth distance or HDI, and a tighter or looser caliper would change how alike the matched cities are in size: the reduction remains 30.5--30.9\% across the tested calipers (Supplementary Fig.~\ref{fig:si-geography-dou}B). Matches inside the anchor's own country could carry the effect on their own, since a country shares more than continent and climate; they make up 11.9\% of the cell sharing both and contribute 3.9 percentage points of the reduction, and excluding them leaves 26.9\%. In repeated splits that exclude whole countries, across 992 cities, continent and climate explain 24.3\% of tangent variation ($Q^2=0.243$, interval 0.196 to 0.297), while population alone explains essentially none. Two related estimates of the same transfer sit alongside it, one under an earlier grouping of the climates and one whose tangent frame the excluded cities help define (Supplementary Fig.~\ref{fig:si-geography-dou}C).

The audit with equal samples in every degree retains 544 cities in 117 countries with 250 observations in each class. Differences among the four class means account for 8.88\% of the variation inside a city (8.32 to 9.37\% when whole countries are resampled), leaving 91.12\% within classes; corrected for noise, the share between classes is 8.60\% (Supplementary Fig.~\ref{fig:si-geography-dou}D). On the same subsample, only 70.8\% of cities have all four class means in the expected order, and pixels from peri-urban areas and from urban centres swap order in 21.6\% of pairs drawn at random inside a city. Recovery from one half of a city's observations to the other is strongest at the two endpoints and weakest in the middle (Supplementary Fig.~\ref{fig:si-geography-dou}E). The degree of urbanisation is detectable, but does not form four clean pixel classes.

What the four degrees leave unresolved could be noise, in which case subdivisions of it would carry nothing beyond the cities they were learned in. Subdivisions learned without the confirmation countries nevertheless separate targets that were not used to fit them. At four subdivisions per class, the variance left inside a cell, measured by cross-fitting, falls by 46.6\% for vegetation, 44.5\% for population, 21.2\% for built fraction, 15.4\% for building volume, 13.9\% for night lights and 3.3\% for building height (Supplementary Fig.~\ref{fig:si-geography-dou}F). Four subdivisions are illustrative rather than optimal, and because the sampled locations were drawn without their coordinates, none of this establishes that the subdivisions are contiguous on the ground or coherent as a map.

\clearpage

\clearpage

\clearpage

\clearpage
\begin{figure*}[p]
    \centering
    \includegraphics[width=\textwidth,height=0.68\textheight,keepaspectratio]{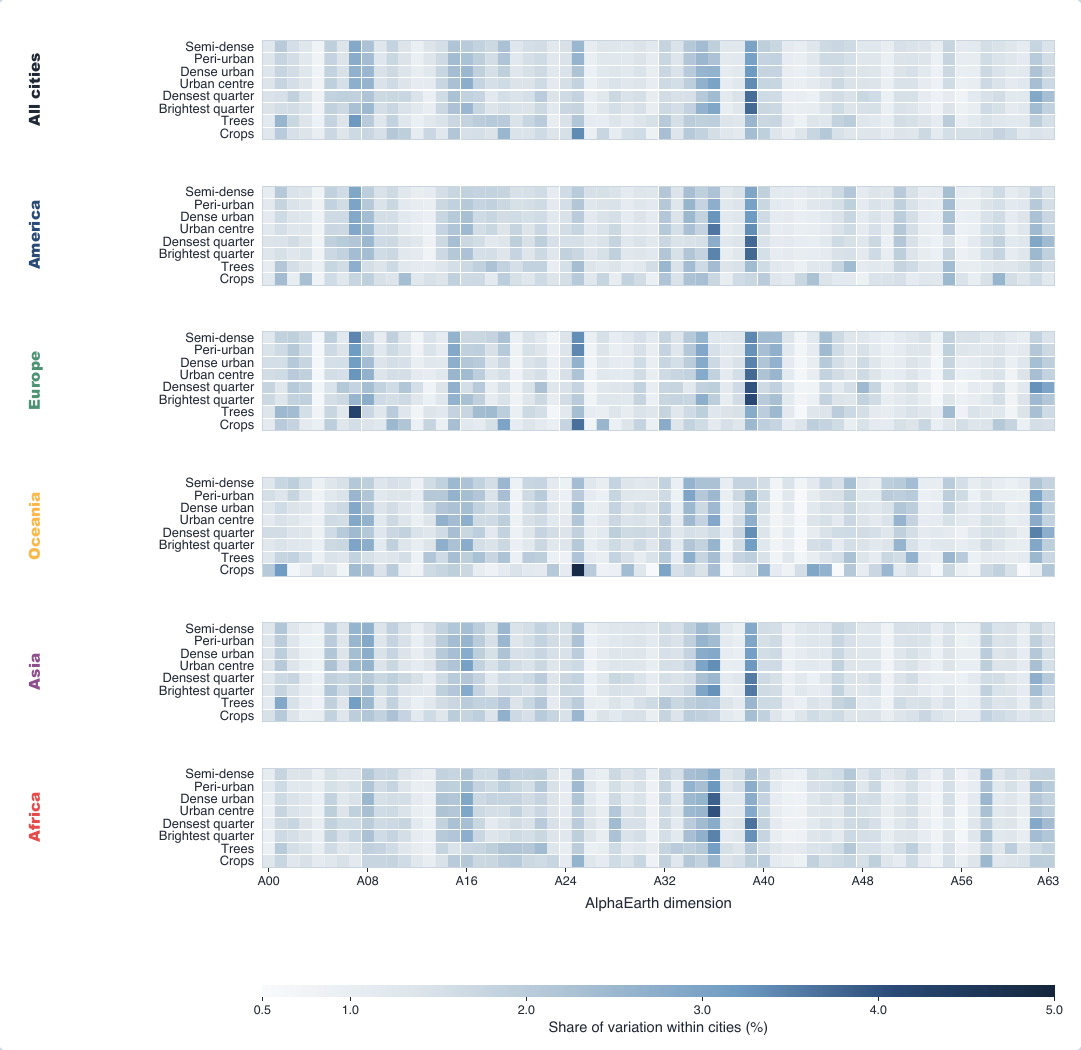}
    \caption{\textbf{Variance is spread across the released coordinates rather than concentrated in a few of them.} Each cell gives one coordinate's share of within-city variance after city centring and unit normalisation. A handful of coordinates stand out and recur across continents, but no row rests on one of them, and the urban rows share a profile that the tree and crop rows, and the rows for the brightest and densest quarter of cities, only partly follow. Rows for the degrees of urbanisation balance 544 cities in 117 countries at 250 pixels per class, giving 136,000 pixels per class row, while tree and crop rows retain 539 and 490 eligible cities under equal-city weight and Oceania crops rest on four cities. The profiles describe the released basis, not principal components.}
    \label{fig:si-native-coordinate-activity}
\end{figure*}

\clearpage

\clearpage

\clearpage
\subsubsection{City position and the orientation of internal variation}

A city's mean direction and the orientation of the variation inside it are different quantities, and across the same 1,000 cities in 162 countries shown in Fig.~\ref{fig:city-structure}A they agree locally. Over 499,500 city pairs, the great-circle distance between mean directions and the distance between the leading ten directions of internal covariance, transported to a common plane, correlate at $\rho=0.703$, exceeding every one of 999 permutations of the labels ($p=0.001$). The same city is the nearest neighbour in both spaces for 33.8\% of cities, and their sets of ten nearest neighbours overlap by 53.9\% on average.

\begin{figure*}[p]
    \centering
    \includegraphics[width=\textwidth,keepaspectratio]{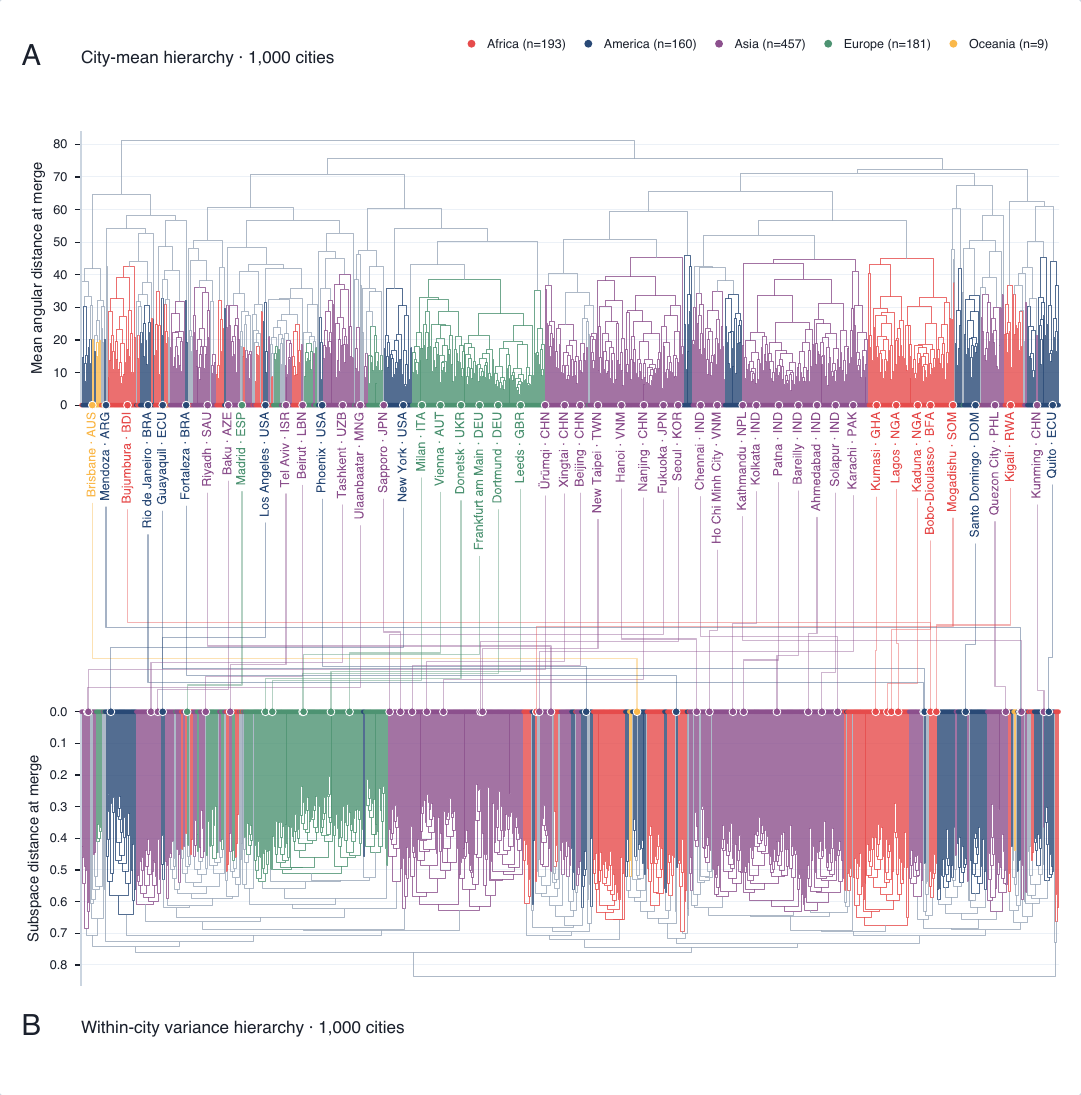}
    \caption{\textbf{Mean position and variance orientation share neighbourhoods but not major branches.} Across all 1,000 cities the two pairwise distance rankings correlate at $\rho=0.703$ and ten-nearest-neighbour sets overlap by 53.9\%, so a city's near neighbours are largely the same under both constructions. The trees built from them are not: cophenetic correlation falls to 0.336, and only five exact clades of ten or more cities are shared, the largest holding 26. Read the two against one another rather than down either one---the lower tree is mirrored at 484 of its 999 binary nodes to bring identical leaves alongside, leaving topology and merge heights untouched, so the displayed Kendall $\tau=0.440$, the rank agreement between the two leaf orders, diagnoses that alignment rather than any fit, and is $-0.065$ before it. Terminal colours mark continent and the 50 names match Fig.~\ref{fig:city-structure}A.}
    \label{fig:si-mean-variance-hierarchies}
\end{figure*}

Complete trees agree much less. The correlation between their cophenetic distances is $\rho=0.336$, and only five clades shared exactly between them contain ten or more cities, the largest holding 26. Mean position and internal variance therefore share local geometry without defining one taxonomy at the level of large branches (Supplementary Fig.~\ref{fig:si-mean-variance-hierarchies}).

\begin{figure*}[p]
    \centering
    \includegraphics[width=\textwidth,keepaspectratio]{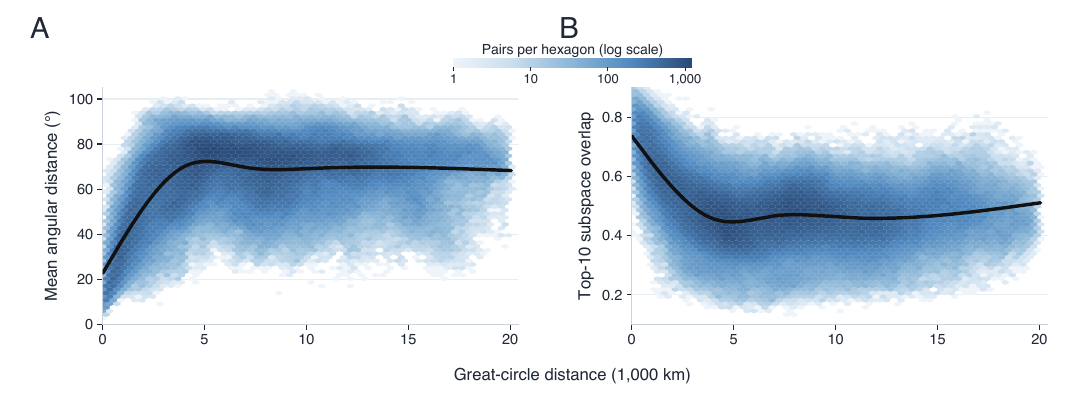}
    \caption{\textbf{Geographic distance organises both where a city's urban core sits and how its internal variation is oriented.} Cities near one another on the ground have closer urban-core mean directions (A) and the leading ten directions of their internal covariance overlap more (B). Neither relation decays to indifference: overlap settles well above the reference line, the 0.159 expected of two random ten-dimensional subspaces, rather than at zero. Five folds that exclude cities from fitting give spline $R^2=0.413$ for mean angular distance and 0.297 for the overlap of the leading ten directions, and the curves are descriptive rather than inferential. The comparison keeps 992 of the 1,000 cities, in 162 countries, each with 250 valid observations from its urban centres; the eight excluded have none, and are named with the deposited values.}
    \label{fig:si-urban-core-distance-decay}
\end{figure*}

\paragraph{Geographic distance decay.}
The largest valid 2024 comparison retains 992 of the 1,000 cities in 162 countries; eight lack valid observations in urban centres. Across its 491,536 unordered pairs, mean angular distance rises and the overlap between the leading ten directions falls mainly within the first 5,000\,km, and splines fitted with cities excluded give $R^2=0.413$ and 0.297 respectively (Supplementary Fig.~\ref{fig:si-urban-core-distance-decay}). These curves describe how similarity falls with distance; each city enters many pairs, so the observations are not independent of one another.

\subsubsection{Principal directions on the sphere and their transfer}

\begin{figure*}[p]
    \centering
    \includegraphics[width=0.97\textwidth,keepaspectratio]{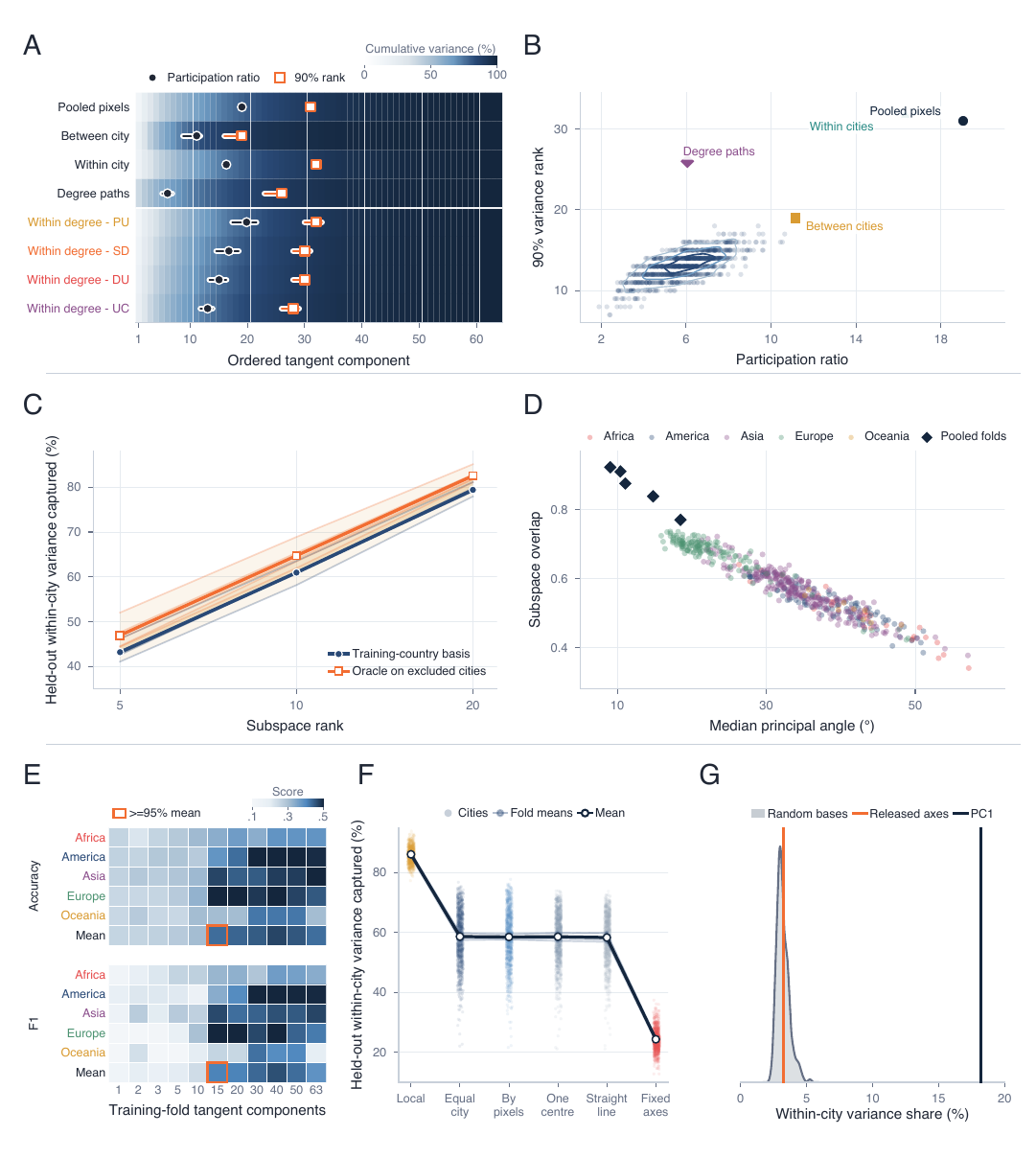}
    \caption{\textbf{Dimensionality depends on which covariance on the sphere is asked about, and shared covariance transfers far better than the directions of a single city.} Participation ratio, in circles, and the rank reaching 90\% of variance, in squares, sort by which covariance is asked about, not by any fixed property of the representation: the aggregates sit high and an individual city far lower (A,B). Transfer separates the constructions. A basis of ten directions from the training countries reaches 94.2\% of what an oracle of the same rank, ten directions fitted to the excluded cities themselves, manages on those cities (C), yet an excluded city's own directions lie a median 33.0\textdegree{} from that basis where two pooled folds lie 11.1\textdegree{} apart (D), and capture falls from 86.1\% on bases fitted inside a city to 24.4\% on ten fixed released axes (F). Classification is cheaper, 15 components reaching 95\% of the rank-63 mean on both scores (E), and no released coordinate carries the leading component's weight---a median 3.11\% across 500 random bases against 18.20\% (G).}
    \label{fig:si-spherical-pca}
\end{figure*}

In the 2024 comparison, quantities computed over a whole city use all 1,000 cities, and those requiring equal samples use the 544 cities that carry all four degrees. Pooled pixels have participation ratio 19.05 and need 31 components for 90\% of variance; covariance within cities, each weighted equally, has ratio 16.32 and needs 32. By contrast, a typical city's own covariance has ratio 5.95 (interquartile range 4.85--6.87) and needs 13 components (12--14). Ratios within each degree decline from 19.86 in peri-urban areas to 13.06 in urban centres, while their 90\% ranks decline from 32 to 28. Dimension therefore depends on which covariance is asked about rather than on a single property of AlphaEarth (Supplementary Fig.~\ref{fig:si-spherical-pca}A,B), and the ratios for pooled pixels and for variation within cities are point estimates without intervals.

Shared directions transfer efficiently relative to an oracle of the same rank. At rank ten, a basis fitted on the training countries captures 60.9\% of the complete covariance in the excluded countries, and 94.2\% of what an oracle of rank ten fitted there itself could capture. The pooled training and test subspaces differ by 11.1\textdegree{} on average, whereas the median angle between an excluded city's own directions and the training basis is 33.0\textdegree{} (Supplementary Fig.~\ref{fig:si-spherical-pca}C,D).

Classifying takes fewer directions than reconstructing. With one continent left out at a time, the first rank tested that retains 95\% of the mean score at the full 63 directions is 15 for both balanced accuracy and macro-F1, whose means at 63 directions are 0.425 and 0.383. The task classifies the directions of the four degree means rather than the path a city traces through the degrees of urbanisation (Supplementary Fig.~\ref{fig:si-spherical-pca}E).

A gap between a shared basis and a city's own could be an artefact of how the shared basis is built, so we vary those choices across all cities. An oracle fitted inside a city on its own leading ten directions captures 86.1\% of that city's covariance, a basis shared across cities weighted equally 58.6\% and ten fixed released axes 24.4\%; weighting by pixel, using one shared tangent centre and measuring covariance along straight lines all land within 58.3--58.6\%. In the pooled covariance within cities, the largest share any single coordinate carries under 500 random orthogonal bases has a median of 3.11\% (2.55--4.39\%), whereas the leading component, which does not move under rotation, carries 18.20\%. Individual released coordinates, their signs and their loadings therefore carry no meaning of their own (Supplementary Fig.~\ref{fig:si-spherical-pca}F,G).

\clearpage

\clearpage

\clearpage

\clearpage
\subsubsection{Placebo footprints}

Pooling cities raises the apparent dimensionality of the residual, and a placebo asks how much of that belongs to cities rather than to any landscape of the same size. Copying each city's footprint to non-urban land and sampling both on land pixels leaves 495 of 504 pairs, in 138 source and 90 destination countries; the other nine have a member that fell almost wholly on open water.

\begin{figure*}[p]
    \centering
    \includegraphics[width=\textwidth,keepaspectratio]{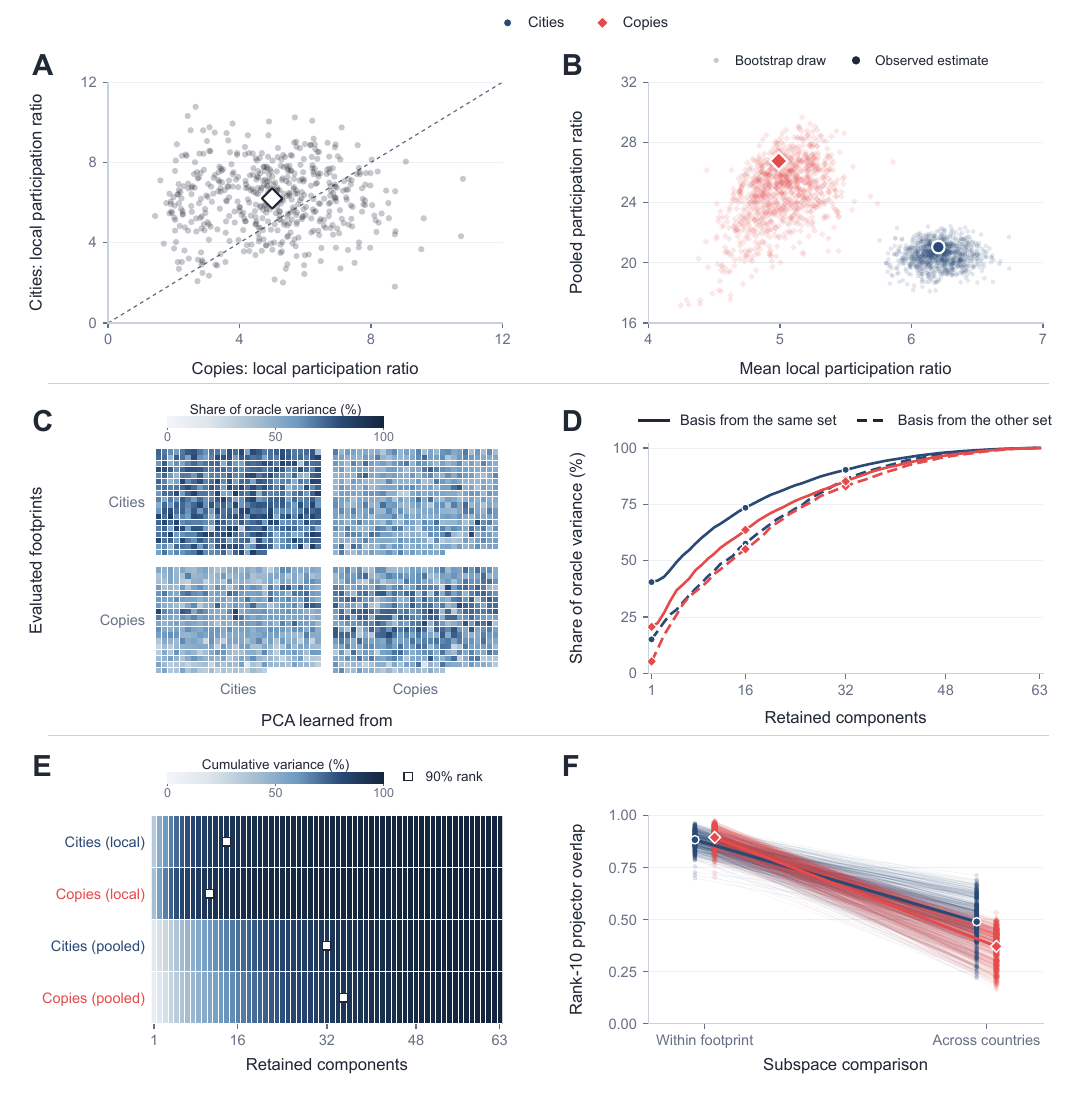}
    \caption{\textbf{Copied to non-urban land, a city's footprint pools to higher dimension than the city itself; what cities keep is structure that transfers.} Both the city and its copy are locally simple (A), yet once footprints are pooled the copies span more directions than the cities, the two bootstrap clouds sitting well apart (B). Transfer runs the other way. A basis learned in other countries' cities recovers most of an excluded city's variance and a basis learned in their copies noticeably less, the dark diagonal blocks against the paler off-diagonal ones (C), with the same gap at every rank (D); pooled cities also reach 90\% of variance sooner (E). Within a footprint the leading directions are stable across spatial halves in both sets (F), so the contrast is not sampling noise.}
    \label{fig:si-placebo-pca}
\end{figure*}

The copies reproduce the expansion. Taken one at a time, a copy has a mean participation ratio of 4.99 against 6.20 for its city, a paired difference of 1.21 (0.67 to 1.72) over 1,000 bootstrap draws that resample the cities' countries and the copies' countries independently. Pooled, the copies span more directions than the cities, 26.76 against 21.04, a difference of $-5.72$ ($-7.94$ to 0.71). The expansion from a single footprint to the pool is therefore a property of sampled landscapes rather than of cities, and of the two it is cities that pool to the lower dimension. The two part in how well their directions transfer. A basis learned in other countries' cities retains 69.3\% of the variance an oracle of the same rank captures in an excluded city, averaged over ranks one to thirty, where a basis learned in other countries' copies retains 53.9\% on the same cities, an advantage of 15.4 points (12.0 to 19.2). Settlement labels copied from a city to its placebo likewise separate the city's pixels more than the copy's, 7.9\% of variation against 3.2\%. Part of that contrast could be climate, because a copy moved far enough leaves its source climate behind: a companion set of 224 pairs whose copies stay inside the source continent and Köppen--Geiger climate family keeps the advantage in transfer, 17.9 points (14.7 to 20.5), and keeps the pooled gap, but halves the difference within a single footprint to 0.58 with an interval that covers zero (Supplementary Fig.~\ref{fig:si-placebo-pca}). Sampling a whole footprint on land keeps rural classes that the sample stratified by settlement class excludes, so these ratios are not interchangeable with the 1,000-city values above.

\subsubsection{National development, measured form and the opportunity to observe}
\label{sec:supp-results-development}

\begin{figure*}[p]
    \centering
    \includegraphics[width=\textwidth,height=0.70\textheight,keepaspectratio]{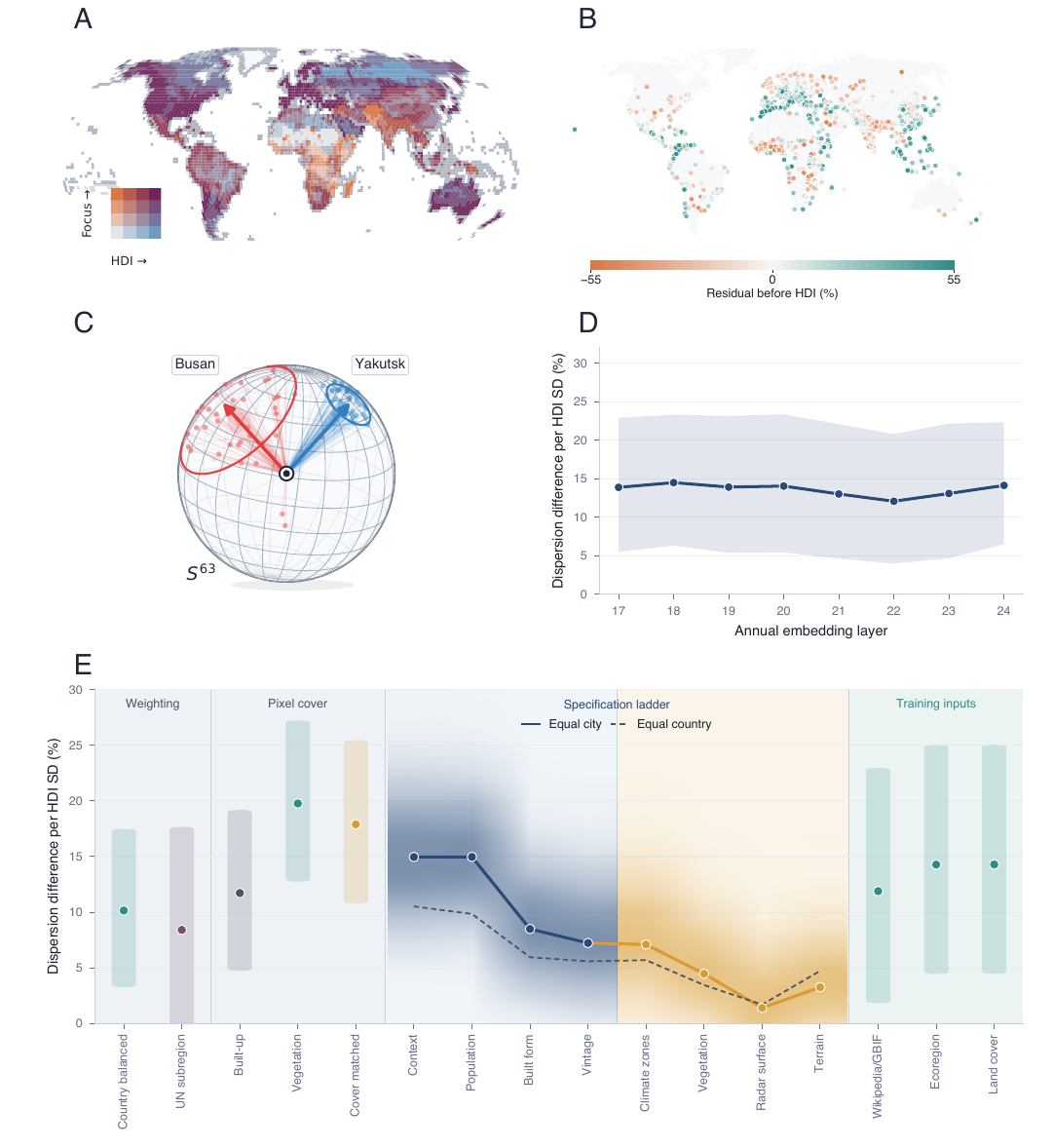}
    \caption{\textbf{Dispersion rises with national development in every year and under every specification, and the two choices that move it are geographic.} Released training sites concentrate where development is high (A), and the residual after context is mapped for every city (B). A city's pixels fan out around its mean direction, and dispersion is the width of that fan: 41\textdegree{} for Busan, 14\textdegree{} for Yakutsk (C). The association stays between 12.1 and 14.5\% in every annual layer (D). Equal-country weighting lowers it to 10.2\% and UN-subregion adjustment to 8.4\% with an interval crossing zero; built form takes it from 15.0 to 8.5\% and the environmental blocks to 1.4\%, a terrain block added on the 937 cities the elevation model covers leaves 3.3\%, and the proxies for released training sites leave 11.9 to 14.3\% on the 516 cities that carry them (E). HDI is national context, not a causal treatment, and focus maps the density of released v2.0 training sites rather than what AlphaEarth was exposed to.}
    \label{fig:si-development-robustness}
\end{figure*}

Dispersion within urban centres rises with national development, and no route to the outcome changes that. After adjustment for population, land area, stage of collection and continent, one standard deviation higher HDI is associated with 14.1\% greater dispersion across 977 cities in 157 countries (95\% interval 6.5 to 22.3\% when countries are resampled), and the estimate stays between 12.1 and 14.5\% in every annual layer from 2017 to 2024 (Supplementary Fig.~\ref{fig:si-development-robustness}D). Two supporting outcomes on the expanded catalogue give 16.7\% (9.3 to 24.6\%) and 18.3\% (11.5 to 25.5\%). Supplementary Table~\ref{tab:si-development-sensitivities} gathers those outcomes computed both over every valid cell a degree covers and over the capped sample of 250 pixels, together with controls for the extent of the urban centre and three ways of forming intervals. Extent works against the association: higher-HDI cities in this catalogue have smaller urban centres, so carrying it suppresses the estimate rather than producing it.

\begin{table*}[p]
    \centering
    \fontsize{8}{10}\selectfont
    \setlength{\tabcolsep}{4pt}
    \begin{tabular*}{\textwidth}{@{\extracolsep{\fill}}llrrl@{}}
        \toprule
        \multicolumn{5}{l}{\bf A. Every dispersion outcome, computed both ways} \\
        \midrule
        {\bf Outcome} & {\bf Reduced over} & {\bf Cities} & {\bf \% per HDI SD} & {\bf 95\% CI} \\
        \midrule
        Urban centre (principal) & every valid cell & 977 & 14.13 & 6.50 to 22.32 \\
        Urban centre (principal) & 250 sampled pixels & 977 & 16.04 & 9.00 to 23.53 \\
        All available urban      & every valid cell & 985 & 16.72 & 9.30 to 24.64 \\
        All available urban      & 250 sampled pixels & 985 & 18.75 & 11.69 to 26.26 \\
        Balanced four degrees    & every valid cell & 532 & 17.52 & 9.95 to 25.62 \\
        Balanced four degrees    & 250 sampled pixels & 537 & 18.31 & 11.53 to 25.51 \\
        \bottomrule
    \end{tabular*}

    \vspace{0.8em}
    \begin{tabular*}{\textwidth}{@{\extracolsep{\fill}}lrrl@{}}
        \toprule
        \multicolumn{4}{l}{\bf B. Urban-centre extent added to the principal specification (977 cities, 157 countries)} \\
        \midrule
        {\bf Specification} & {\bf \% per HDI SD} & \multicolumn{2}{l}{\bf 95\% CI} \\
        \midrule
        Published: log Functional Urban Area population and area & 14.13 & \multicolumn{2}{l}{6.50 to 22.32} \\
        $+$ log urban-centre area                                & 13.99 & \multicolumn{2}{l}{6.58 to 21.92} \\
        $+$ log urban-centre area and number of urban centres    & 13.85 & \multicolumn{2}{l}{6.55 to 21.65} \\
        $+$ the same two and log urban-centre density            & 14.01 & \multicolumn{2}{l}{6.67 to 21.87} \\
        $+$ log urban-centre population                          & 14.06 & \multicolumn{2}{l}{6.41 to 22.26} \\
        $+$ log urban-centre reduction cells                     & 14.13 & \multicolumn{2}{l}{6.74 to 22.02} \\
        Urban-centre population and area replacing the pair above & 15.65 & \multicolumn{2}{l}{7.73 to 24.14} \\
        \bottomrule
    \end{tabular*}

    \vspace{0.8em}
    \begin{tabular*}{\textwidth}{@{\extracolsep{\fill}}lrl@{}}
        \toprule
        \multicolumn{3}{l}{\bf C. Three inference regimes on the principal coefficient} \\
        \midrule
        {\bf Regime} & {\bf 95\% CI (\%)} & {\bf $p$ against zero} \\
        \midrule
        Country-clustered with a $t$ reference (reported throughout) & 6.50 to 22.32 & 0.00023 \\
        Cluster jackknife with a $t$ reference & 5.96 to 22.93 & 0.00057 \\
        Wild cluster bootstrap, Rademacher, null imposed, 9,999 draws & 6.46 to 23.28 & 0.00030 \\
        \bottomrule
    \end{tabular*}

    \vspace{0.8em}
    \begin{minipage}{\textwidth}
        \textit{Notes.} Every row uses the same specification---log dispersion
        on standardised HDI, log population, log land area, sampling stage
        and continent---with country-clustered intervals unless stated. Panel A's
        two routes measure the same city either over every valid cell of its footprint or on the capped 250-pixel sample, and sample sizes differ only
        where a city lacks a valid summary in one of the four degrees. In panel
        B log urban-centre area carries its own coefficient of $-7.36$\% per
        standard deviation ($-15.74$ to $+1.86$), and its partial correlation
        with HDI given log population and log Functional Urban Area area is
        $-0.197$. Panel C's bootstrap interval inverts the test rather than
        dividing by a standard error.
    \end{minipage}
    \caption{\textbf{The association with national development survives every route to the outcome, every control for the extent of the urban centre and all three ways of forming intervals.} Every cell is the association between one standard deviation of national HDI and log embedding dispersion. Across the three panels it ranges from 13.9 to 18.8\% per standard deviation, and no interval reaches zero.}
    \label{tab:si-development-sensitivities}
\end{table*}

Two geographic choices move the estimate more than any covariate does, because countries hold very different numbers of cities and continents differ in how much of the HDI range they cover. In the fixed 977-city sample, weighting every country equally leaves 10.2\% (3.3 to 17.5\%), replacing continent with UN subregion leaves 8.4\% with an interval from $-0.1$ to 17.7\% that includes zero, and omitting Africa leaves 7.2\% (1.7 to 13.1\%) across 784 cities in 109 countries (Supplementary Fig.~\ref{fig:si-development-robustness}E). Because HDI is standardised inside each sample, those are contrasts on different scales: its standard deviation runs from 0.110 without Africa to 0.168 without Asia. On a common increment of 0.1 HDI the full sample gives 9.8\% (4.6 to 15.4\%) and the sample without Africa 6.6\% (1.6 to 11.8\%). That rescaling compares point estimates and does not test their difference; omitting each continent in turn spans 7.2 to 20.7\% per standard deviation within the sample, Africa the lower bound. Prediction in unseen countries agrees: adding HDI to population and land area raises $R^2$ in those countries by 8.7 points (1.0 to 20.9), and adding continent afterwards by $-1.7$ ($-13.8$ to 9.6).

Adjustment for what can be measured on the ground removes most of the gradient, though the order of removal decides the credit. In the fixed 942-city ladder used in Fig.~\ref{fig:development-temporal-dynamics}B the context baseline of 15.0\% (7.5 to 22.9\%) falls to 8.5\% when the distribution of built form enters, and to 1.4\% ($-3.8$ to 6.9\%) once settlement vintage, detailed climate, vegetation and radar surface follow (Supplementary Fig.~\ref{fig:si-development-robustness}E). That last interval includes zero, so the reduction is read as a paired difference of 0.125 log units (0.064 to 0.172), which excludes zero. Averaged over every ordering the same reduction divides differently (Table~\ref{tab:shapley-blocks}): vegetation, radar surface structure and built form take near-equal shares, and entering third inflates the contribution of built form to 1.61 times its average across orderings.

How much of the gradient survives depends on how measured surface structure is represented, so we fit both specifications on identical rows. On the 938 cities in 155 countries covered by both, the ladder's raw backscatter summaries leave 1.4\% ($-3.8$ to 6.9\%) and the optical--radar index defined in Supplementary Section~\ref{sec:supp-alternatives} leaves 6.4\% (0.05 to 13.2\%), an interval excluding zero. The gap of 0.047 log units against that index, entered as its two components, separates exactly into two causes, both excluding zero: $-0.027$ from the covariate, raw backscatter against the index, and $-0.020$ from the summary statistic, a mean over every pixel against a median over the 250 sampled ones. Neither dominates, at 57 and 43\% of the gap.

The public record narrows the candidate mechanisms without identifying one. AlphaEarth's own documentation supplies two candidate channels and rules out a third. Its batch-uniformity objective rotates each training batch along its batch axis and penalises the alignment of an embedding with its rotated partner, assuming a uniform sample from the training set \citep{alphaearth2025}. That is a constraint on the distribution of embeddings rather than on any one of them, so it guarantees no region a share of the sphere and permits unequal dispersion rather than forbidding it. Reading the sphere as allocated in proportion to where the training sites fall is our inference from that objective, not the source's statement, and the objective is a minor term: the supplement settles on a weight of 0.05 and reports 0.005 as the best value in its sweep, against 1.0 for the reconstruction loss. The developers also name the clustering of training sites in cities themselves, and report gridding the stratum seeded from text at 1.28\,km as the step taken to account for it. The second channel is a precedent rather than a mechanism. The National Land Cover Database is a target rather than an input to the encoder, restricted to the conterminous United States, and sampled at half of the training rows. Its presence in the mixture produced a crop-classification regression in that same geography, which was fixed by lowering its weight and adding a second crop target, also confined to that geography \citep{alphaearth2025}. Geographic coordinates, by contrast, are not an input at all, so the gradient cannot be a lookup by position. Where the released training sites fall is the only public proxy for the first channel, and it is steeply uneven: the median count of 1.28\,km chips seeded by Wikipedia or GBIF records within 25\,km of a study city rises from 14 in the lowest HDI tertile to 381 in the highest. That proxy nevertheless does not carry the gradient once measured environment is included, leaving an 8.5\% association (3.4 to 13.8\%) on the earlier 495-city ladder. The v2.1 training sample that produced the layers we analyse is not released, nor is the mixture of sources behind each layer, nor which sensors were present at each chip when the layers were produced.

We call the released density of training sites ``focus'', mapped in Supplementary Fig.~\ref{fig:si-development-robustness}A: how densely a place is represented among the targets the model was trained on. If the gradient were an artefact of where the model was taught, or of how often a city can be seen from orbit, adjusting for either would remove it. Neither does. Entered one at a time against the context baseline, the density of Wikipedia and GBIF sites, the density of ecoregion sites and an indicator for the target confined to the conterminous United States each leave 11.9 to 14.3\% across the 516 cities for which the density of training sites is released. In a separate 214-city subset, adding Sentinel-1, Sentinel-2 and Landsat counts with their clear fractions moves 19.6\% to 18.6\%, an attenuation of 4.7\% whose interval runs from $-7.6$ to $+19.9\%$. None of these fields reconstructs AlphaEarth's unpublished v2.1 inputs.

The association spans the representation rather than concentrating in a few directions. Across the 977 cities with HDI and context, one standard deviation higher HDI is associated with 14.3\% more variance in the leading shared direction, 14.0\% more in the first ten and 19.9\% more in the remaining 53, so the tail gains more than the leading span. Shares of variance show essentially nothing: one of the 63 tests on shares passes a 5\% false discovery rate, the fifty-eighth direction, which carries 0.092\% of the pooled variation within cities.

\begin{figure*}[!t]
    \centering
    \includegraphics[width=\textwidth,keepaspectratio]{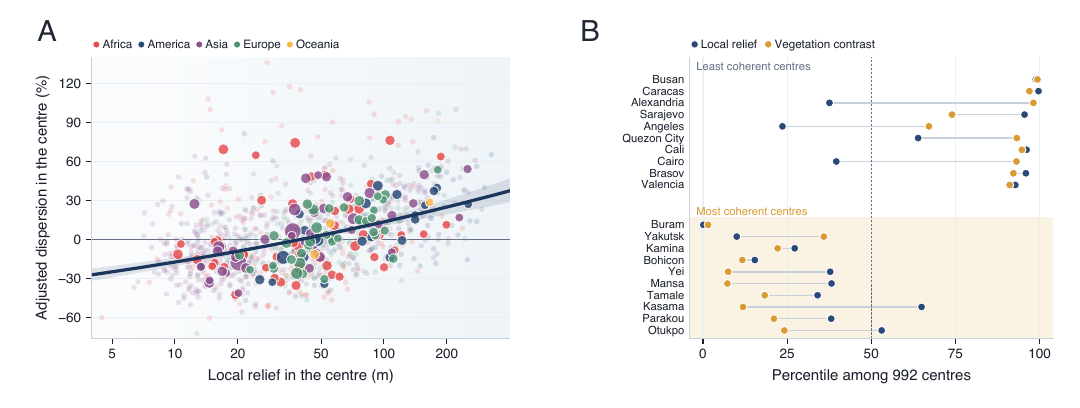}
    \caption{\textbf{Terrain is a second channel for dispersion within cities.} The same population-, area- and continent-adjusted dispersion that Figure~\ref{fig:development-temporal-dynamics}A relates to national development rises 37\% for every tenfold increase in local relief within the centre, with country means sized by city count (A). The ten least coherent centres, those whose pixels spread widest around their mean, divide into rugged cities and flat delta cities with strong vegetation contrast, and the ten most coherent sit low on both, each city placed at its percentile of relief and of vegetation contrast among 992 centres (B).}
    \label{fig:si-terrain-coherence}
\end{figure*}

Terrain is a second channel for dispersion, and it carries almost none of the development gradient. The same adjusted dispersion rises 37\% for every tenfold increase in local relief within the centre across 972 cities in 154 countries (29 to 47\%), a rank correlation of 0.44 (Supplementary Fig.~\ref{fig:si-terrain-coherence}A). That association survives the environmental covariates already in the ladder: it keeps four fifths of its strength when the spread of NDVI and of radar backscatter are held at their ranks, and in a joint model on 939 cities relief, vegetation contrast and radar contrast each add about 10\% per standard deviation with the others fixed, every interval excluding zero. Relief is nearly unrelated to national development, a rank correlation of 0.17, so it predicts which cities are diffuse and leaves the gradient where it was: as an eighth rung the terrain block leaves 3.3\% ($-1.7$ to 8.5\%) on the 937 cities it covers, and as a seventh player it takes 1\% of the reduction ($-21.5$ to 23.1\%), moving the three leading shares by no more than six points. The ten least coherent centres divide into rugged cities, with Busan, Caracas, Cali, Sarajevo, Brasov and Valencia all above the ninety-second percentile of relief, and flat delta cities, with Alexandria and Cairo below the fortieth percentile of relief and above the ninety-third of vegetation contrast; the ten most coherent, nine of them in sub-Saharan Africa and the tenth Yakutsk, sit low on both (Supplementary Fig.~\ref{fig:si-terrain-coherence}B).

The gradient's survival and its removal have the same cause. The two covariates that carry the reduction are themselves patterned by development and by climate: the spread of NDVI within a centre correlates with national HDI at 0.48 and the spread of radar backscatter at 0.42, climate family explains a quarter of the variance of the first and a quarter of the variance of HDI, and mean HDI runs from 0.66 in tropical centres to 0.86 in continental ones. Adjusting for vegetation and radar contrast therefore removes development together with the landscape that accompanies it, and the gradient's persistence under weighting, subregion, land cover and annual layer measures how far those checks leave that entanglement untouched. Relief is the one physical covariate that escapes it. Climate family explains 4\% of its variance and it correlates with HDI at 0.17, yet it predicts dispersion as strongly as vegetation contrast does; it is the only measured channel that separates the landscape from development, and it leaves the gradient where it was.

Cities inside one country share its HDI, so contrasts within countries hold development fixed. Fixed effects for country leave the landscape coefficients near their pooled values across 859 cities in 94 countries: vegetation contrast is associated with 10.8\% greater dispersion per standard deviation within countries against 12.7\% between them, radar contrast with 8.5 against 9.9\% and relief with 7.9 against 8.6\%. No paired difference excludes zero (Supplementary Table~\ref{tab:si-within-country}). The gradient is carried by countries differing in mean landscape: country means of the three covariates take 14.9 to 4.0\% ($-3.3$ to 11.9\%) across 937 cities, a reduction of 0.099 log units (0.030 to 0.165). Across countries on one continent, pairs matched on climate family, relief and population follow development at 11.5\% per standard deviation (4.6 to 20.9\%), and at 7.5\% ($-0.9$ to 19.3\%) once vegetation contrast is matched too. A tenfold increase in relief buys the same dispersion at every level of development, 37.7\% in the lowest third of HDI and 41.8\% in the highest, with an interaction of 1.4\% per standard deviation of HDI ($-4.7$ to 7.8\%).

\begin{table*}[p]
    \centering
    \fontsize{8}{10}\selectfont
    \setlength{\tabcolsep}{4pt}
    \begin{tabular*}{\textwidth}{@{\extracolsep{\fill}}lrlrll@{}}
        \toprule
        \multicolumn{6}{l}{\bf A. Landscape coefficients between countries and within them (859 cities, 94 countries)} \\
        \midrule
        {\bf Covariate} & \multicolumn{2}{l}{\bf Between countries (\% per SD, 95\% CI)} & \multicolumn{2}{l}{\bf Within countries (\% per SD, 95\% CI)} & {\bf Within / between} \\
        \midrule
        Vegetation contrast & 12.7 & 8.7 to 16.9 & 10.8 & 7.2 to 14.4 & 0.86 (0.60 to 1.25) \\
        Radar contrast      &  9.9 & 5.7 to 14.3 &  8.5 & 5.0 to 12.0 & 0.86 (0.59 to 1.32) \\
        Relief              &  8.6 & 5.3 to 12.0 &  7.9 & 5.9 to 9.8  & 0.92 (0.71 to 1.50) \\
        \bottomrule
    \end{tabular*}

    \vspace{0.8em}
    \begin{tabular*}{\textwidth}{@{\extracolsep{\fill}}lrllr@{}}
        \toprule
        \multicolumn{5}{l}{\bf B. The HDI coefficient once countries' mean landscape enters (937 cities, 152 countries)} \\
        \midrule
        {\bf Landscape entered as} & {\bf \% per HDI SD} & {\bf 95\% CI} & {\bf Reduction, log units (95\% CI)} & {\bf Share removed (\%)} \\
        \midrule
        None (principal specification)            & 14.9 & 7.4 to 22.9    & ---                    & --- \\
        City values                               &  4.8 & $-1.6$ to 11.7 & 0.091 (0.047 to 0.133) & 65.9 \\
        Country means                             &  4.0 & $-3.3$ to 11.9 & 0.099 (0.030 to 0.165) & 71.7 \\
        City values and country means             &  4.2 & $-3.0$ to 11.9 & 0.098 (0.035 to 0.162) & 70.6 \\
        Country means of landscape and built form &  2.5 & $-4.2$ to 9.6  & 0.114 (0.036 to 0.184) & 82.4 \\
        \bottomrule
    \end{tabular*}

    \vspace{0.8em}
    \begin{tabular*}{\textwidth}{@{\extracolsep{\fill}}lrrllr@{}}
        \toprule
        \multicolumn{6}{l}{\bf C. Matched pairs within countries and across them} \\
        \midrule
        {\bf Contrast within countries} & {\bf Pairs} & {\bf Countries} & {\bf Spearman, all pairs} & {\bf Spearman, balanced} & {\bf Share (\%)} \\
        \midrule
        Vegetation contrast & 7,082 & 69 & 0.31 (0.09 to 0.38) & 0.23 (0.14 to 0.34)    & 66.4 \\
        Radar contrast      & 6,915 & 68 & 0.21 (0.07 to 0.30) & 0.17 (0.06 to 0.27)    & 61.9 \\
        Relief              & 7,082 & 69 & 0.28 (0.16 to 0.36) & 0.11 ($-0.01$ to 0.23) & 63.5 \\
        \midrule
        {\bf Matched across countries, HDI gap $\geq 0.15$} & {\bf Pairs} & {\bf Countries} & {\bf \% per HDI SD} & {\bf 95\% CI} & {\bf Share (\%)} \\
        \midrule
        Population only                       & 13,396 & 133 & 13.1 & 4.1 to 25.9    & 68.1 \\
        Climate family, relief and population &    707 & 106 & 11.5 & 4.6 to 20.9    & 67.0 \\
        The same and vegetation contrast      &    172 &  81 &  7.5 & $-0.9$ to 19.3 & 58.7 \\
        \bottomrule
    \end{tabular*}

    \vspace{0.8em}
    \begin{tabular*}{\textwidth}{@{\extracolsep{\fill}}lrl@{}}
        \toprule
        \multicolumn{3}{l}{\bf D. Relief against development (972 cities, 154 countries)} \\
        \midrule
        {\bf Statistic} & {\bf Value} & {\bf 95\% CI} \\
        \midrule
        Rank correlation of relief with HDI                                                 & 0.17             & --- \\
        Slope of $\log_{10}$ relief per unit HDI                                            & 0.39             & 0.03 to 0.76 \\
        Centres above the 90th percentile of relief by HDI tertile, low / middle / high (\%) & 8.4 / 12.7 / 9.3 & --- \\
        Rank correlation of relief with HDI within the Americas / within Asia               & $-0.39$ / 0.33   & --- \\
        \bottomrule
    \end{tabular*}

    \vspace{0.8em}
    \begin{tabular*}{\textwidth}{@{\extracolsep{\fill}}lrlrl@{}}
        \toprule
        \multicolumn{5}{l}{\bf E. Dispersion per tenfold relief by level of development (972 cities, 154 countries)} \\
        \midrule
        {\bf Slope per tenfold relief (\%)} & \multicolumn{2}{l}{\bf With continent (95\% CI)} & \multicolumn{2}{l}{\bf With fixed effects for country (95\% CI)} \\
        \midrule
        Lowest third of HDI                          & 37.7   & 25.5 to 51.0    & 26.6    & 12.1 to 42.8 \\
        Middle third of HDI                          & 40.0   & 31.2 to 49.4    & 27.9    & 20.9 to 35.3 \\
        Highest third of HDI                         & 41.8   & 29.2 to 55.6    & 28.9    & 18.7 to 40.0 \\
        Change in the slope per HDI SD (interaction) & 1.4    & $-4.7$ to 7.8   & 0.9     & $-6.7$ to 9.0 \\
        The same interaction for vegetation contrast & $-4.4$ & $-20.9$ to 15.5 & $-13.1$ & $-31.2$ to 9.8 \\
        \bottomrule
    \end{tabular*}

    \vspace{0.8em}
    \begin{minipage}{\textwidth}
        \textit{Notes.} The outcome is log dispersion within urban centres, and
        intervals cluster on countries with a $t$ reference. Panel A fits
        vegetation contrast (log spread of NDVI), radar contrast (log spread of
        VH backscatter), relief (log 90th percentile of 500\,m local relief),
        five built form covariates, log population, log land area and stage,
        once with continent and once with fixed effects for country, on
        identical rows with predictors standardised on those rows; the ratio's
        interval resamples countries within continent, 2,000 draws. Panel B adds
        the three landscape covariates to the principal specification; the
        reduction is a paired difference from the same draws. Panel C pairs
        cities within the 69 countries holding at least five, population within
        0.3 $\log_{10}$ units, and reads the rank correlation of the contrast
        with the difference in log dispersion; China and India hold 5,891 of
        7,082 pairs, so the balanced column weights every country equally. The
        rows across countries pair cities on one continent in the same climate
        family, relief within 0.15 $\log_{10}$ units and population within 0.3,
        the last row also vegetation contrast within 0.15 log units. Share is
        the percentage of pairs in which the more contrasted, or higher HDI,
        city is the more dispersed. Panel E fits log dispersion on $\log_{10}$
        relief, standardised HDI and their product with the same controls; each
        tertile's slope is evaluated at the tertile's mean HDI.
    \end{minipage}
    \caption{\textbf{Landscape predicts dispersion as strongly within countries, where development is held fixed, as between them.} Fixed effects for country leave vegetation contrast, radar contrast and relief at 86 to 92\% of their pooled coefficients (A), country means of the same covariates take the HDI coefficient from 14.9 to 4.0\% (B), pairs across countries follow development less once vegetation contrast is matched (C), relief rises weakly with HDI (D), and a tenfold increase in relief buys the same dispersion in the lowest third of HDI as in the highest (E).}
    \label{tab:si-within-country}
\end{table*}

\clearpage
\clearpage

\subsubsection{Where Sentinel-1B coverage was lost}
\label{sec:si-s1b-shock}

Sentinel-1B stopped delivering radar observations after a power-system anomaly on 23 December 2021 \citep{esa2022sentinel1b}, between the 2021 and 2022 annual AlphaEarth products, and the loss fell unevenly: a location could retain both pass directions, lose one or receive no public Sentinel-1 scene in 2022. If the annual layers track the surface rather than the sensors, nothing about a city should change on that boundary. We test whether movement and dispersion in AlphaEarth follow the timing and the local severity of the loss, a design that shows sensitivity to how a place is observed rather than isolating a causal effect of Sentinel-1B.

\begin{figure*}[p]
    \centering
    \includegraphics[width=\textwidth,keepaspectratio]{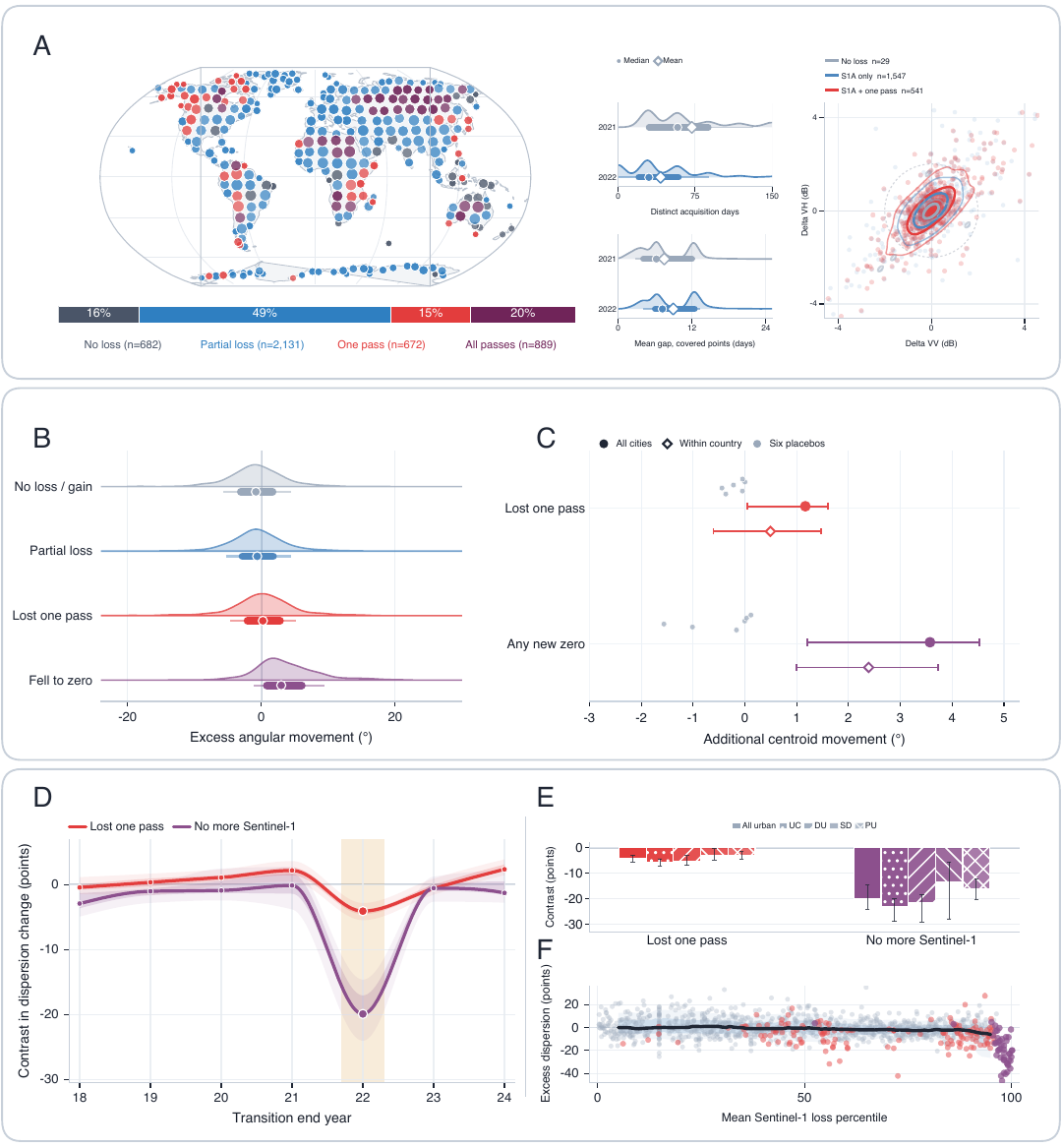}
    \caption{\textbf{Where Sentinel-1B coverage was lost, AlphaEarth moves further and holds less internal variation.} Fixed locations move most where public coverage falls to zero, and replaying the same locations and year through a restricted view shifts the annual radar medians they were built from; both rest on an independent global frame of 4,374 fixed points (A,B). Exposed city centroids then translate $1.17^{\circ}$ and $3.57^{\circ}$ farther than retained ones once background motion within each continent is removed (C). Dispersion contracts at the outage boundary, by 4.08 percentage points where a direction was lost and 19.93 points where every scene was, across all 1,000 cities in 162 countries, negative in every degree of urbanisation and deepest in the cities that lost most, while no other annual transition comes close (D--F). The public scenes intersected with each point measure the opportunity to observe, not AlphaEarth's inputs.}
    \label{fig:si-s1b-shock}
\end{figure*}

Among 4,374 independent fixed points with positive 2021 coverage, losing some scenes while keeping both passes produces no clear excess movement. Losing one pass adds $1.04^{\circ}$ (95\% interval 0.16 to 1.86 when the equal-area cells are resampled), while losing every scene for the year adds $3.79^{\circ}$ (2.56 to 5.49; Supplementary Fig.~\ref{fig:si-s1b-shock}A,B). Replaying the same locations in the same year through a restricted view shifts the 2021 median across the two bands, VV and VH, by 0.50 dB at 1,547 points restricted to Sentinel-1A and by 0.82 dB at the 541 points also restricted to the pass retained in 2022. That shows a pathway through the inputs; it does not reconstruct AlphaEarth's own inputs.

The response also reaches city centroids. After removing background motion estimated within each continent, cities that lost a direction translate $1.17^{\circ}$ more than cities that kept theirs (0.05 to 1.61; 148 exposed against 806 retained), and cities that lost every scene translate $3.57^{\circ}$ more (1.21 to 4.52; 46 against 806; Supplementary Fig.~\ref{fig:si-s1b-shock}C). Two caveats bound that result. Countries differ in ways the comparison cannot see, and comparing exposed and unexposed cities inside a single country would remove them, but only 31 countries contain both a city that lost a direction and a city that did not, and only 10 contain both a city that lost every scene and one that did not, so that comparison agrees in sign on thin evidence. And the placebo on the earlier 2019--2020 transition is not clean: it returns a coefficient of $-0.43^{\circ}$ that survives Holm correction at $p=0.020$, so the outage transition is the largest and the only positive contrast, but it is not the only nonzero one.

Translation is distinct from contraction of the embedding cloud around its centroid, and the contraction is the stronger response. At the outage boundary, losing a direction is associated with a $-4.08$ percentage-point change in dispersion within degrees (95\% interval $-5.88$ to $-2.50$ from resampling countries and then cities; 159 exposed cities in 41 countries against 841), and losing every scene with $-19.93$ points ($-25.81$ to $-10.87$; 46 cities in 15 countries against 954). Both contrasts are estimated across all 1,000 cities in 162 countries, and weights from the sampling design agree (Supplementary Fig.~\ref{fig:si-s1b-shock}D,E). All four degrees of urbanisation move the same way, by 2.87 to 5.65 points where a direction was lost and 13.53 to 23.02 points where every scene was. Applying the same exposure labels to years without a loss should produce nothing; no such transition approaches those values, but two are not clean. The cities that lost every scene had already contracted 2.92 points more than their comparison across 2017--2018 ($-5.42$ to $-0.57$), about a seventh of their outage-year contrast, and the cities that lost a direction expand 2.36 points more across 2023--2024 (0.78 to 4.01).

A direction fitted where the loss was discovered does not transport. The counterpart fitted independently in the untouched cells has cosine 0.162 with it, and its coefficient there is not distinguishable from zero ($p=0.145$). Removing the fixed direction requires all four degrees in every city and so retains 538 cities: it takes out 1.26\% of the total variance under equal city weights but changes no share of the hierarchy by more than 0.07 percentage points, or 0.10 points under design weighting. That restriction belongs to this removal alone and does not carry over to the movement or dispersion analyses.

The pattern of timing and severity supports sensitivity to how a place is observed locally, but not a universal radar direction. The 2024 hierarchy is effectively unchanged. Here contraction means a tighter embedding cloud, not physical change in a city; the public scenes we intersect with each point remain proxies for what the model saw, and the analysis neither isolates a causal effect of Sentinel-1B nor explains the association with national development.

\subsubsection{Annual movement against the floor set by sampling}
\label{sec:si-temporal-floor}

Mean directions drawn from different pixels of the same city, degree and year are close together. Across 215,616 half-sample splits within one degree, the separation rescaled to the full sample size has median 2.579\textdegree{} and root mean square 2.990\textdegree{}; across 63,360 splits pooled to the level of a city the median is 1.419\textdegree{} and the root mean square 1.574\textdegree{}. Both root mean square values barely move from year to year. These are the errors that drawing different places puts on a mean direction, not repeated runs of the model at one location.

Observed annual movement is several times larger. Across the 23,583 adjacent transitions within one degree with an estimable floor, the root mean square annual separation of mean directions is 12.834\textdegree{} against 2.990\textdegree{} expected from finite spatial sampling, a ratio of 4.29. On that collection 94.57\% of pooled squared movement lies beyond the estimated component (95\% interval 93.95 to 95.04\% when cities are resampled). Across the 6,930 transitions pooled to city level the corresponding figures are 12.456\textdegree, 1.573\textdegree, a ratio of 7.92, and 98.41\% (98.35 to 98.46\%). Observed movement exceeds the expected sampling component on every city transition and on 99.90\% of transitions within one degree.

The ordering repeats in every subset we can test: each of the seven adjacent transitions, each of the four degrees and weighting by the sampling design, all of them tabulated with the deposited values. The 2021--2022 Sentinel-1B transition carries the lowest excess ratio at both levels, which is the direction a sensor effect predicts, yet it remains far above sampling error within a year. Drawing different places cannot by itself reproduce the annual step; that is the finding, not that sensor effects are absent.

This excess does not establish physical change: the original samples bound spatial resampling but do not follow fixed sites. The next two sections take up weather and fixed construction sites in turn.

\subsubsection{Weather and annual wobbles}
\label{sec:supp-weather-wobbles}

\begin{figure*}[p]
    \centering
    \includegraphics[width=\textwidth,height=0.70\textheight,keepaspectratio]{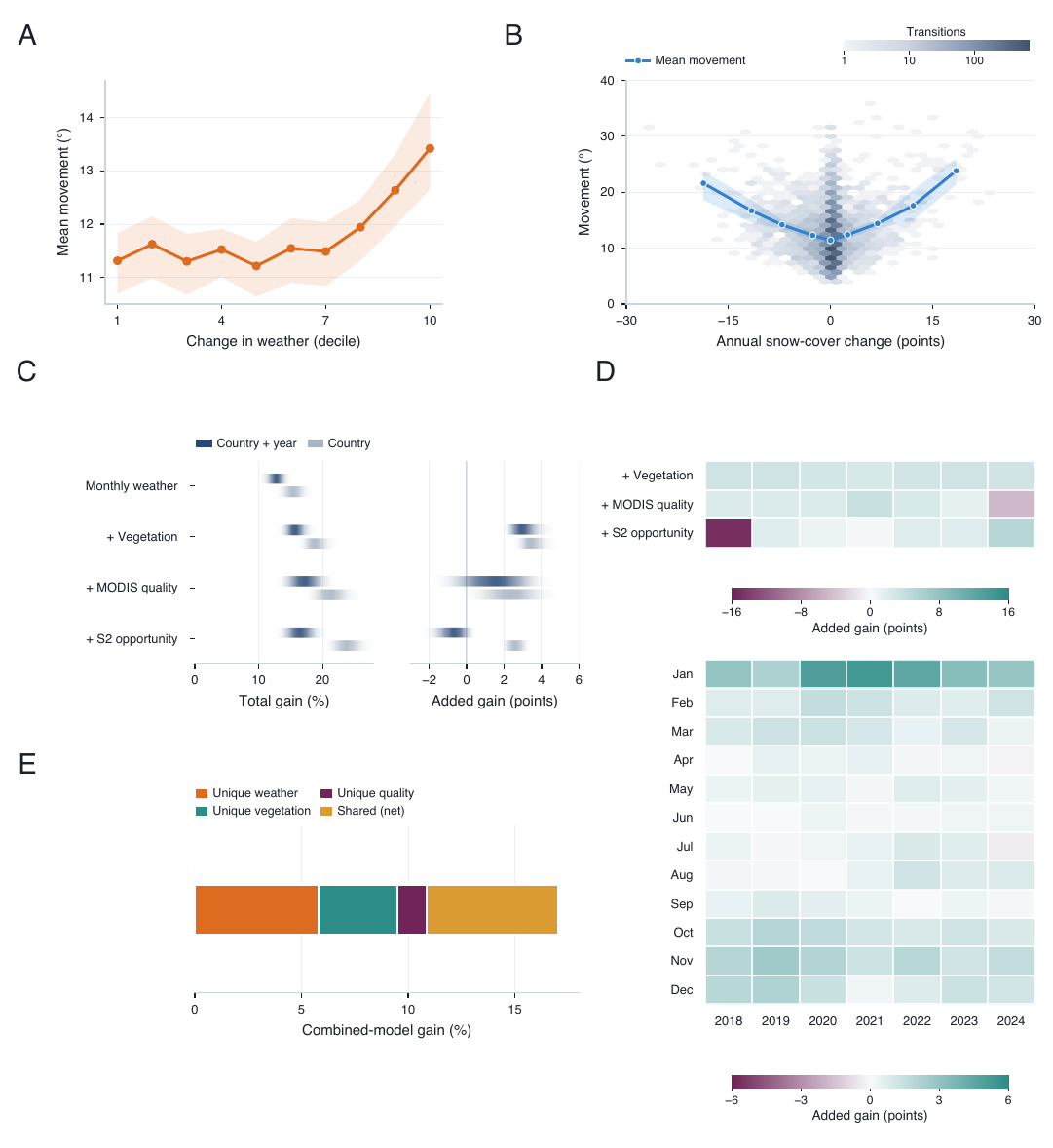}
    \caption{\textbf{Weather and surface state predict part of the annual wobble.} Mean movement rises from 11.3\textdegree{} in the lowest decile of annual weather change to 13.4\textdegree{} in the highest (A), and both gains and losses of snow cover accompany larger movement, with most of the 7,000 transitions crowded near zero change (B). Monthly weather reduces held-out prediction error by 12.7\%, vegetation lifts this to 15.6\% and MODIS quality to 17.0\%, while adding Sentinel-2 opportunity lowers it to 16.4\% (C). By transition year (D), vegetation helps in every cohort, quality turns negative in the last, opportunity fails badly in the first, and removing January's weather costs the most in six of seven years. Of the 17.0\%, 5.8 points are unique to weather, 3.7 to vegetation and 1.4 to quality, with 6.1 shared (E).}
    \label{fig:si-weather-wobbles}
\end{figure*}

Weather reaches the embeddings. A city's annual change can be predicted in part from what the weather did that year: ridge models of monthly ERA5-Land radiation, precipitation, snowfall and snow cover reduce the squared error of that prediction by 12.7\% (95\% interval 11.2 to 14.0\%) in countries and years excluded from fitting, against a baseline of year effects alone (Supplementary Fig.~\ref{fig:si-weather-wobbles}C). The association is visible before any model is fitted: mean movement rises from 11.3\textdegree{} in the lowest decile of annual weather change to 13.4\textdegree{} in the highest (A).

Snow is the sharpest term. Physical snow-cover change in place of locally standardised anomalies raises the weather gain to 15.4\%, and on its own it reduces error by 4.9\% across all cities, by 12.0\% in the 379 cities whose peak climatological monthly cover reaches 1\% and by 0.04\% in the other 621. Gains and losses of cover both move a city: mean movement climbs symmetrically away from zero change (B), where 5,483 of the 7,000 transitions sit, and the association survives adjustment for city and year and reappears in monthly root mean square change, which retains the seasonal swings that annual means cancel.

Vegetation adds to weather rather than repeating it. Monthly MODIS vegetation indices with seasonal timing lift the gain to 15.6\%, adding 3.0 points (2.4 to 3.8) and improving every one of the seven transition cohorts (D). Retrieval quality, viewing geometry and snow flags from the same product lift it again to 17.0\%, although that 1.4-point increment spans zero ($-0.4$ to 3.2) and turns negative in the final cohort. Fitting all seven combinations of the three blocks apportions the 17.0\%: 5.8 points belong to weather alone, 3.7 to vegetation alone and 1.4 to quality alone, with 6.1 shared (E). The overlap does not say how much of that shared part is weather acting through vegetation or clouds.

Weather also governs what the satellites see. Change in the clear-sky fraction follows radiation at a mean monthly rank correlation of $+0.537$ and precipitation at $-0.376$ across 83,984 of the 84,000 pairs of month and transition, and the correlations barely move after adjusting for city, year and acquisition count. Yet measured opportunity itself predicts poorly. Adding Sentinel-2 acquisition counts and cloud scores lowers the joint held-out gain to 16.4\%, improving transfer across countries while failing across years, worst of all in the 2017 to 2018 cohort (C,D). Observing conditions carry information, but that information does not transfer from one year to the next.

Which month's weather matters is stable across years. Refitting the weather model without each month in turn, January is the most costly to remove in six of seven cohorts and November helps in all seven (D). The pattern holds under extreme-shock scoring and when training transitions that share an annual endpoint are excluded, which leaves an overall weather gain of 10.5\%. Weather, surface state and observing conditions therefore predict a transferable part of the annual wobble. What they leave unexplained is not thereby change on the ground, and because these gains measure squared prediction error they cannot be added to the sampling or Sentinel-1B contrasts (Supplementary Section~\ref{sec:supp-movement-tests}).

\clearpage

\clearpage
\subsubsection{Temporal signal at construction sites}
\label{sec:supp-construction}

\begin{figure*}[p]
    \centering
    \includegraphics[width=\textwidth,height=0.74\textheight,keepaspectratio]{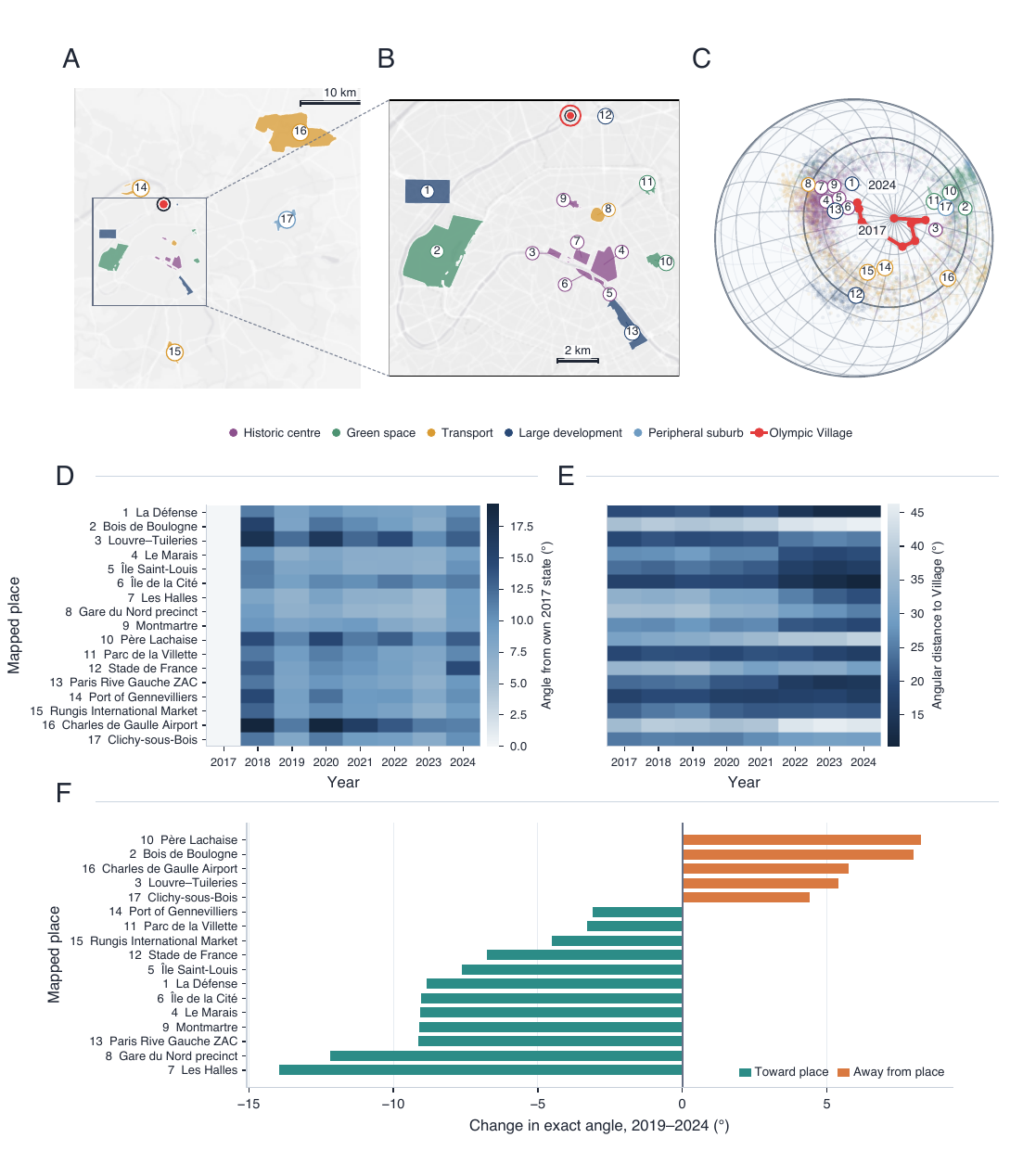}
    \caption{\textbf{The olympic village changes its relation to Paris places as it is rebuilt.} Seventeen comparator places and the 500\,m target circle are mapped (A,B) and followed on one sphere whose pole is the village's 2017 direction, the red path tracing the village and thin paths the comparators (C). Every place drifts from its own 2017 state across the eight layers (D), so the village's movement is read against same-year distances to each place, on which Île de la Cité stays its nearest place at both endpoints (E). Between 2019 and 2024 the village moves 14.0\textdegree{} closer to Les Halles, 12.2\textdegree{} closer to Gare du Nord and 8.2\textdegree{} farther from Père Lachaise (F).}
    \label{fig:si-paris-construction}
\end{figure*}

Where the ground is known to have changed, the embeddings move. The Paris olympic village was deconstructed from November 2019 and its public spaces were delivered in February 2024 \citep{solideoVillageTimeline}, and all 7,847 pixels in the 500\,m circle around it survive the eight annual layers, as do all 554,637 pixels in the 17 comparator places (Supplementary Fig.~\ref{fig:si-paris-construction}). Between 2019 and 2024 the village mean moves 14.0\textdegree{} closer to Les Halles, 12.2\textdegree{} closer to Gare du Nord and 9.1\textdegree{} closer to Paris Rive Gauche, and 8.2\textdegree{} farther from Père Lachaise, while Île de la Cité stays its nearest place at both endpoints. The site does not acquire a single new identity: the largest share of its pixels nearest any one place falls from 34.2\% to 18.6\%, so redevelopment spreads the village across more of the city's places.

\begin{figure*}[p]
    \centering
    \includegraphics[width=\textwidth,height=0.76\textheight,keepaspectratio]{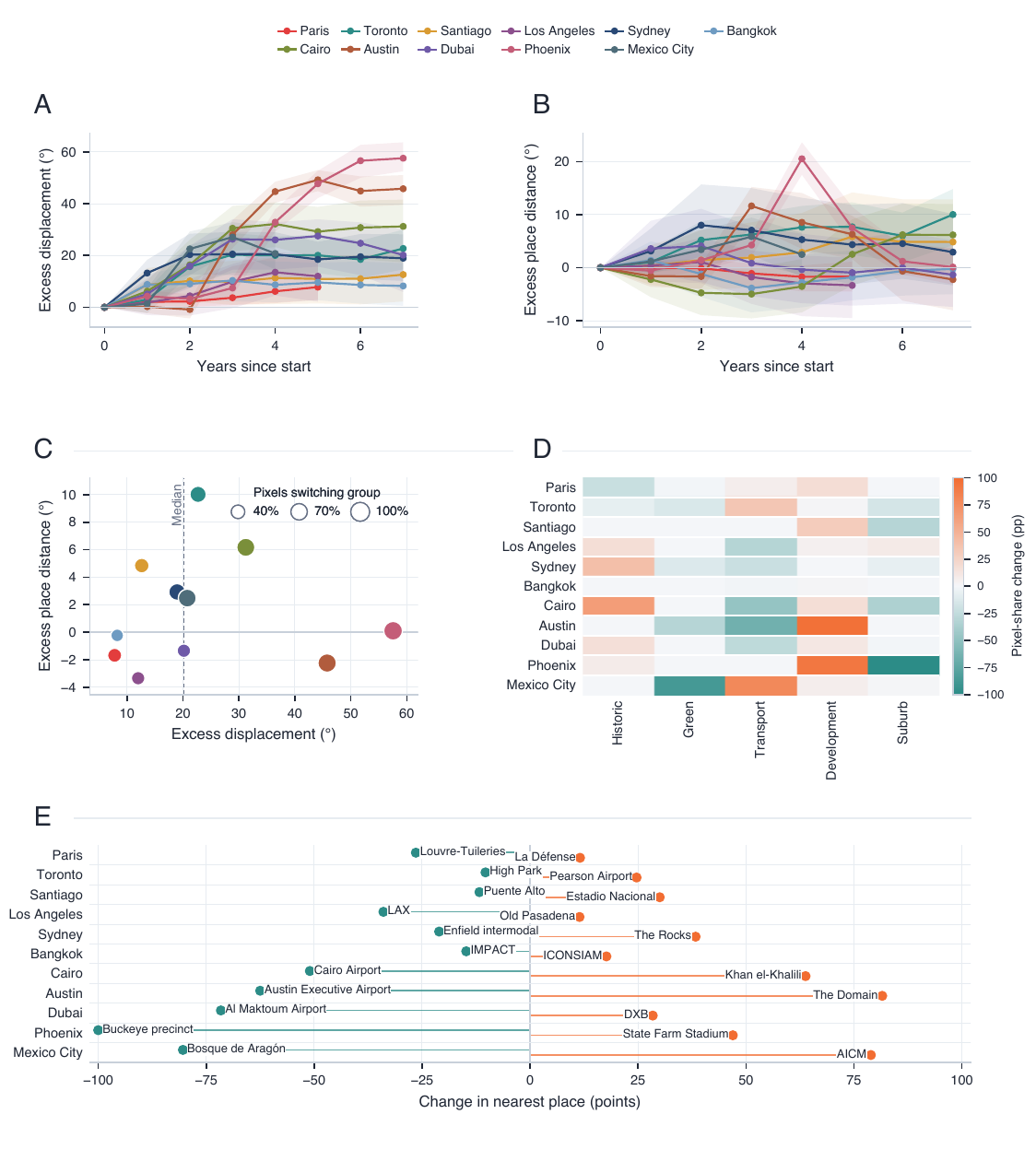}
    \caption{\textbf{Construction sites move farther than matched pixels elsewhere in their cities.} Median excess displacement over matched controls rises above zero at every one of the eleven projects, from 7.8\textdegree{} in Paris to 57.5\textdegree{} in Phoenix (A), whereas the excess change in distance to the nearest named place runs either way, Paris moving 1.7\textdegree{} closer to its local vocabulary and Toronto 10.0\textdegree{} farther (B). At the endpoints (C) the dashed line marks the 20.2\textdegree{} median across projects and marker area the share of pixels that switch comparator group. Which groups gain and lose those pixels differs by city (D), as do the named places that lose and gain most (E), both in percentage points. The largest movers change vocabulary almost wholesale: every Phoenix pixel and 96\% of Austin's switch group, Austin ending nearest The Domain rather than its executive airport, and the airport built at Mexico City ends nearest the city's existing one (C,E).}
    \label{fig:si-construction-diagnostics}
\end{figure*}

Paris is one project. Across eleven, in nine countries, the declared analysis intervals retain 81 city-years and 159,545 fixed target pixels, with 9 to 11 cities present in every year from 2017 to 2024 (Supplementary Fig.~\ref{fig:si-construction-diagnostics}). Every one of them moves farther than its matched controls. The median excess displacement is positive in all eleven cases, from 7.8\textdegree{} in Paris to 57.5\textdegree{} at the semiconductor plant in Phoenix, with a median across projects of 20.2\textdegree{}, and Cairo's new capital keeps its excess when controls are drawn from the narrower functional boundary, which raises it from 31.2\textdegree{} to 38.2\textdegree{}.

Where the sites move to differs. The excess change in distance to the nearest named place runs in both directions: Paris ends 1.7\textdegree{} closer to its local vocabulary than its controls do, and Toronto's Port Lands ends 10.0\textdegree{} farther from it. Mexico City behaves as a positive control should: 81\% of its target pixels begin nearest the Bosque de San Juan de Aragón and 89\% end nearest the city's existing international airport, so the new airport comes to resemble the old one. Construction therefore shows up in the annual layers as movement beyond a city's background at every project we could date. Yet eleven selected projects, measured against controls that are not certified unchanged, cannot say what fraction of movement across 1,000 cities is physical change, and sensitivity to the matching and to the Sentinel-1B discontinuity remains to be established (Supplementary Section~\ref{sec:supp-movement-tests}).

\clearpage

\clearpage
\clearpage